\documentclass[journal]{IEEEtran} 

\usepackage{times}

\usepackage{mathptmx} 
\usepackage{amsmath} 
\usepackage{amssymb}  
\usepackage{subcaption}
\usepackage{graphicx}
\usepackage{epsfig}
\usepackage{mathptmx}
\usepackage{siunitx}
\usepackage{relsize}

\usepackage{makecell}
\usepackage{algorithm}
\usepackage{algpseudocode}
\usepackage{algorithmicx}

\usepackage{wrapfig}

\usepackage[hidelinks]{hyperref}

\usepackage[normalem]{ulem}

\DeclareMathAlphabet{\mathcal}{OMS}{cmsy}{m}{n}

\title{RLOC: Terrain-Aware Legged Locomotion using
Reinforcement Learning and Optimal Control}

\author{Siddhant Gangapurwala, Mathieu Geisert, Romeo Orsolino, Maurice Fallon and Ioannis Havoutis

	\thanks{This work was supported by the UKRI/EPSRC RAIN Hub
[EP/R026084/1] and the EU H2020 Projects MEMMO and THING,
the EPSRC grant ‘Robust Legged Locomotion’ [EP/S002383/1] and a
Royal Society University Research Fellowship (Fallon). This work was
conducted as part of ANYmal Research, a community to advance legged
robotics. The authors are with Dynamic Robots Systems (DRS) group, Oxford Robotics Institute, University
of Oxford, UK. Email: {\tt\footnotesize \{siddhant, mathieu, rorsolino, mfallon, ioannis\}@robots.ox.ac.uk}}
}

\date{}

\begin{document}
\maketitle
\thispagestyle{empty}
\pagestyle{empty}

\begin{abstract}
We present a unified model-based and data-driven
approach for quadrupedal planning and control to achieve
dynamic locomotion over
uneven terrain.  We utilize on-board proprioceptive and
exteroceptive feedback to map sensory information and desired
base velocity commands into footstep plans
using a reinforcement learning (RL) policy. This RL policy is
trained in simulation over a wide range of 
procedurally generated terrains.
When ran online, the system tracks the generated footstep plans using a model-based motion controller. We
evaluate the robustness of our method over a wide variety of 
complex terrains. It exhibits behaviors which prioritize stability over aggressive locomotion.
Additionally, we introduce two ancillary RL policies for corrective whole-body motion tracking and
recovery control. These policies account for changes in physical parameters and external perturbations.
We train and evaluate our framework on a complex quadrupedal system, ANYmal version B,
and demonstrate transferability to a larger and heavier robot, ANYmal C, without requiring retraining.
\end{abstract}


\section{Introduction}
    Legged robot locomotion research has been driven by the goal of developing
    robust and efficient control solutions with the potential to exhibit
    a level of agility that matches their biological counterparts. Their advanced mobility
    is characterized by the ability to
    form support contacts with the environment at discontinuous locations as required.
    This makes such systems suitable
    for traversal over terrains that are inaccessible for wheeled or tracked
    robots of equivalent size. This maneuverability, however, comes
    at the expense of increased control complexity. Such systems often need to
    maintain dynamic stability, select appropriate contact sequences, and actively balance during locomotion.

\begin{figure}
            \begin{subfigure}{.24\textwidth}
              \centering
              \includegraphics[width=\linewidth]{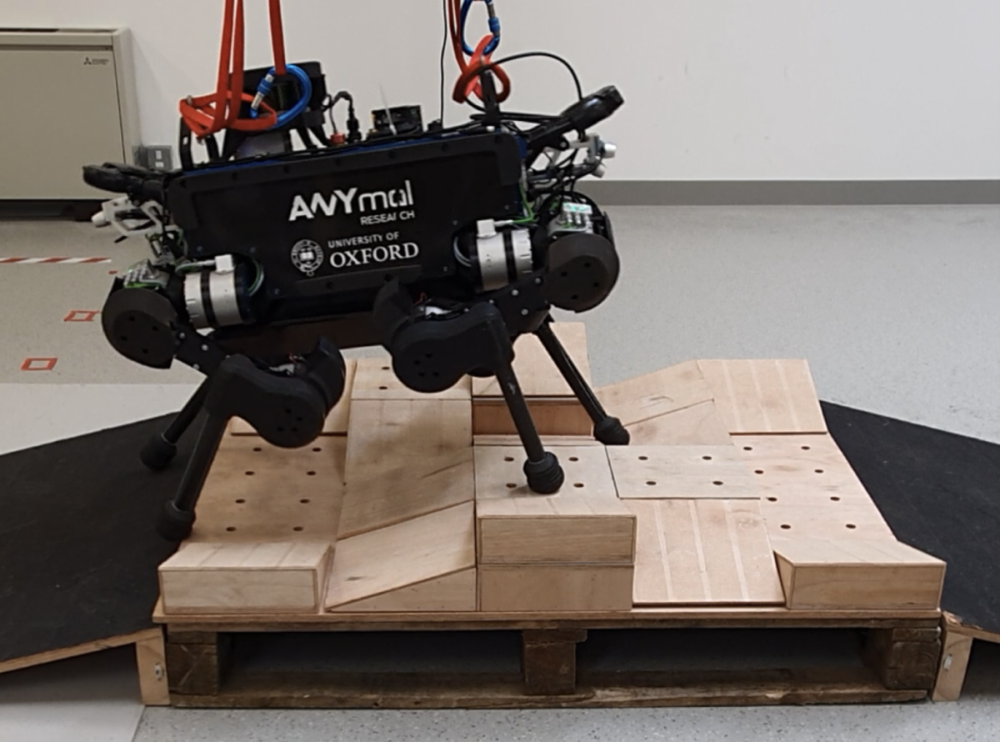}
            \end{subfigure}%
            \begin{subfigure}{.24\textwidth}
              \centering
              \includegraphics[width=\linewidth]{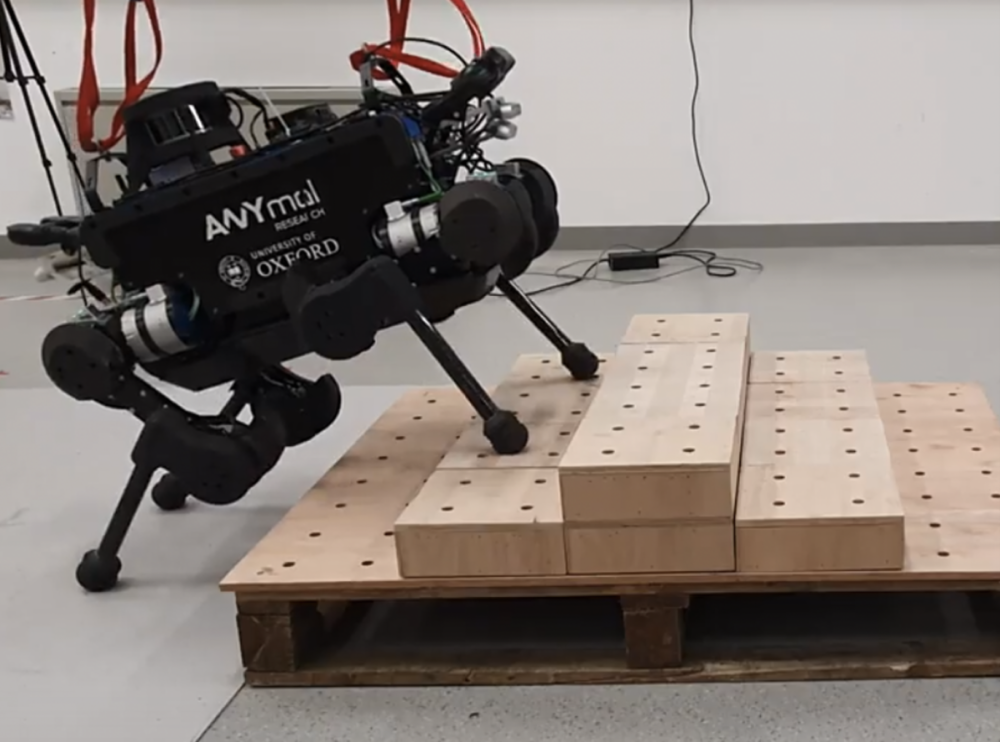}
            \end{subfigure}\\
            \begin{subfigure}{.24\textwidth}
              \centering
              \includegraphics[width=\linewidth]{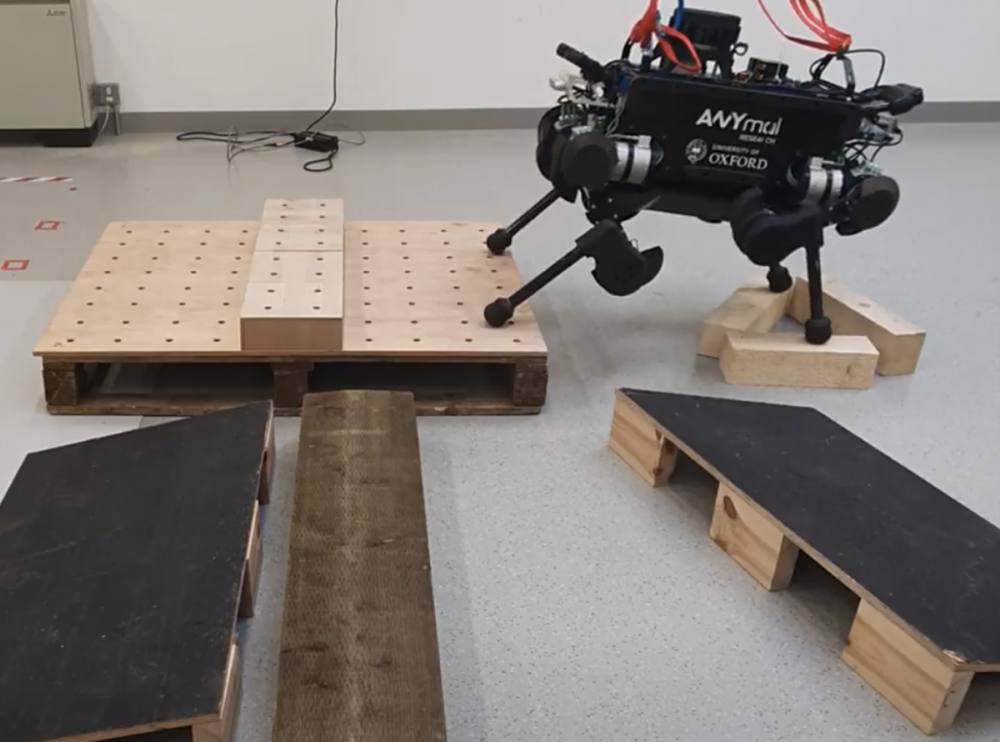}
            \end{subfigure}%
            \begin{subfigure}{.24\textwidth}
              \centering
              \includegraphics[width=\linewidth]{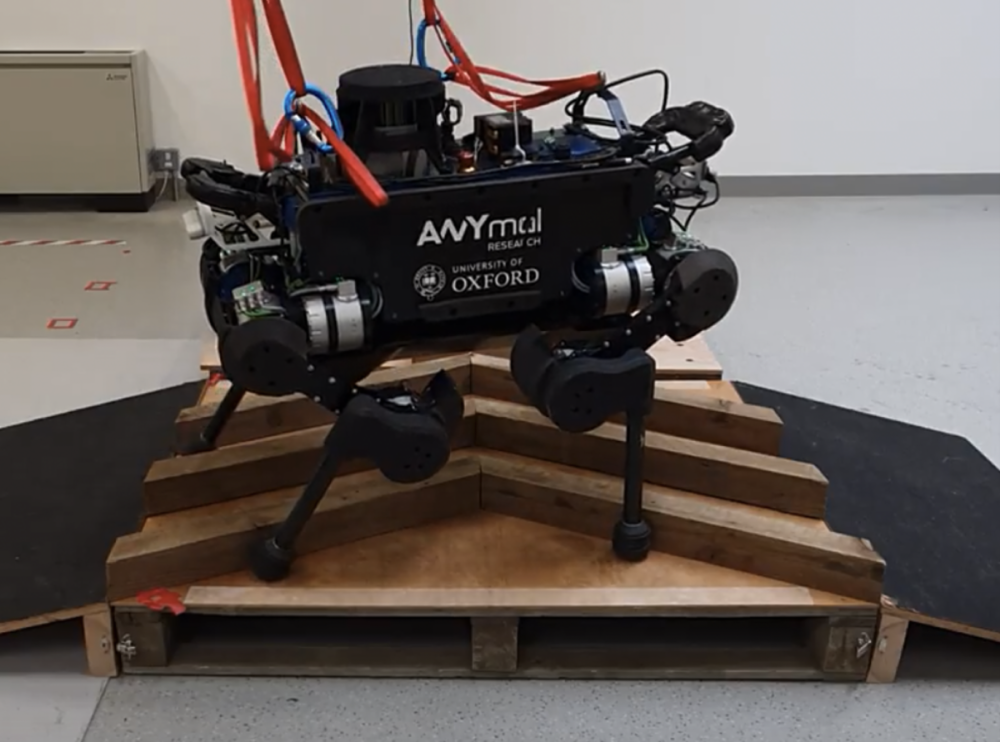}
            \end{subfigure}
            \caption{ANYmal B quadruped used to perform experiments on different terrains using our RLOC framework. Overview video can be found at: \href{https://youtu.be/GTI-0gl6Hg0}{https://youtu.be/{\footnotesize GTI-0gl6Hg0}}.}
            \label{fig:anymal_boxy_real_tests}
            \vspace{-15pt}
        \end{figure}

\begin{figure*}
            \centering
            \includegraphics[width=\textwidth]{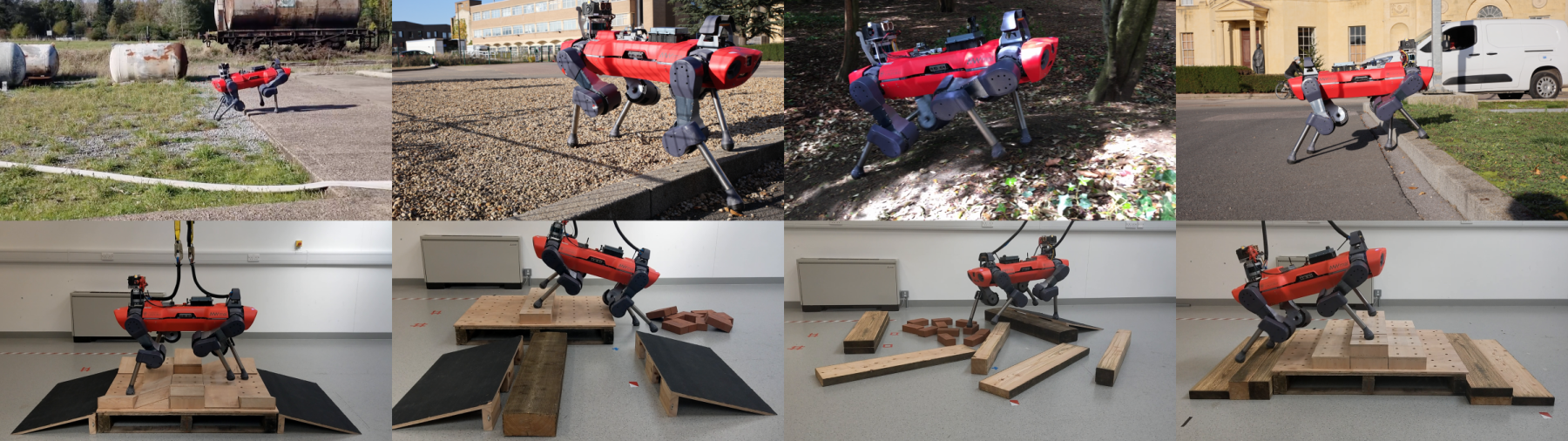}
            \caption{ANYmal C quadruped used to perform experiments on different terrains outdoors; on hard, grassy and muddy terrain also including slopes, and indoors; similar to the experiments performed using ANYmal B. The policies trained using ANYmal B were transferred to ANYmal C without retraining. The video clips of the demonstrations are provided in \href{https://youtu.be/rIr0tyqTjGw}{Movie S6}.}
            \label{fig:anymal_coyote_real_tests}
        \vspace{-12pt}
        \end{figure*}

    In this work, we design, 
    train and evaluate a control framework for legged locomotion
    over uneven terrain
    using the \textit{ANYbotics ANYmal B~\cite{hutter2016anymal}} and \textit{ANYmal C}~\cite{anymalc} quadrupedal robots. ANYmal version B is a \SI{30}{\kilogram} rugged
    robot developed with a focus on high mobility and the ability to perform
    dynamic locomotion behaviors for rough terrain operations. It features
    a large workspace which enables a wide range of leg motions and is suitable for a broad domain of applications, 
    for example,
    inspection tasks on offshore platforms~\cite{anymaloffshore} and sewer systems~\cite{doi:10.1002/rob.21964}. Figure~\ref{fig:anymal_boxy_real_tests}
    shows the ANYmal B robot used for testing our proposed framework on different uneven terrains.
    We additionally highlight the generality of our control framework by testing our approach
    with the newer ANYmal, version C~\cite{anymalc} --a \SI{50}{\kilogram} quadrupedal robot with
    a similar kinematic structure but different dynamic characteristics-- and by evaluating its 
    performance \textit{without retraining} any of our planning and control policies. Experiments
    performed with ANYmal C are shown in Fig.~\ref{fig:anymal_coyote_real_tests}.
    
    The control of such underactuated robotic systems is an active area of research.

    Approaches based on whole-body trajectory optimization such as contact-implicit optimization (CIO)~\cite{mordatch2012discovery} 
    can generate complex multi-contact motions but often remain too computationally extensive to be used online.
    More recent works based on differential dynamic programming (DDP) algorithms~\cite{mastalli2020crocoddyl, grandia2019feedback}
    enable efficient computation of whole-body motions suitable for online execution. These methods, however,
    are still under development, and their effectiveness on physical systems and complex environments is yet to be demonstrated.
    
    A rich body of work has focused on using simplified dynamic models which capture only the main dynamics of the system
    so as to achieve a balance of computational complexity~\cite{Full1999}. 
    Such models are also known as \textit{templates} and are typically lower 
    dimensional than the full (whole-body) model, and also linear with respect to the states of the system; 
    the mechanism that relates them back to the whole-body model is instead usually called the \textit{anchor}. 
    Examples of templates include the linear inverted pendulum (LIP)~\cite{Kajita2001} and the spring-loaded inverted pendulum (SLIP)~\cite{Poulakakis2006},
    and their derived variants~\cite{Poulakakis2009}. 
    
    For the development of robust model-based controllers, 
    the synthesis of stability criteria based on ground reference points (2D points on a flat ground plane)
    such as the zero moment point (ZMP)~\cite{Vukobratovic1968} and the instantaneous capture point (ICP)~\cite{Pratt2007} is also introduced into the control problem.
    The templates can be seen as a way to constrain the admissible domain of such ground reference points. In this regard, the balancing problem for legged robots 
    can be translated into an inclusion problem that requires the reference point to belong to its admissible region, i.e. the \textit{support region},
    for the robot to be stable. In this way, a measure of robustness,
    the stability \textit{margin}, can be defined as the distance between the reference point 
    and the edges of its admissible region.
    
    Often, simplified models 
    and ground reference points are limited to flat terrain and their descriptive 
    accuracy deteriorates when applied to environments with complex geometry.
    For such \textit{multi-contact} conditions, the centroidal dynamics model~\cite{Orin2013} 
    is widely used. It represents the system in a compact manner such that 
    it focuses on the inputs (contact wrenches) and the outputs 
    (center of mass accelerations) in the task space of the robot discarding the intermediate
    joint-space physical quantities. In this case, the reference point for evaluating stability 
    can be described by a 6-dimensional wrench. Therefore, for evaluation of stability, templates that illustrate suitable 
    6D admissible regions need to be defined~\cite{Dai2016}. Despite their low dimensionality, 
    such simplified models are still typically too numerically inefficient to be employed in the robot's 
    real-time control loop.
    An alternative approach is to evaluate the stability of the robot directly using the 
    contact forces~\cite{carpentier2018multicontact}. This method, however, necessitates 
    introduction of the contact force variables in the locomotion problem thereby adding to computational delays which result in 
    reduced reactive control.
    
    Recent research on model-predictive control (MPC) based strategies have demonstrated perceptive locomotion over uneven terrain~\cite{jenelten2020perceptive, kim2020vision}. 
    These controllers, however, 
    often employ system
    model approximations, due to which, the potential of the physical robot is not fully exploited. 

    Moreover, these methods require exhaustive tuning of planning and control parameters, requiring considerable
    development time.
    
    Alternatively, an entirely different approach has focused on investigating 
    solutions which can perform a direct mapping between sensory information
    to desired control signals with minimal computational overhead.
    Model-free data-driven techniques such as RL have introduced
    promising substitutes to model-based control methods, and achieved highly dynamic and
    robust locomotion skills on complex legged robot systems such as 
    ANYmal B~\cite{hwangbo2019learning, lee2019robust, gangapurwala2020guided}, Laikago~\cite{peng2020learning} and Cassie~\cite{xie2018feedback}. The major
    challenge of using RL lies in the fact that training these methods is extremely sample inefficient,
    especially for high-dimensional problems such as legged locomotion. When applied to
    a robot such as ANYmal B, the vast exploration space associated with an RL policy,
    not only necessitates considerable amount of training samples, but also implies that
    without precise reward function tuning, the RL agent could converge towards an
    undesired control behavior. As an alternative, a guided policy exploration approach, such as guided policy search (GPS)~\cite{levine2013guided}, which relies on utilizing model-based control solutions in conjunction with data-driven policy optimization can be employed to direct
    policy convergence and increase sample efficiency. 
    
    The possibility of damaging the robotic system itself and the environment of
    operation during policy exploration necessitates the use of simulation platforms.
    Ease of access
    to physics simulators, and the ability to execute simulations much faster
    than real-time, makes this approach appealing for training RL policies. However,
    the inaccuracy associated with the complex dynamics that are hard to model in simulation often 
    affect the performance of RL policies 
    when deployed onto real systems. 

    Although there have been works where training was done on the real system itself~\cite{Haarnoja2019, yang2020data}, 
    this has been appropriate for simpler and inherently stable robots which cannot be employed for training complex quadrupedal systems such as the ANYmal robots.
    
    To address the reality gap problem, more
    accurate models of the training environment can be directly introduced into the
    simulators. For example, \textit{Hwangbo et al.} introduced an actuator 
    network~\cite{hwangbo2019learning} to model the dynamics of
    the physical actuators and further performed ablation studies to justify the necessity for
    accurate models to achieve simulation-to-real transfer. Moreover, for a training environment, 
    it is very likely that not all the dynamics can be precisely modeled. As a solution,
    domain randomization~\cite{tobin2017domain} techniques have been used to train
    policies which are robust to unaccountable factors~\cite{hwangbo2019learning, gangapurwala2020guided}.

    Most of the above mentioned RL research focuses on quadrupedal locomotion on flat ground, 
    overlooking the primary purpose of quadrupedal systems -- mobility over rough terrain.
    
    \textit{Tsounis et al.} presented an approach for terrain-aware gait planning
    and control using RL for the ANYmal B quadruped~\cite{tsounis2020deepgait}. \textit{Tsounis et al.}
    used a hierarchical approach to plan gaits suitable for navigation to a desired
    goal position, and then performed whole-body control, both using RL. However, different
    gait planning policies were trained for different terrains, while sim-to-real transfer
    was not addressed. \textit{Lee et al.} presented an RL based approach for quadrupedal locomotion over
    challenging terrains~\cite{lee2020learning}, however, this work only focused on use of proprioceptive feedback
    for locomotion without explicitly processing terrain information. Recently, \textit{Miki et al.}
    demonstrated robust and perceptive locomotion skills in
    unstructured environments~\cite{miki2022learning}. Their control framework exhibited impressive terrain-adaptive
    behavior even with invalid exteroceptive feedback. This work, however, did not demonstrate significant robustness to variations
    in kinematic and dynamic properties. For transfer to a different robot not introduced during training, their proposed
    approach is expected to require further retraining.

    \begin{figure}
        \centering
        \includegraphics[width=0.48\textwidth]{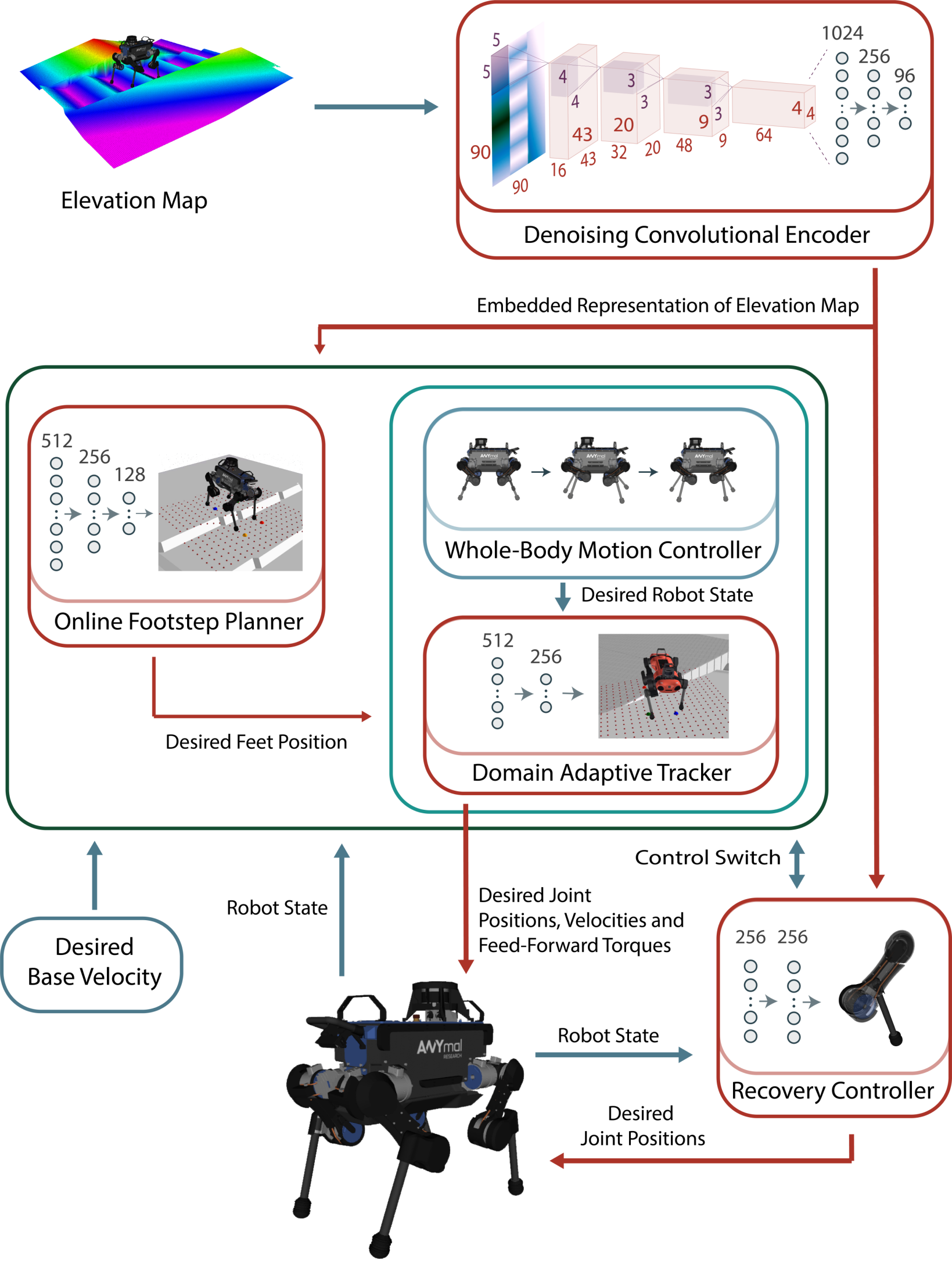}
        \caption{Overview of the proposed control framework - RLOC. We use an online footstep planning RL policy to generate desired feet positions which are tracked using a model-based whole-body motion controller. In order to
        perform accurate tracking even in case of changes in robot dynamics, we introduce a domain adaptive tracking policy to generate feed-forward corrective torques. Furthermore, in order to exhibit a quick response to external perturbations, we introduce a recovery control policy which generates low-level joint position commands.}
        \label{fig:overview}
    \vspace{-20pt}
    \end{figure}

    In this work, we propose a \textit{unified RL and optimal control (OC)} based terrain-aware legged
    locomotion framework as illustrated in Fig.~\ref{fig:overview}. We train 
    a footstep planning RL policy which maps the robot state and terrain 
    information to desired feet positions. In order to track these positions,
    we extend the \textit{dynamic gaits}
    motion controller of \textit{Bellicoso et al.}~\cite{bellicoso2018dynamic} to enable foothold tracking over non-flat terrain.
    This forms the core of our locomotion framework. In order to account
    for changes in physical parameters and actuation dynamics, we train a domain
    adaptive tracking policy which introduces
    corrective feed-forward joint torques for accurate tracking of the trajectories
    generated by the motion controller. Furthermore, to enable
    recovery to a stable state upon perturbations or unexpected events like foot slippage, we train an RL policy to generate
    low-level joint position commands such that during unstable motions,
    the RL policy stabilizes the robot and then the control switches back to the
    footstep planning and tracking modules. A video overview of our proposed framework is presented in
    \href{https://youtu.be/GTI-0gl6Hg0}{Movie 1}.
    
    Our main contribution with this work is realizing a modular framework which combines RL and OC
    based approaches to obtain a system-adaptive control solution for perceptive and dynamic robotic locomotion.
    The adaptive characteristic further facilitates
    robustness to variations in dynamic and kinematic properties of the system.
    The key elements of this work thus
    include (1) Presenting a training approach for obtaining a perceptive
    footstep planning policy which employs a model-based dynamic motion controller.
    This setup enables us to generate valid dynamic motion plans while
    ensuring the obtained policy adapts to and exploits the domain largely described by the characteristics of the motion controller.
    (2) Obtaining a domain 
    adaptive tracking policy which can be utilized to address system modeling inaccuracies during whole-body
    motion tracking. This allows us to use a model-based controller on a robot where the modeled characteristics
    differ from the actual kinematic and dynamic properties of the system. This further enables us to
    perform dynamics randomization while training the footstep planning policy without requiring retuning of the motion
    controller. 
    (3) Demonstrating zero-shot generalization to a 
    domain not explicitly modeled during training. For this,
    we show transfer of our framework to the physical ANYmal C robot despite having been trained in a simulator with ANYmal B.

    Our framework allows for high-level tracking of reference
    base velocity commands over complex uneven terrain, enabling the robot to be operated 
    either by a user (through a remote control) or a higher-level goal planner. We evaluate the performance of RLOC on a wide range of simulated terrains
    and present the results for the ANYmal B and C quadrupeds. We further transfer RLOC on to the real robots.

\section{Preliminaries}
    \subsection{System Model}
        A general quadrupedal robotic system can be modeled as a floating base $B$ with four
        actuated limbs attached to it. The robot state can be described w.r.t. a global inertial
        reference frame $W$ where the base position is expressed as 
        ${}_{W}r_{\,WB}\in\mathbb{R}^3$, and the orientation, $\mathrm{q}_{WB}\in\mathit{SO}(3)$, 
        is represented using Hamiltonian unit quaternions with 
        a corresponding rotation matrix $\mathbf{R}_{WB}\in\mathit{SO}(3)$. We assume that the $z$-axis of $W$, $\mathbf{e}_z^W$,
        aligns with the gravity axis.
        The angular positions of
        the rotational joints in each of the limbs are described as a vector $\mathrm{q}_{j}\in\mathbb{R}^{n_{j}}$.
        For the ANYmal robots, $n_j=12$. The linear and angular velocities of the base w.r.t. the global
        frame are written as ${}_{W}\mathrm{v}_{WB}\in\mathbb{R}^3$ and ${}_{W}\mathrm{\omega}_{\,WB}\in\mathbb{R}^3$
        respectively. The generalized coordinates and
        velocities are thus stacked as vectors $\mathrm{q}$ and $\mathrm{u}$ where
        
        \begin{equation}
            \mathrm{q} = \begin{bmatrix}
                {}_{W}r_{\,WB} \\
                \mathrm{q}_{WB} \\
                \mathrm{q}_{j}
            \end{bmatrix} \in\mathit{SE}(3)\times\mathbb{R}^{n_j},\quad \mathrm{u} = \begin{bmatrix}
                {}_{W}\mathrm{v}_{WB} \\
                {}_{W}\mathrm{\omega}_{\,WB} \\
                \mathrm{\dot{q}}_{j}
            \end{bmatrix} \in\mathbb{R}^{6 + n_j}.
        \end{equation}

    We denote positions of the feet in the global reference frame as ${}_{W}r_{\,WF}\in\mathbb{R}^{3\times n_{f}}$, where
    $n_f=4$ is the number of feet. We additionally define ${}_{W}r_{\,WF_{xy}}\in\mathbb{R}^{2\times n_{f}}$ as a vector
    of $x$ and $y$ components of the feet positions.
    
    We introduce a horizontal frame $H$ such that ${}_{W}r_{\,WH}={}_{W}r_{\,WB}$ and
    $\mathbf{R}_{WH}=\mathbf{R}_{WB_z}$, where $\mathbf{R}_{WB_z}$ is the $z$ decomposition of the
    base rotation matrix which can also be expressed as $\mathbf{R}_{WB}=\mathbf{R}_{WB_z}\mathbf{R}_{WB_y}\mathbf{R}_{WB_x}$. 
    The frame $H$ can thus be interpreted as having the same translation as the robot base frame $B$ w.r.t. the global inertial
    reference frame, but rotation
    in the frame $H$ is only permissible along $\mathbf{e}_z^H$, implying that the gravity axis always aligns with the $z$-axis of frame $H$. Additionally, the yaw rotation of frame $H$ also always aligns with the yaw rotation of frame $B$.
    We then represent the local robocentric terrain 
    elevation in the horizontal frame as 
    $\mathbf{M}_{H}:\mathbb{R}^2\times\mathbb{R}\rightarrow\mathbb{R}^{91\times91}$, an elevation map
    corresponding to an area of $1.8\times1.8\,$\si{\metre\squared}
    and a resolution of \SI{0.02}{\metre}.
    
    The quadrupedal system is actuated using the joint control torques $\tau_j\in\mathbb{R}^{n_j}$. These 
    torques are computed at the actuator level using an impedance control model written as
    
    \begin{equation}
        \tau_j = K_p(\mathrm{q}^{\ast}_j-\mathrm{q}_j) + K_d(\mathrm{\dot{q}}^{\ast}_j-\mathrm{\dot{q}}_j) + \tau_{j_{FF}}
    \label{eq:impedance_controller}
    \end{equation}
    where $K_p$ and $K_d$ refer to the position and velocity tracking gains respectively, $\mathrm{q}^{\ast}_j$ is the
    vector representing desired joint positions, $\mathrm{\dot{q}}^{\ast}_j$, the desired joint velocities, and $\tau_{j_{FF}}$ refers to the
    feed-forward joint torques.
    
    Note that, for concise presentation, we will assume that all of the translations and rotations
    are measured and expressed in the global reference frame $W$ unless otherwise stated. In this regard, for the rest of the manuscript,
    the notation ${}_{W}r_{WB}$ will
    be shortened to $r_B$, $\mathbf{R}_{WB}$ to $\mathbf{R}_{B}$, and similarly for all the other introduced notations.
    
    \subsection{Motion Parameterization}
        The motion of the quadrupedal system is expressed in terms of the center of mass (CoM) and feet motions. In consistency
        with the definitions introduced in~\cite{bellicoso2018dynamic}, we represent the CoM motion plan in the form 
        of a sequence of quintic splines where the $j$-th spline for the $x$ component is represented as
        
        \begin{equation}
        \begin{split}
            & x(t) = \alpha^{x}_{j5}t^5 + \alpha^{x}_{j4}t^4 + \alpha^{x}_{j3}t^3 + \alpha^{x}_{j2}t^2 + \alpha^{x}_{j1}t + \alpha^{x}_{j0} \\
            & \qquad = [t^5 \quad t^4 \quad t^3 \quad t^2 \quad t \quad 1] \\
            & \qquad \quad \cdot [\alpha^{x}_{j5} \quad \alpha^{x}_{j4} \quad 
            \alpha^{x}_{j3} \quad \alpha^{x}_{j2} \quad \alpha^{x}_{j1} \quad \alpha^{x}_{j0}]^T \\
            & \qquad = \eta^T(t)\alpha^x_j.
        \end{split}
        \end{equation}
        
        The spline $j$ corresponds to a duration of $\Delta\, t_j$ such that $t\in[\overline{t},\, \overline{t}+\Delta\, t_j]$. Here
        $\overline{t}$ refers to the total elapsed time of the motion described by $j-1$ splines. We therefore
        express the CoM position, $\mathrm{p}_{CoM}\in\mathbb{R}^3$, as $\mathrm{p}_{CoM}=\mathbf{K}(t)\alpha_j$ where

        \begin{equation}
            \mathbf{K}(t)=\begin{bmatrix}
                 \eta^T(t) & 0 & 0\\
                0 & \eta^T(t) & 0 \\
                0 & 0 & \eta^T(t)
            \end{bmatrix},
        \end{equation}
        
        and $\alpha_j=[{\alpha_j^x}^T \quad {\alpha_j^y}^T \quad {\alpha_j^z}^T]^T$. The CoM velocity is thus represented as
        $\mathrm{\dot{p}}_{CoM}=\mathbf{\dot{K}}(t)\alpha_j$, 
        and the acceleration as $\mathrm{\ddot{p}}_{CoM}=\mathbf{\ddot{K}}(t)\alpha_j$.
        
        We parameterize the feet motions based on their contact state
        given by the vector $\mathrm{c}_F\in\left\{0, 0.5, 1\right\}^4$ where an open-contact is represented by 0 
        and close-contact (with zero foot velocity) by 1. Depending on the terrain, there remains a possibility of foot slippage i.e. a close-contact with
        non-zero foot velocity. This \textit{contact-uncertainty}
        state is therefore denoted by 0.5. For the purpose of motion planning, we only consider the open (0) and close (1) contact states.
        The contact-uncertainty state is instead utilized for adapting the joint PD control gains during motion
        tracking. This is detailed in Section~\ref{sec:motion_controller}.
        
        We define the swing phase, $\Phi_{o,i}$, for foot $i\in\{0,1,2,3\}$ to 
        correspond to the foot motion where $\mathrm{c}_{F,i}=0$.
        Conversely, the stance phase, $\Phi_{c,i}$, corresponds to the motion where $\mathrm{c}_{F,i}=1$. 
        Both the swing and stance phases can be described by the tuple
        \begin{equation}
        \Phi=\langle r_F^l, r_F^d, r_F^h, \mathrm{c}_F, t_F^l, t_F^d, v^l_F, v^d_F \rangle    
        \end{equation}
         where $r_F^l$, $r_F^d$ and $r_F^h$ refer to the
        vectors describing the feet lift-off, touch-down and clearance positions respectively. The foot clearance position
        represents the maximum $r_{F_z,i}$ during the given phase. For $\Phi_{o,i}$, $t_{F,i}^l$ is the time elapsed since previous
        lift-off and $t_{F,i}^d$ is the time until touch-down. For $\Phi_{c,i}$, $t_{F,i}^l$ refers to the time until lift-off and 
        $t_{F,i}^d$ is the time elapsed since previous touch-down. $v^l_F$ and $v^d_F$ refer to the
        feet lift-off and touch-down velocities respectively. Additionally, for $\Phi_{c,i}$, we consider $r_F^l = r_F^d = r_F^h$.

    \subsection{Reinforcement Learning}
        The problem of sequential decision making can be formulated in the framework of 
        a discrete time Markov decision process (MDP)~\cite{sutton1998introduction}. 
        An MDP is defined by the tuple - $\langle S, A, R, P, \mu \rangle$, where 
        $S$ represents a set of states, $A$ a set of actions,
        $R:S\times A\times S\rightarrow \mathbb{R}$ the reward function, 
        $P:S\times A\times S\rightarrow \left[0,1 \right]$ 
        the state transition probability, and $\mu$ the initial 
        state distribution. In the context of deep RL, as utilized in our work, an RL agent described in this MDP framework interacts with an environment
        based on a policy $\pi:S\rightarrow \mathcal{P}\left(A\right)$ which is modeled as a neural network. This policy can be defined as
        a function mapping states to probability
        distributions over actions such that $\pi\left(a|s\right)$ 
        denotes the probability of selecting action $a$ in state $s$. The objective of the RL agent is to then obtain a policy $\pi_{\theta}$ for parameters
        $\theta$ in order to maximize
        the expected cumulative discounted return,
        \begin{equation}
        J\left(\pi\right)\doteq\underset{\mathcal{T}\sim\pi_\theta}{\text{\textup{E}}}\left[\sum_{t=0}^{\infty}{{\gamma}^{t}R\left({s}_{t},{a}_{t},{s}_{t+1}\right)}\right]\text{,}
        \end{equation}
        where $\gamma\in\left[0,1\right)$ is the discount factor and $\mathcal{T}$ 
        denotes a trajectory dependent on $\pi_\theta$.
        
\section{Methodology}
We present a modular terrain-aware quadrupedal planning and control framework for locomotion
over uneven terrain. We propose a footstep planner which utilizes the proprioceptive and exteroceptive 
information to generate the desired feet touch-down positions, ${r_F^d}^\ast$, that are tracked during
the immediate swing phases. The footstep planner is sequentially 
evaluated after each of the previous four swing phases
are executed, completing a \textit{gait stride}. 
The desired footholds are thus updated after each gait stride at the instance when each
of the feet is in stance phase, $\mathrm{c}_{F}=\overrightarrow{1}_F$ where 
$\overrightarrow{1}_F\in\mathbb{R}^4$ is a vector of ones.
The dynamic gaits motion controller, which serves as a motion planner and whole-body controller,
is then employed to generate and track
the desired CoM and feet motion plans.
The whole-body tracking of motion plans occurs at \SI{400}{\hertz}, while 
the motion planning optimization is performed in parallel to the control thread and a
new optimization problem is run everytime a solution is found.

The domain adaptive tracker, also executed at
\SI{400}{\hertz}, is trained to introduce additive deviations to the 
feed-forward joint torques generated by the motion controller, $\tau_{j_{FF}}$. 
These additive deviations, $\delta\tau_{j_{FF}}$, are included
in the impedance controller shown in eq.~\ref{eq:impedance_controller} and account
for discrepancies in the modeled and actual system dynamics. Finally, the recovery
controller, running at \SI{400}{\hertz}, is activated to perform joint position control 
when a robot state is encountered such
that the motion controller becomes unstable and is unable to execute a state recovery behavior.

We employ this modular control architecture for following reasons:
\begin{enumerate}
    \item The individual trained modules can be precisely tuned to adapt to different tasks. 
    For example, transferring the framework for use on a quadruped with significantly larger mass
    would only require adapting either the motion controller or the domain adaptive tracker. This, in fact,
    also allows us to perform a zero-shot transfer from the simulated ANYmal B training domain to the physical ANYmal C.
    \item In our preliminary experiments, we observed that training a terrain-aware locomotion
    policy which outputs desired joint position commands required comprehensive tuning of the 
    RL environment parameters, without which, the network often failed to learn, resulting 
    in a behavior where the robot would not respond to velocity commands, instead,
    would stand in place. This observation significantly contributed towards our modular 
    control architecture design. A modular framework enabled us to perform easier tuning
    of individual components. The contributions of each of the modules are detailed in Section~\ref{sec:results_and_discussion}.
    \item Distributing the planning and control tasks between the RL footstep planner
    and the model-based motion controller enabled us to obtain a footstep planner which was
    able to generalize to wide range of terrains in less than \SI{32}{\hour} of total training.
    In contrast, the gait planner of DeepGait~\cite{tsounis2020deepgait} required \SI{82}{\hour}
    of training time for each of the terrains. Moreover, the gait controller of DeepGait
    required further \SI{116}{\hour} of training time for each of the uneven terrains,
    and an additional \SI{58}{\hour} of training for locomotion on flat-ground, totaling 
    the training time to approximately \SI{454}{\hour} for locomotion over three different
    terrains: Flat-World, Temple-Ascent, and Random-Stairs (see DeepGait~\cite{tsounis2020deepgait}).
\end{enumerate}

An overview of the control framework is presented in Fig.~\ref{fig:overview}. This
section details upon each of the modules. We also describe the MDPs for the footstep planner, 
domain adaptive tracker and the recovery controller in this section. The detailed RL training 
setups are presented in Section~\ref{sec:training}.

\subsection{Elevation Map Encoding}
            \begin{figure}
                    \begin{subfigure}{.24\textwidth}
                    \centering
                        \includegraphics[width=.99\linewidth]{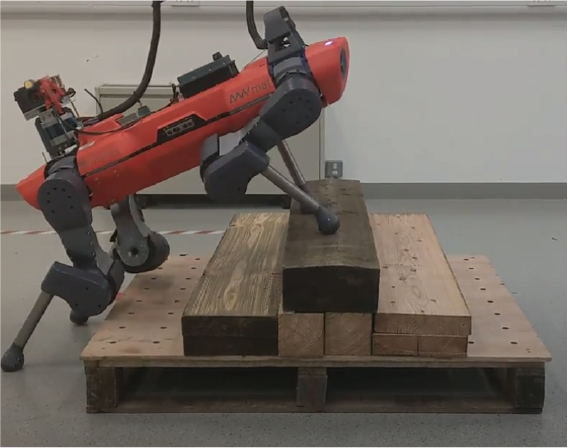}
                    \end{subfigure}
                     \begin{subfigure}{.24\textwidth}
                    \centering
                        \includegraphics[width=.99\linewidth]{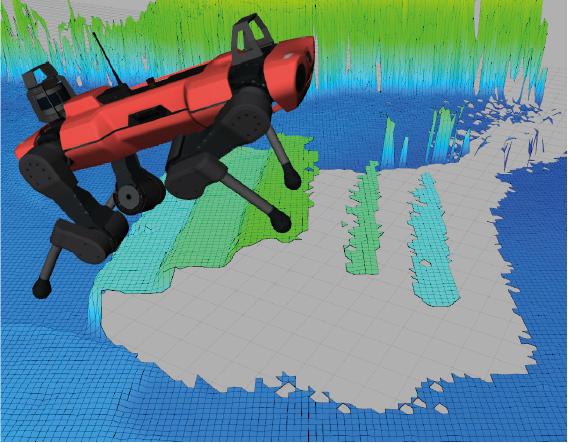}
                    \end{subfigure} \\
                    \begin{subfigure}{.24\textwidth}
                    \centering
                        \includegraphics[width=.99\linewidth]{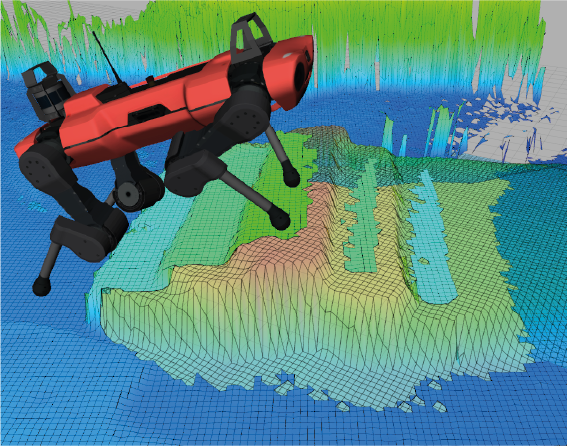}
                    \end{subfigure}
                    \begin{subfigure}{.24\textwidth}
                    \centering
                        \includegraphics[width=.99\linewidth]{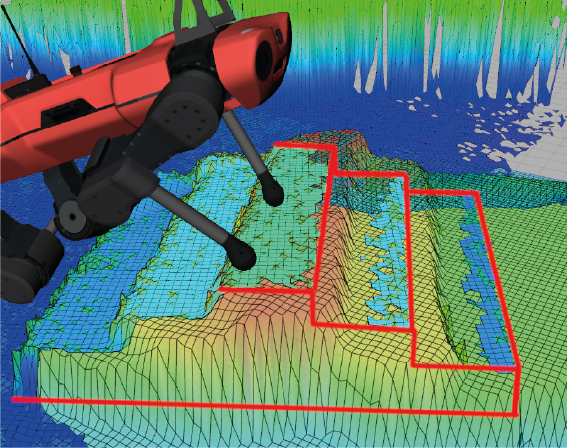}
                    \end{subfigure}    
                \caption{\textbf{top:} Ascending stairs with ANYmal C (left) and the corresponding raw elevation map (right).
                The unfilled regions in the raw elevation map represent non-observable terrain information which
                could not be captured by the depth camera on the robot due to occlusions. \textbf{bottom:} Filtered
                elevation map obtained using nearest-neighbors interpolation (left) and the 
                terrain edges (right).}                              
                \label{fig:elevation_map:occlusions}
            \vspace{-12pt}
            \end{figure}

            We extract robocentric terrain elevation, $\mathbf{M}_{H}$, using the \textit{elevation mapping}
            framework presented in~\cite{Fankhauser2018ProbabilisticTerrainMapping} and embed it
            to obtain a latent representation, $\mathrm{Z}_{\mathbf{M}_{H}}\in\mathbb{R}^{96}$. This representation
            is generated by an encoding network, $E_{\mathbf{M}_{H}}$ which is trained using a denoising convolutional
            auto-encoding strategy. Our motivation for performing this embedding (including the encoding approach) 
            is based on the following reasons:
            \begin{enumerate}
                \item Three of the modules constituting our RLOC framework utilize exteroceptive terrain
                information. Pre-processing the elevation map to extract terrain features eliminates
                repeated processing of the elevation map by each of the modules limiting further computational delays.
                \item The denoising property of $E_{\mathbf{M}_{H}}$ is relevant to perform a sim-to-real
                transfer. The elevation map extracted on the physical system is often noisy and contains
                occlusions that cannot be precisely modeled. An example of this is shown in 
                Fig.~\ref{fig:elevation_map:occlusions}. This has further been studied in~\cite{miki2022learning}.
                A nearest-neighbors interpolation technique
                is employed to obtain a filtered elevation map which often results in
                distortions (e.g. the edges of the steps are blurred and shifted). The denoising auto-encoding strategy 
                enables extraction of the most relevant terrain features like
                edges and slopes even with distorted terrain information.
                \item A pre-trained terrain feature extractor implies that, during training of RL policies,
                the RL agent does not need to learn the parameters for the terrain feature extraction block thereby
                making the training faster while also adapting to a pre-trained encoder which can be shared with
                multiple modules.
                \item Even though different strategies such as fast Fourier transform (FFT) can be 
                employed to compress the elevation map, our work focuses on utilizing
                deep learning strategies for \textit{feature extraction}. This enables us to adapt the previously
                trained network to a diverse class of terrains by retraining the encoder. Additionally, strategies
                such as transfer learning can be utilized to reduce training complexity. The 
                feature extractor can also be retrained along with an RL policy to adapt to different tasks not introduced during prior training.
            \end{enumerate}

            \begin{figure*}
                \centering
    \includegraphics[width=0.92\textwidth]{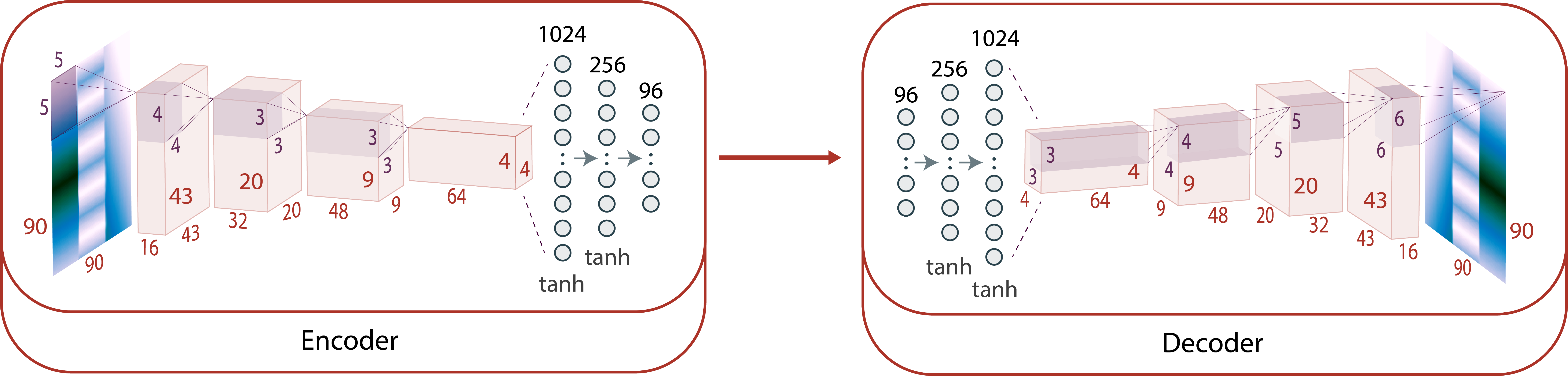}
                \caption{Network architecture of the denoising convolutional autoencoder used to 
                generate an embedded representation of the local elevation map. We use the \textit{leaky ReLU} activation function for the convolutional layers and the \textit{tan hyperbolic} activation function for the dense layers.}
                \label{fig:autoencoder}
            \vspace{-10pt}
            \end{figure*}

            \begin{figure}
                \centering
                \includegraphics[width=0.48\textwidth]{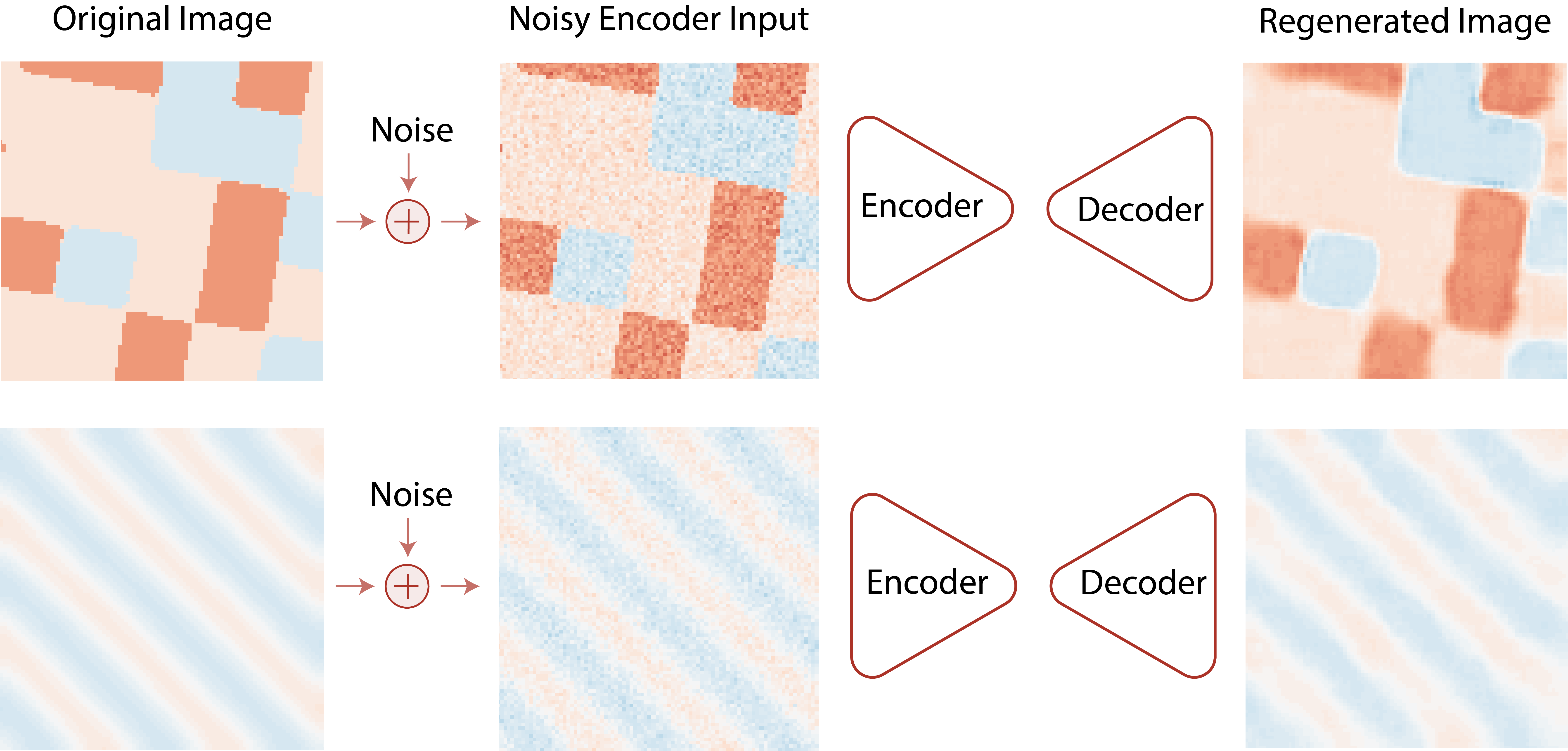}
                \caption{Examples of regeneration of noisy images using our trained autoencoder - \textbf{top:} a patch from the bricks terrain and \textbf{bottom:} a patch from the sine wave terrain.}
                \label{fig:autoencoder_examples}
            \vspace{-10pt}
            \end{figure}

            The architecture of the denoising convolutional auto-encoder is presented in
            Fig.~\ref{fig:autoencoder}. Note that, while extracting
            the elevation map, we clip $\mathbf{M}_H$ in the range $[-2, 2]\,$\si{\meter}. For network training and 
            evaluation stability,
            we further normalize $\mathbf{M}_H$ by performing a scalar
            division by 2 to obtain $\mathbf{M}_H^n$ with elements in the range $[-1, 1]$. 
            As part of training, our objective is then to learn the weights 
            of the convolutional encoder $E_{\mathbf{M}_{H}}$, such that
            \begin{equation}
                \mathrm{Z}_{\mathbf{M}_H} = E_{\mathbf{M}_{H}}(\mathbf{M}_H^n + \mathbf{N}_{\mathbf{M}}(\mathbf{M}_H^n))
            \end{equation}
            where $\mathbf{N}_{\mathbf{M}}$ represents the artificial noise introduced in $\mathbf{M}_H^n$, and
            \begin{equation}
                {\mathbf{\hat{M}}}_H^n = D_{\mathbf{M}_{H}}(\mathrm{Z}_{\mathbf{M}_H})
            \end{equation}
            where $D_{\mathbf{M}_{H}}$ is the decoder network which maps the latent embedding of $\mathbf{M}_{H}^n$
            to its approximate reconstruction ${\mathbf{\hat{M}}}_H^n$. Both $E_{\mathbf{M}_{H}}$ and 
            $D_{\mathbf{M}_{H}}$
            are trained together using supervised learning to minimize the reconstruction loss
            \begin{equation}
                \mathcal{L}_{\mathbf{M}_{H}} = ({\mathbf{M}}_H^n - {\mathbf{\hat{M}}}_H^n)^2 \,.
            \end{equation}
            
            The artificial noise, 
            $\mathbf{N}_{\mathbf{M}}:\mathbb{R}^{91\times91}\rightarrow\mathbb{R}^{91\times91}$, is expressed as
            \begin{equation}
            \label{eq:encoder_input_noise}
            \begin{split}
                & \mathbf{N}_{\mathbf{M}}(\mathbf{M}) = n_\mathcal{N}\mathcal{N}_{91\times91}(0,\,1) \\
                & \qquad\qquad +\, (\mathcal{I}({\mathbf{M}}, p_\mathcal{I}) - \mathbf{M}) \\
                & \qquad\qquad +\, n_\mathcal{C}(\mathcal{C}(\mathbf{M}, k_b) - \mathbf{M})
            \end{split}
            \end{equation}
            where $\mathbf{M}\in\mathbb{R}^{91\times91}$ is a matrix similar to $\mathbf{M}_H^n$, $\mathcal{N}$ 
            is the additive Gaussian noise, $\mathcal{I}$ is the impulse noise also know as salt-and-pepper noise,
            $p_\mathcal{I}\sim\mathcal{U}(0, 0.05)$
            is the probability of introduced impulses sampled uniformly in the range $[0, 0.05]$, 
            $\mathcal{C}$ is the 2D 
            convolution function, and $k_b\in\mathbb{R}^{3\times3}$ is the convolutional blurring kernel written as
            \begin{equation}
                    k_{b} = \frac{1}{9}\times
                        \begin{bmatrix}
                            1 & 1 & 1\\
                            1 & 1 & 1\\
                            1 & 1 & 1\\
                        \end{bmatrix}\, .
            \end{equation} $n_\mathcal{N}\sim\mathcal{U}(0, 0.05)$ and
            $n_\mathcal{C}\sim\mathcal{U}(0, 1)$ are the coefficients for weighting the noise functions.

            To train the denoising convolutional autoencoder, we use our terrain generator tool, 
            described in Section~\ref{section:terrain_generation}, 
            to generate
            $20\times20\,$\si{\metre\squared} terrains, where each terrain is represented as
            $\mathbf{T}\in\mathbb{R}^{1001\times1001}$.
            These terrains contain objects such as stairs,
            wave, bricks, planks and unstructured ground.
            We slice a $91\times91$ matrix from $\mathbf{T}$ at randomly selected positions and further
            use data-augmentation techniques such as
            rotation, mirroring and modifying contrast during training of the autoencoder online. This augmentation
            is done in parallel to the main training session in a separate thread. Figure~\ref{fig:autoencoder_examples} 
            shows examples of regeneration of noisy terrain inputs using the trained autoencoder.  

    \subsection{Motion Controller}
    \label{sec:motion_controller}
    
        \textit{Rudin et al.} propose an architecture which allows for direct mapping of proprioceptive and exteroceptive state information
        to desired joint positions~\cite{rudin2022learning}. This, however, often requires exhaustive reward tuning to achieve a desired locomotion behavior~\cite{hwangbo2019learning, gangapurwala2020guided}. As an alternative, \cite{miki2022learning} utilizes central
        pattern generators (CPGs) to achieve an underlying oscillatory behavior for each of the quadrupedal limbs.
        This architecture enables online adaptation of the locomotion gait while also making the RL training less demanding~\cite{iscen2018policies}.
        In this work, we are driven by the motivation that the proposed framework should be robust to variations in system kinematics and dynamics. For this, we utilize a motion controller to generate and track base and feet motion plans for traversal over uneven terrain. The desired feet positions are instead generated by the RL 
        footstep planner described in Section~\ref{section:methodology:footstep_planner}. This eliminates the
        need for the RL agent to perform motion control thereby simplifying training. Additionally, the model-based motion 
        control architecture 
        ensures that safety-critical constraints such as joint kinematic limits and peak joint torques are
        still enforced by the low-level controller. The system-adaptive property is then induced by the
        domain adaptive tracker presented in Section~\ref{section:methodology:domain_adaptive_tracker}.
        
        In this work, we extend the \textit{dynamic gaits} motion controller of \textit{Bellicoso et al.}~\cite{bellicoso2018dynamic}
        to generate CoM and feet motion
        plans for traversal over uneven terrain. 
        The CoM motion plan is generated by solving a nonlinear optimization problem to track the velocity command, $\mathbf{c}^\ast$,
        for a time horizon $t_H^c$. This time horizon
        corresponds to the period of
        the active locomotion gait.
        We introduce a control frame $C$ such that
        $r_C=r_B$, $\mathbf{R}_{C_z}=\mathbf{R}_{B_z}$, and $\mathbf{e}_z^C$ aligns with the normal of the estimated terrain 
        plane 
        $\mathbf{e}_z^T$. The velocity command is therefore expressed in the control
        frame as
        \begin{equation}
            \mathbf{c}^\ast = [\mathrm{v}_x^\ast\mathbf{e}_x^C \quad \mathrm{v}_y^\ast\mathbf{e}_y^C \quad
            \mathrm{\omega}_z^\ast\mathbf{e}_z^C]^T
        \end{equation} where $\mathrm{v}_x^\ast$ and $\mathrm{v}_y^\ast$
        are the desired heading and lateral velocities expressed along $\mathbf{e}_x^C$ and $\mathbf{e}_y^C$ 
        respectively, and $\mathrm{\omega}_z^\ast$ is the desired yaw rate around the vertical axis $\mathbf{e}_z^C$.

        The motion controller employs a quadratic programming (QP) based
        foothold planning module to generate the desired feet touch down positions, ${r_F^d}^\ast$.
        This is replaced by the RL footstep planner in our proposed framework. The vertical component of the
        desired touch down positions, ${r^d_{F_{z}}}^\ast$,
        is directly extracted from the terrain plane estimate for each swing phase. The contact
        scheduler module of the motion controller manages the feet contact events for a full
        stride of
        an active gait. It generates the lift-off,
        $t_F^l$, and touch-down, $t_F^d$, timings for the feet swing and stance phases. The $\Phi_{o}$ and 
        $\Phi_{c}$ thus obtained is used to generate feet motion plans to track ${r_F^d}^\ast$.
        
        A detailed description of the modifications introduced in the original motion controller is provided in
        the Appendix.
        
         The generated reference motion plans are tracked using a whole-body controller~\cite{bellicoso2017dynamic}
         based on a hierarchical QP optimization framework to optimize for the generalized accelerations $\mathrm{\dot{u}}$ and feet contact forces to generate the feed-forward torques $\tau_{j_{FF}}$. The
         impedance control torques are then computed by extracting the desired joint positions, $\mathrm{q}^{\ast}_j$,
         and joint velocities, $\mathrm{\dot{q}}^{\ast}_j$, from the feet motion plans using inverse kinematics.
        
        The work of \textit{Jenelten et al.} on locomotion over slippery surfaces~\cite{Jenelten2019dynamicL} 
         proposes the use of impedance gains depending on the feet contact-states. For a foot $i$, the gains are 
         selected according to the rule \begin{equation}
             K_{p,d}^{\mathrm{c}_{F,i}=0.5} >\!> K_{p,d}^{\mathrm{c}_{F,i}=1} > K_{p,d}^{\mathrm{c}_{F,i}=0}
         \end{equation} such that, in the contact-uncertainty state $\mathrm{c}_{F,i}=0.5$, 
         the high gains used for joints in the leg 
         containing the foot $i$ prevent further leg slippage due to increased stiffness. For detailed description, 
         we defer the reader to the original work. We use this approach of adaptive impedance for whole-body tracking
         of the reference motion plans in RLOC.
    
        \subsection{Footstep Planner}
        \label{section:methodology:footstep_planner}
            Approximating the perceptive footstep planning policy
    using a neural network implies that we can directly map the exteroceptive and proprioceptive robot state information to
    desired feet positions. This rids us of the need for computationally exhaustive task of combinatorial search among a 
    large space of feasible footstep plans. Additionally, as presented in Section~\ref{sec:results_footstep_planning},
    training the footstep planner using deep RL enables natural emergence of planning behavior learned through
    interactions with the environment of operation. This offers better performance for traversal over unstructured 
    terrains compared to a baseline sampling-based perceptive footstep planner.
        
            Computing new set of desired feet positions after each gait stride, specifically after
            each of the previous four feet swing phases have been executed, allows us to
            formulate the problem of footstep planning as a sequential MDP. In particular,
            we represent the problem as an infinite-state space and infinite-action space MDP.
            
            \subsubsection{State and Action}   
            We define the MDP state as a tuple $\mathbf{s}_f:=\langle s_R, s_v, s_j, s_c, s_M \rangle$, and
            the action $\mathbf{a}_f:=\langle {}_{H}r_{H{F_{xy}}}^\ast \rangle$. The state $\mathbf{s}_f\in\mathbb{R}^{204}$
            comprises terms representing 
            robot base orientation $s_R\in\mathbb{R}^3$, base linear and angular velocities
            $s_v\in\mathbb{R}^6$, sparse representation of history of joint positions and velocities 
            $s_j\in\mathbb{R}^{96}$, desired base velocity $s_c\in\mathbb{R}^3$, and an embedded representation
            of the robocentric elevation map $s_M\in\mathbb{R}^{96}$. The exact
            definitions of the footstep planner (FP) state terms are provided in Table~\ref{tab:fp_state_definitions}.
            The action $\mathbf{a}_f\in\mathbb{R}^8$ 
            comprises the desired foothold locations, ${}_{H}r_{H{F_{xy}}}^\ast$, for each of the four feet represented in the
            horizontal frame. The $z$ components of the desired feet positions, ${}_{H}r_{H{F_{z}}}^\ast$, 
            are directly extracted from $\mathbf{M}_H$.
            Therefore, the desired feet positions, ${}_{H}r_{HF}^\ast$, measured and expressed in horizontal frame
            can be transformed into the world frame representation using the transformation
            \begin{equation}
                r_{F,i}^{FP}=(\mathbf{R}_H)\;({}_{H}r_{HF,i}^\ast) \ + \ r_B 
            \end{equation} for foot $i$. The feet positions generated by the footstep planner, $r_{F}^{FP}$, are then forwarded to
            the dynamic gaits motion controller.
        
            \begin{table}[h!]
                \centering
                \caption{State term definitions for the footstep planner MDP. Here, $t$ refers to the current measurement and the 
                duration between $t-1$ and $t$ corresponds to \SI{2.5}{\milli\second}.}
                \begin{tabular}{|c|}
                    \hline
                     FP State \rule{0pt}{2.6ex} \rule[-0.9ex]{0pt}{0pt} \\
                     \hline
                     $s_R=\mathbf{e}_z^B$ \rule{0pt}{2.6ex} \\
                     $s_v=[(\mathbf{R}_B^T\mathrm{v}_B)^T \quad (\mathbf{R}_B^T\mathrm{\omega}_B)^T]^T $ \rule{0pt}{2.6ex} \\
                     $s_j=[{\mathrm{q}_{j}}_{t}^T \ \ {\mathrm{q}_{j}}_{t-5}^T \  \ {\mathrm{q}_{j}}_{t-10}^T \ \ {\mathrm{q}_{j}}_{t-50}^T \ \ 
                     {\dot{\mathrm{q}}_{j}}{}_{t}^T \ \ {\dot{\mathrm{q}}_{j}}{}_{t-5}^T \ \ {\dot{\mathrm{q}}_{j}}{}_{t-10}^T \ \ {\dot{\mathrm{q}}_{j}}{}_{t-50}^T]^T$ \rule{0pt}{2.6ex} \\
                     $s_c=\mathbf{c}^\ast$ \rule{0pt}{2.6ex} \\
                    $s_M=\mathrm{Z}_{\mathbf{M}_H}$ \rule{0pt}{2.6ex} \rule[-0.9ex]{0pt}{0pt} \\
                     \hline 
                \end{tabular}
                \label{tab:fp_state_definitions}
            \end{table}
        
            Note that, we introduce the history of joint states in the state space to infer the 
            active locomotion gait. The different gaits result
            in different \textit{stability margin}~\cite{Orsolino2020,orsolinorapid}, a metric introduced in the reward function to
            assess the dynamic stability of the robot.
            This implies that the desired feet positions generated by the footstep planner depend
            on the gait in use. The state space is thus required to contain
            information relating to the robot locomotion gait.
        
            \subsubsection{Reward Function}
            The reward function in the MDP formulation
            encourages stable dynamic locomotion over uneven terrain in order to track 
            a desired base velocity command while minimizing energy consumption. During execution of 
            the sequential footstep planning MDP in an episodic framework, the planning state
            transition, ${\mathbf{P}_f}_n := \langle {\mathbf{s}_f}_n, {\mathbf{a}_f}_n, {\mathbf{s}_f}_{n+1} 
            \rangle$, is only observed 
            after the motion plans for each of the feet are executed. The
            tracking of these motion plans is instead performed at each control 
            step executed at \SI{400}{\hertz}. Therefore, each
            $\mathbf{P}_f$ in the footstep planning MDP comprises several control steps. The
            control behavior observed within these state transitions is still relevant to assess the performance
            of the footstep planner especially pertaining to velocity tracking and foot slippage. In this regard,
            our reward function includes a \textit{running} reward term, $\mathbf{r}_F^\mathrm{r}\in\mathbb{R}$, accumulated 
            over the control
            horizon for each $\mathbf{P}_f$, and a \textit{final} reward term $\mathbf{r}_F^\mathrm{f}\in\mathbb{R}$,
            computed after a state transition is observed. The reward function can then be expressed as
            \begin{equation}
                \mathbf{r}_F = \mathbf{r}_F^\mathrm{f} + \frac{1}{(n+1)_t - (n_t+1)}
                \left(\mathlarger{\mathlarger{\sum}}_{t=n_t+1}^{(n+1)_t}{\mathbf{r}_F^\mathrm{r}}_t\right)
            \end{equation} where $n_t$ corresponds to the control step
            at which state $\mathbf{s}_{f_n}$ is observed. The
            duration between $t=n_t+1$ and $t=(n+1)_t$ is not fixed and depends upon the touch-down and lift-off
            timings. We express the running reward at control step $t$ as
            \begin{equation}
            \label{eq:running_reward_fp}
                {\mathbf{r}_F^\mathrm{r}}_t=8{\mathbf{r}_v}_t - 0.016d_f{\mathbf{r}_{\tau}}_t - 2.5d_f{\mathbf{r}_{\mu}}_t + 4{\mathbf{r}_{m}}_t
            \end{equation} comprising the velocity tracking reward $\mathbf{r}_v$, control joint torque penalty
            $\mathbf{r}_{\tau}$, feet slippage penalty $\mathbf{r}_{\mu}$, and the stability margin reward $\mathbf{r}_{m}$. The weighting factor $d_f\in[0, 1]$ corresponds to the training curriculum
            and is increased as training progresses to fine tune the behavior of the RL agent. We update
            the curriculum factor after each policy iteration by the rule $d_{f_{k+1}}\leftarrow{d}_{f_{k}}^{0.9986}$
            where $k$ refers to the policy iterations and $d_{f_{0}}=0.02$.
            The final reward is represented as
            \begin{equation}
                \label{eq:final_reward_fp}
                \mathbf{r}_F^\mathrm{f} = -2d_f\mathbf{r}_n - 1d_f\mathbf{r}_e - 2.5d_f\mathbf{r}_\mu + 
                4\mathbf{r}_m + 2\mathbf{r}_h
            \end{equation} comprising the penalty for deviation from nominal footholds $\mathbf{r}_n$, 
            reward for distance from terrain edges $\mathbf{r}_e$, feet slippage penalty $\mathbf{r}_\mu$, 
            the robot stability margin reward $\mathbf{r}_m$ based on~\cite{orsolinorapid},
            and the feet height reward relating to the closeness
            to the nominal feet height $\mathbf{r}_h$. The definitions for the reward terms are presented in
            Table~\ref{tab:fp_reward_definitions}.

            \begin{table}[h!]
                \centering
                \caption{Reward term definitions for the footstep planner MDP. Here $K$ refers to the
        logistic kernel defined as ${K(x):=({{e^{x}+2+e^{-x}}})^{-1}}$. $\mathcal{S}_m$ is the function
        mapping robot state information to the robot stability margin, $N_F$ represents the feet
        contact normals, $\protect\overrightarrow{F}_B$ represents the external force on base,
        $\protect\overrightarrow{\tau}_B$ is
        the external torque applied on base, $\mathbf{T}_\mu$ is the terrain friction coefficient, and
        $\mathbf{C}_{\mathbf{M}}$ maps the 
        foot position to a cost describing the distance of the foot from the terrain edges. The functions 
        $\mathcal{S}_m$ and $\mathbf{C}_{\mathbf{M}}$ are detailed in the Appendix. 
        ${{}_{H}r^n_{HF_{z}}}$
        is the $z$ component of the nominal feet positions.}
                \begin{tabular}{|c|}
                    \hline
                     FP Rewards \rule{0pt}{2.6ex} \rule[-0.9ex]{0pt}{0pt} \\
                     \hline
                     $\mathbf{r}_v = K\left(5\Vert \mathbf{c}^\ast - \mathbf{c} \Vert^2 \right)$ \rule{0pt}{2.6ex} \\
                     $\mathbf{r}_\tau = \Vert \tau_j \Vert^2$ \rule{0pt}{2.6ex} \\
                     $\mathbf{r}_{\mu} = \sum{\Vert \dot{r}_{F,i} \Vert}, 
                        \ \ \forall{i},\mathrm{c}_{F,i}=1,i\in\{0,1,2,3\}$ \rule{0pt}{2.6ex} \\
                     $\mathbf{r}_m = \mathcal{S}_m(\mathrm{q}, \mathrm{{u}}, \mathrm{\dot{u}}, \mathrm{r}_F, \mathrm{c}_F, N_F,         \overrightarrow{F}_B, \overrightarrow{\tau}_B, \mathbf{T}_\mu)$ \rule{0pt}{2.6ex} \\
                     $\mathbf{r}_n = \Vert {{}_{H}r_{HF_{xy}}} - {{}_{H}r^n_{HF_{xy}}} \Vert^2$ \rule{0pt}{2.6ex} \\
                     $\mathbf{r}_e = \sum{{\mathbf{C}_{\mathbf{M}}}(r_{F_{xy},i})}, \ \ \forall{i}$ \rule{0pt}{2.6ex} 
                     \\
                    $\mathbf{r}_h= K(8\Vert {{}_{H}r^\ast_{HF_{z}}} - {{}_{H}r^n_{HF_{z}}} \Vert^2)$ \rule{0pt}{2.6ex} \rule[-2.0ex]{0pt}{0pt} \\
                     \hline 
                \end{tabular}
                \label{tab:fp_reward_definitions}
            \end{table}

        \subsubsection{Episode Termination Criteria}
            We execute the footstep planning MDP in an episodic framework employing
            the following termination criteria while training the footstep planner:
            \begin{itemize}
                \item Angle between the terrain normal and base vertical axis exceeds a threshold, $|\mathrm{cos}^{-1}(\mathbf{e}_z^{B}\cdot\mathbf{e}_z^{C})| > \pi/4$
                \item Any of the robot links except for the feet collide with the terrain.
                \item Robot self-collision
            \end{itemize} Upon termination,
            the robot state is reset and a new episode is executed.
            
        \subsubsection{Policy Parameterization}
            We parameterize the footstep planning policy as a bounded Gaussian distribution
            ${\pi_{\theta_f}}(\mathbf{a} | 
            \mathbf{s}_f):=\mathrm{tanh}(\mathcal{N}(\mathbf{a} | {\mathbf{\mu}_{\theta_f}}(\mathbf{s}_f), 
            {\mathbf{\sigma}_{\theta_f}}(\mathbf{s}_f)))$ where both the mean 
            ${\mathbf{\mu}_{\theta_f}}(\mathbf{s}_f)$
            and the log standard deviations $\mathrm{log}({\mathbf{\sigma}_{\theta_f}}(\mathbf{s}_f))$
            are state-dependent
            outputs of a neural network. The footstep planning network architecture representing the function mapping 
            state to the action mean is shown in Fig.~\ref{fig:training}. The log standard deviation is 
            extracted using a linear mapping
            of the latent vector obtained as an output of the last hidden layer. This state-dependent 
            stochasticity is used to encourage exploration during training. We utilize
            a custom implementation of soft actor critic (SAC)~\cite{haarnoja2018soft}, heavily based 
            on \textit{RLlib}~\cite{liang2018rllib}, to train the footstep planning policy ${\pi_{\theta_f}}$
            by performing entropy-regularized reinforcement learning. For a slow environment setup
            of footstep planning, which includes prediction of desired footholds and further includes tracking of these feet positions using the 
            motion controller, SAC offers better sample efficiency in comparison to the widely used on-policy 
            alternatives,
            trust region policy optimization (TRPO)~\cite{schulman2015trust} and proximal policy optimization 
            (PPO)~\cite{schulman2017proximal},
            and has further been demonstrated to perform efficiently on quadrupedal locomotion 
            tasks~\cite{haarnoja2018softapp}. The training is detailed in Section~\ref{section:footstep_planner_training_eval}.
        
        \subsection{Domain Adaptive Tracker}
        \label{section:methodology:domain_adaptive_tracker}
        The model-based characteristics of the dynamic gaits motion controller implies that, in order to exhibit robust motion tracking, significant tuning of the controller parameters
        is required. This is especially relevant for computing $\tau_{j_{FF}}$ (see eq.~\ref{eq:impedance_controller}) which is written as
        \begin{equation}
            \tau_{j_{FF}} = \mathrm{M}_j{\mathrm{\dot{u}}}^\ast + \mathrm{h}_j - \mathrm{J}^T_{sj}\lambda^\ast
        \end{equation} where $\mathrm{M}_j$ represents the mass matrix relative to the joints, $\mathrm{h}_j$ comprises Coriolis, centrifugal and gravity terms, $\mathrm{J}_{sj}$ corresponds to
        the support Jacobian, ${\mathrm{\dot{u}}}^\ast$ is the desired generalized accelerations, and $\lambda^\ast$ represent the desired contact forces. This is detailed in~\cite{bellicoso2017dynamic}.
        Executing such a controller on a system where the robot dynamics differ from the modeled dynamics will result in failure to perform motion plan tracking. In this regard
        we introduce a domain adaptive tracker to generate corrective joint torques, $\delta\tau_j$, for impedance control which can then be rewritten as
        \begin{equation}
        \label{eq:dat_impedance_controller}
            \tau_j^c = K_p(\mathrm{q}^{\ast}_j-\mathrm{q}_j) + K_d(\mathrm{\dot{q}}^{\ast}_j-\mathrm{\dot{q}}_j) + \tau_{j_{FF}} + \delta\tau_j \ .
        \end{equation}
        This reformulation offers us the ability to deploy the motion controller on systems with different dynamics without the need for retuning the control parameters. The experimental evaluation presented in Section~\ref{sec:results_dat} includes ablation studies which
        support this statement. This is further illustrated in Fig.~\ref{fig:adaptive} and Fig.~\ref{fig:adaptive_torque_scaling}.
        Our motivation for introducing the domain adaptive tracker is therefore two-fold:
        \begin{enumerate}
            \item When deployed on a physical system, RLOC can be made robust to system disturbances, encouraging accurate motion plan tracking even with change in dynamic
            properties, such as added mass, without the need for explicit modeling of system inaccuracies.
            \item Introducing the domain adaptive tracker in an RL environment setup employing a model-based motion controller allows us to perform dynamics randomization without
            retuning the controller. This enables us to utilize the domain adaptive tracker for training the footstep planning policy with wide domain of kinematic and dynamics
            properties eventually letting us perform a zero-shot transfer of the footstep planner trained with ANYmal B to ANYmal C. 
        \end{enumerate}
        
        \subsubsection{State and Action}
        We formulate the problem of domain adaptive tracking in the framework of MDP, such that, given the robot state and the history of tracking errors, the domain adaptive tracker (DAT) implicitly models the
        deviation in system dynamics from the modeled dynamics
        to generate corrective joint torques at each control step. We define
        the DAT state as 
        \begin{equation}
            \mathbf{s}_d:=\langle s_r, s_{\delta{r}}, s_{\mathbf{R}}, s_{\mathbf{R}^\ast},
            s_\mathbf{v}, s_{\delta\mathbf{v}}, s_{\mathbf{j}}, s_F, s_\tau, s_M, s_b, 
            s_{\mathbf{b}} \rangle
        \end{equation} where $\mathbf{s}_d\in\mathbb{R}^{330}$ comprises vector representing the 
        history of robot base position $s_r\in\mathbb{R}^6$, 
        history of base position tracking error $s_{\delta{r}}\in\mathbb{R}^9$, history
        of robot orientation $s_{\mathbf{R}}\in\mathbb{R}^9$, 
        history of desired orientation $s_{\mathbf{R}^\ast}\in\mathbb{R}^9$, history of robot velocities 
        $s_\mathbf{v}\in\mathbb{R}^{18}$, history of velocity tracking errors 
        $s_{\delta\mathbf{v}}\in\mathbb{R}^{18}$,
        history of joint state $s_{\mathbf{j}}\in\mathbb{R}^{72}$, history of desired feet positions and 
        velocities $s_F\in\mathbb{R}^{72}$, previous corrective torques $s_\tau\in\mathbb{R}^{12}$, 
        embedded elevation map $s_M\in\mathbb{R}^{96}$, robot base position relative to the 
        elevation map update frame $s_b\in\mathbb{R}^3$, and robot orientation in the elevation map
        update frame $s_{\mathbf{b}}\in\mathbb{R}^{6}$. It is important
        to note that, unlike in the case of footstep planning where each call to the policy
        occurs after each gait stride, the
        domain adaptive tracking policy is executed at \SI{400}{\hertz}. Since the 
        perception sensors on the robot are significantly slower, the elevation map on the physical
        robot cannot be updated at \SI{400}{\hertz}. In order to account for the
        asynchronous updates of the exteroceptive and proprioceptive feedback, we store
        the reference frame of the robot's base during each update of the elevation map and 
        include the current state of the robot in this frame in the state tuple.
        The \textit{elevation map update frame} is represented as $B_{\mathbf{M}_H}$. The definitions 
        of the state terms are provided in Table~\ref{tab:dat_state_definitions}.

            \begin{table}[h!]
                \centering
                \caption{State term definitions for the domain adaptive
                tracking MDP. Here, subscript notation $t$ on a frame indicates the state of the
                frame at control step $t$, for example, $B_{t-4}$ represents the state of frame $B$ at 
                control step $t-4$. If no subscript exists, the frame is considered to be in the current
                state $t$. ${}_{B}r^\ast_{B{B_t}}$ is the desired base position at control step 
                $t$ expressed in the current measured 
                base frame, $\mathbf{\hat{e}}_z^{B}$ the desired 
                orientation, $r^\ast_F$ the desired feet positions, $\dot{r}^\ast_F$ the desired 
                feet velocity, $\mathrm{v}^\ast_B$ desired linear velocity and $\mathrm{\omega}^\ast_B$ is the
                desired angular velocity. The desired robot state at control step $t$ is extracted from
                the motion plans generated by dynamic gaits.}
                \begin{tabular}{|c|}
                    \hline
                     DAT State \rule{0pt}{2.6ex} \rule[-0.9ex]{0pt}{0pt} \\
                     \hline
                     $s_r= [ {}_{B}r_{B{B_{t-4}}}^T \ \ {}_{B}r_{B{B_{t-8}}}^T ]^T$ \rule{0pt}{2.6ex} \\
                    
                     $s_{\delta{r}} = [ {{}_{B}r^\ast_{B{B_{t}}}}^T \ \ ({}_{B}r^\ast_{B{B_{t-4}}} - 
                     {}_{B}r_{B{B_{t-4}}})^T \ \ ({}_{B}r^\ast_{B{B_{t-8}}} - {}_{B}r_{B{B_{t-8}}})^T ] $ 
                     \rule{0pt}{2.6ex} \\
                    
                     $s_{\mathbf{R}} = [ (\mathbf{e}_z^{B})^T \ \ (\mathbf{e}_z^{B_{t - 4}})^T \ \ 
                     (\mathbf{e}_z^{B_{t - 8}})^T ] ^ T$ 
                     \rule{0pt}{2.6ex} \\
                    
                     $s_{\mathbf{R}^\ast} = [ (\mathbf{\hat{e}}_z^{B})^T \ \ (\mathbf{\hat{e}}_z^{B_{t - 4}})^T \ \ 
                     (\mathbf{\hat{e}}_z^{B_{t - 8}})^T ] ^ T$ 
                     \rule{0pt}{2.6ex} \\ \\
                    
                    $s_\mathbf{v}=[(\mathbf{R}_{B}^T\mathrm{v}_{B})^T \ \ 
                    (\mathbf{R}_{B_{t-4}}^T\mathrm{v}_{B_{t-4}})^T \ \ 
                    (\mathbf{R}_{B_{t-8}}^T\mathrm{v}_{B_{t-8}})^T$ \\ 
                    $\quad (\mathbf{R}_B^T\mathrm{\omega}_B)^T \ \ 
                    (\mathbf{R}_{B_{t-4}}^T\mathrm{\omega}_{B_{t-4}})^T \ \ 
                    (\mathbf{R}_{B_{t-8}}^T\mathrm{\omega}_{B_{t-8}})^T]^T$ \rule{0pt}{2.6ex} \\ \\
                    
                    $s_{\delta\mathbf{v}}=[(\mathbf{R}_{B}^T(\mathrm{v}^\ast_{B} - \mathrm{v}_{B}))^T \ \ 
                    (\mathbf{R}_{B_{t-4}}^T(\mathrm{v}^\ast_{B_{t-4}} - \mathrm{v}_{B_{t-4}}))^T \ \
                    (\mathbf{R}_{B_{t-8}}^T(\mathrm{v}^\ast_{B_{t-8}} - \mathrm{v}_{B_{t-8}}))^T$ \\ 
                    $\quad (\mathbf{R}_B^T(\mathrm{\omega}^\ast_B - \mathrm{\omega}_B))^T \ \ 
                    (\mathbf{R}_{B_{t-4}}^T(\mathrm{\omega}^\ast_{B_{t-4}} - \mathrm{\omega}_{B_{t-4}}))^T \ \ 
                    (\mathbf{R}_{B_{t-8}}^T(\mathrm{\omega}^\ast_{B_{t-8}} - \mathrm{\omega}_{B_{t-8}}))^T]^T$ 
                    \rule{0pt}{2.6ex} \\ \\
                    
                     $s_\mathbf{j}=[{\mathrm{q}_{j}}_{t}^T \ \ {\mathrm{q}_{j}}_{t-4}^T \  \ {\mathrm{q}_{j}}_{t-8}^T 
                     \ \ 
                     {\dot{\mathrm{q}}_{j}}{}_{t}^T \ \ {\dot{\mathrm{q}}_{j}}{}_{t-4}^T \ \ 
                     {\dot{\mathrm{q}}_{j}}{}_{t-8}^T]^T$ \rule{0pt}{2.6ex} \\ \\
                    
                    $s_F = [ ({}_{B}r_{BF}^\ast)^T \ \ ({}_{B}r_{BF_{t - 4}}^\ast)^T \ \ ({}_{B}r_{BF_{t - 8}}^\ast)^T$ 
                    \\ $\quad (\mathbf{R}_{B_{t}}^T\dot{r}_{F_{t}}^\ast)^T \ \ (\mathbf{R}_{B_{t-4}}^T\dot{r}_{F_{t - 4}}^\ast)^T \ \ (\mathbf{R}_{B_{t-8}}^T\dot{r}_{F_{t - 8}}^\ast)^T]^T$
                    \rule{0pt}{2.6ex} \\ \\
                    
                     $s_\tau=\delta\tau_{t-1}$ \rule{0pt}{2.6ex} \\
                    
                    $s_M=\mathrm{Z}_{\mathbf{M}_H}$ \rule{0pt}{2.6ex} \\
                    
                    $s_b={}_{(B_{\mathbf{M}_H})}r_{({B_{\mathbf{M}_H}}B)}$ \rule{0pt}{3.6ex} \\
                    
                    $s_\mathbf{b}= [({\mathbf{e}_x^{B_{\mathbf{M}_H}}})^T \ \ ({\mathbf{e}_y^{B_{\mathbf{M}_H}}})^T]^T$ \rule{0pt}{3.6ex} \rule[-0.9ex]{0pt}{0pt} \\       
                     \hline 
                \end{tabular}
                \label{tab:dat_state_definitions}
            \end{table}

        The DAT action can be defined as $\mathbf{a}_d:=\langle \delta\tau_j \rangle$, where $\delta\tau_j\in\mathbb{R}^{12}$ is the corrective joint torque introduced in the impedance
        controller as shown in eq.~\ref{eq:dat_impedance_controller}.
        
        \subsubsection{Reward Function}
        Domain adaptive tracking can be expressed as an imitation learning problem where the DAT policy
        tracks the reference motion plan by generating corrective feed-forward joint torques. The reward function
        is then expressed in terms of tracking errors as
        \begin{equation}
            \mathbf{r}_D = \mathbf{r}_b + 0.125\mathbf{r}_{\mathbf{v}} + 1.5{\mathbf{r}}_j + 0.4\mathbf{r}_{\dot{j}}
        \end{equation} where $\mathbf{r}_b$ promotes minimizing base position tracking errors, $\mathbf{r}_{\mathbf{v}}$ encourages base velocity tracking, ${\mathbf{r}}_j$ relates to
        joint position tracking, and $\mathbf{r}_{\dot{j}}$ is the joint velocity tracking reward. The
        reward definitions are given in Table~\ref{tab:dat_reward_definitions}.
        
            \begin{table}[h!]
                \centering
                \caption{Reward term definitions for the domain adaptive tracking MDP.}
                \begin{tabular}{|c|}
                    \hline
                     DAT Rewards \rule{0pt}{2.6ex} \rule[-0.9ex]{0pt}{0pt} \\
                     \hline
                     $\mathbf{r}_b=K(10\Vert {{}_{B}r^\ast_{B{B_{t}}}} \Vert^2)$ \rule{0pt}{2.6ex} \\
                     $\mathbf{r}_{\mathbf{v}}=K(6\Vert \mathbf{R}_{B}^T(\mathrm{v}^\ast_{B} - \mathrm{v}_{B}) 
                     \Vert^2) + K(4\Vert \mathbf{R}_{B}^T(\mathrm{\omega}^\ast_{B} - \mathrm{\omega}_{B}) \Vert^2)$ 
                     \rule{0pt}{2.6ex} \\
                     ${\mathbf{r}}_j = K(20\Vert \mathrm{q}_j^\ast - \mathrm{q}_j \Vert^2)$ \rule{0pt}{2.6ex} 
                     \\
                    $\mathbf{r}_{\dot{j}}= K(8\Vert \mathrm{\dot{q}}_j^\ast - \mathrm{\dot{q}}_j \Vert^2)$ 
                    \rule{0pt}{2.6ex} \rule[-2.0ex]{0pt}{0pt} \\
                     \hline 
                \end{tabular}
                \label{tab:dat_reward_definitions}
            \end{table}

        \subsubsection{Episode Termination Criteria}
            The termination criteria include:
            \begin{itemize}
                \item Any of the robot links except for the feet collide with the terrain.
                \item Robot self-collision.
            \end{itemize}

        \subsubsection{Policy Parameterization}
            Unlike in the case of footstep planning which utilizes a stochastic policy, 
            we parameterize the DAT policy in the form of a discrete probability distribution,
            wherein, for a particular action, $\mathbf{a}_{s_d} \in A$, the distribution
            ${\pi_{\theta_d}}(\mathbf{a} |
            \mathbf{s}_d):= \mathbf{\delta}^{s_d}_{n} \ \forall{\mathbf{a}_{n}\in A}$ where
            $\mathbf{\delta}$ refers to Kronecker delta. Since the probability 
            is non-zero for only a particular action in a given state, we can
            represent the policy as a \textit{deterministic}
            mapping between a state to action. We therefore
            rewrite ${\pi_{\theta_d}}(s) : S \rightarrow A$. We model the policy using a neural
            network. The architecture is presented in Fig.~\ref{fig:training}. Additionally, we
            clip the action in the range $[-40, 40]\,$\si{\newton\meter}.
            We use the twin delayed deep deterministic policy gradient (TD3)~\cite{fujimoto2018addressing} algorithm to train our domain 
            adaptive tracking policy.
            Exploration is performed using
            Gaussian noise regularization as opposed to exploration through Ornstein-Uhlenbeck process.
            For this, we used the RLlib TD3 implementation. Note that, for the footstep planner,
            the stochasticity of the footstep planning policy was leveraged to perform
            exploration during training. In the case of the deterministic DAT policy, however,
            we perform exploration by introducing noise during execution of an action. We detail upon the training
            in Section~\ref{sec:training_dat}.

    \subsection{Recovery Controller}
        \label{sec:recovery_controller_description}
        To address system disturbances and 
        external perturbations, the motion controller
        additionally employs an inverted pendulum model-based stabilizing cost term in its
        foothold optimizer. The cost,
        $w_s(\mathbf{c}^\ast-\mathbf{c}_{{hip}_i})\sqrt{h/g}$, for hip $i$ with height $h$ and acceleration due to gravity $g$, 
        encourages the foothold optimizer to
        minimize the error between the desired velocity $\mathbf{c}^\ast$ and the hip velocity $\mathbf{c}_{{hip}_i}$. This
        results in generation of a desired foothold in the direction of the perturbation. Depending on the
        weighting $w_s$, the foothold optimizer either prioritizes reference velocity tracking or stability. 
        This, however, relates to the motion controller \textit{adapting} to the external disturbances as opposed to
        \textit{resisting} them. Therefore, for perturbations
        of large magnitude, adapting to the disturbances by executing feet swing motions to track footholds in
        the direction of the perturbations may not be responsive enough, especially, considering the joint velocity
        limits utilized in the optimization problem. We therefore introduce the recovery controller for following
        reasons:
        \begin{enumerate}
            \item The motion controller executes motion plans to track the desired feet positions generated by
            the RL footstep planner. These feet positions are only updated after each gait stride. With system
            disturbances, the CoM position may drift while the desired feet positions are still not updated. This
            would result in the ZMP drifting significantly from the 
            support region eventually resulting in failure if no recovery motion is executed.
            \item Perturbations of large magnitude cannot be fully addressed by adapting the motion plans. The ability
            to resist these disturbances allows us to better address such perturbations and also reduce the velocity command 
            tracking errors, while only adapting to the
            disturbances when the perturbations are extremely large.
        \end{enumerate}
        In this regard, the recovery controller (RC), operating at \SI{400}{\hertz}, is 
        activated when any of the empirically tuned criteria 
        relating to the robot state, described by $\mathrm{q}$ and $\mathrm{u}$, exceed the limits. These are provided in the Appendix.
        RC is deactivated and the control switches back to the motion controller when each of the
        robot state criteria lies within its deactivation limit (also provided in the Appendix).
            
            An evaluation of the recovery controller's robustness to external perturbations is provided in
            Section~\ref{sec:results_recovery_controller}. As shown in Fig.~\ref{fig:perturbations}, the recovery controller
            is able to resist lateral forces of more than twice the magnitude compared to the dynamic gaits motion controller.
            
            We express the recovery control problem in the framework of MDP directly building upon
            our prior work~\cite{gangapurwala2020guided}.

            \subsubsection{State and Action}
                For the purpose of recovery control over uneven
            terrain, we augment the state space defined in~\cite{gangapurwala2020guided}
            with additional terrain information.
            The RC state, $\mathbf{s}_r\in\mathbb{R}^{201}$, is defined as $\mathbf{s}_r:=\langle s_R, s_v, s_\mathbf{j}, s_M, 
            s_b, s_\mathbf{b}, s_c,
                \mathrm{q}_{j}^\ast \rangle$ and RC action, $\mathbf{a}_r\in\mathbb{R}^{12}$, is given by 
                the tuple $\mathbf{a}_r:=\langle \mathrm{q}_{j}^\ast \rangle$. The
                RC state term definitions are consistent with those for FP and DAT.
                
            \subsubsection{Reward Function}
                The objective of the recovery controller is to stabilize the robot when perturbed. In this regard,
                we aim to maximize the stability margin, while also penalizing foot slippage and energy consumption. We
                also introduce a velocity tracking reward, such that, when utilized with a high-level goal planner, the
                RC executes motions in the direction of the desired velocity. This is based on the assumption that
                the goal planner prioritizes locomotion around accessible regions thereby resulting in fewer failures. 
                Additionally, we also utilize the velocity tracking property of the recovery controller for training.
                This is described in Section~\ref{sec:training_rc}. The
                RC reward function is then given by
                \begin{equation}
                \begin{split}
                    & \mathbf{r}_R = 10\mathbf{r}_m + 3.5\mathbf{r}_v - 
                    0.008d_r\mathbf{r}_\tau - 0.25d_r\mathbf{r}_{fh} 
                   - 3d_r\mathbf{r}_\mu \\  
                  & \qquad - 0.1d_r\mathbf{r}_e - 0.4d_r\mathbf{r}_\mathrm{\dot{q}}
                  - 0.06d_r\mathbf{r}_\mathrm{\ddot{q}} - 
                  0.01d_r\mathbf{r}_{\ddot{f}}
                \end{split}
                \end{equation} where joint velocity reward term 
                $\mathbf{r}_\mathrm{\dot{q}}=\Vert max(\mathrm{|\dot{q}}_j|-10, 0)
                \Vert^2$, feet clearance reward term
                $\mathbf{r}_{fh}=\sum_{i}{(\mathrm{r}_{{F,i}_z}-0.085)^{2}}\;
                \forall{i},\mathrm{c}_{F,i}=0,i\in\{0,1,2,3\}$, joint acceleration
                reward term $\mathbf{r}_\mathrm{\ddot{q}}=\Vert max(|\mathrm{\ddot{q}}_j|-75,0) \Vert^2$,
                feet acceleration reward term $\mathbf{r}_{\ddot{f}}=\Vert \mathrm{\ddot{r}}_{{F}} \Vert^2$ and $d_r$ is the reward scaling term increased as training 
                progresses, updated by the rule $d_{r_{k+1}}\leftarrow{d}_{r_{k}}^{0.996}$
                where $k$ refers to the policy iterations and $d_{r_{0}}=0.3$. The rest
                of the reward term definitions are consistent with FP.

        \subsubsection{Episode Termination Criteria}
            We utilize the same termination criteria as employed for DAT.

            \subsubsection{Policy Parameterization}
            We parameterize the recovery control policy as a Gaussian 
            distribution
            ${\pi_{\theta_r}}(\mathbf{a} | 
            \mathbf{s}_r):=\mathcal{N}(\mathbf{a} | 
            {\mathbf{\mu}_{\theta_r}}(\mathbf{s}_r), 
            {\mathbf{\sigma}_{\theta_r}})$. The mean 
            ${\mathbf{\mu}_{\theta_r}}(\mathbf{s}_r)$ is
            generated using a neural network as a function of the state,
            while the standard deviation
            ${\mathbf{\sigma}_{\theta_r}}$ is independent of the state and is utilized
            during training for exploration. The recovery control network architecture 
            representing the function mapping 
            state to the action mean is shown in Fig.~\ref{fig:training}. We use
            the RLlib implementation of proximal 
            policy optimization 
            (PPO)~\cite{schulman2017proximal} to train the recovery control policy. The 
            training is detailed in Section~\ref{sec:training_rc}.

\section{Training and Evaluation}
\label{sec:training}
            \begin{figure*}
            \centering
            \includegraphics[width=\textwidth]{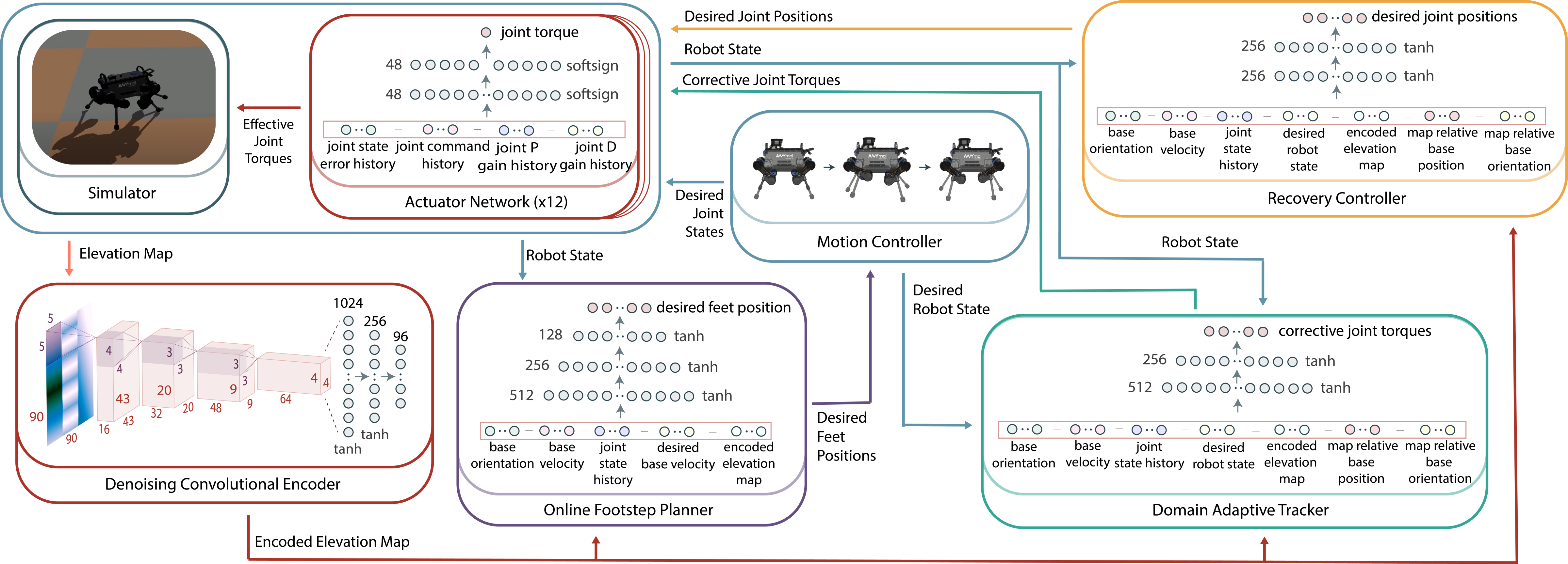}
            \caption{Overview of the approach used to train the RL policies as part of our framework. We introduce the denoising convolutional encoder to embed the
            elevation map into a lower-dimensional representation. This, along with the robot states and desired velocity commands becomes the observation space for each
            of the RL policies. Since we train the RL policies in simulation, we use an actuator network to approximately model the actuation dynamics of the real robot to make
            sim-to-real transfer feasible.}
            \label{fig:training}
            \vspace{-10pt}
            \end{figure*}

    This section details upon the training, testing and deployment setups for
    each of the RL policies. We trained these policies using the \textit{RaiSim}~\cite{raisim} physics simulator for its extremely
    fast contact dynamics computation. Moreover, RaiSim supports online reloading of heightmaps enabling the user to simulate
    terrains making it suitable for training terrain-aware RL policies.
    We defer the reader to our previous work~\cite{gangapurwala2020guided} for a detailed analysis on the preference for
    RaiSim over other widely accessible simulation platforms. Figure~\ref{fig:training}
    represents an overview of the training approach.

    \subsection{Footstep Planner}
    \label{section:footstep_planner_training_eval}
        \subsubsection{Terrain Generation}
        \label{section:terrain_generation}
        Similar to~\cite{lee2020learning}, 
        we introduce a large set of terrains, 10k in our case where each terrain
        is represented as $\mathbf{T}\in\mathbb{R}^{1001\times1001}$, into our simulation
        environment to obtain robust policies
        which can generalize to a wide domain of terrains. Our terrain
        generation procedure is similar to the one introduced in~\cite{gangapurwala2021real}, and is
        further detailed in the Appendix.

    \subsubsection{Actuation Dynamics}
        As introduced in~\cite{hwangbo2019learning},
        we train an actuator network to model the actuation dynamics of the physical system to reduce
        the reality gap by emulating the impedance
        controller response as observed on the real robot. 
        However, unlike in the
        case of~\cite{hwangbo2019learning, gangapurwala2020guided} where the actuators were modeled 
        only for joint position targets with a predefined set of PD gains given by
        \begin{equation}
        \label{eq:recovery_control_actuator_network}
            \tau_j = \mathbf{j}_G(\mathrm{q}_j^\ast, \mathrm{q}_j, \mathrm{\dot{q}}_j) \ ,    
        \end{equation} in this work, 
        we trained the actuator network
        using supervised learning to obtain the effective actuation
        torques
        \begin{equation}
            \tau_j = \mathbf{j}_R(\mathrm{q}_j^\ast, \mathrm{q}_j, \mathrm{\dot{q}}_j^\ast, 
            \mathrm{\dot{q}}_j, \tau_{j_{FF}}, K_p, K_d) \ .
        \end{equation}
        Since the motion controller utilizes different set of PD gains for joints of legs with
        foot in close, open and contact-uncertainty states, we further include the PD gains
        as an input to the approximated actuator model. Note that, even for the same contact states, the actuators located on the
        joints of the same limb can have different PD gains. The actuator network thus maps the input 
        ${j}_s\in\mathbb{R}^{21}$ to an approximation of the joint torque, $\tau_j\in\mathbb{R}$, 
        observed on the physical system. The network architecture and training details are provided in the Appendix.
        
    \subsubsection{Velocity command generation}
        In order to reduce computational overhead, our terrain generation tool does not include collision detection between objects. As a result, these terrains
        contain overlapping objects. In some cases, the terrains
        also have significant deviations in height in small regions which cannot
        be traversed due to the limited size of
        the robots. In this regard, we generate velocity commands with the 
        assumption that the robot tracks the velocity
        commands perfectly. We then check for height deviations in the terrain
        for the next \SI{400}{\milli\second} along the robot's expected trajectory. In case the height deviation exceeds the threshold of
        \SI{0.4}{\metre}, we generate new velocity commands until the height deviation along the expected trajectory remains less than the threshold. 
        This height deviation test is performed every \SI{100}{\milli\second}. The described
        velocity command generation approach significantly
        helps boost the performance of the footstep planning and domain adaptive tracking policies during training.
        Without introducing a strategic velocity command generator, we observed that our policies failed to converge, often
        resulting in reward collapse.
    
    \subsubsection{Domain randomization}
    \label{sec:domain_randomization_footstep_planner}
        Our footstep planner is trained to adapt to 
        the behavior of the motion controller which significantly depends on the dynamic
        properties of the system. In order to achieve a locomotion behavior robust to change
        in system properties, such as added mass, we further utilize 
        domain randomization strategies in addition to terrain randomization. These are shown in
        Table~\ref{table:footstep_planner_dynamics_randomization}. Note that the sampling distributions
        presented in Table~\ref{table:footstep_planner_dynamics_randomization} relate to 
        the limits which the motion controller is robust to. We observed that 
        increasing the range of the distributions caused motion tracking failure.
        In this regard, we perform two training sessions for the footstep planner. Session 1
        employs the distributions presented in Table~\ref{table:footstep_planner_dynamics_randomization}, while
        Session 2 utilizes the domain adaptive tracker to enhance the robustness of the motion controller and 
        therefore employs the distributions presented in Table~\ref{table:dat_dynamics_randomization}.
        Furthermore, since the perception sensors present on the real 
        robot cannot 
        be used to perceive
        the elevation along regions occluded by an obstacle, we apply blurring filter to the elevation map, as shown in 
        eq.~\ref{eq:encoder_input_noise}, to emulate 
        interpolation around unobservable terrain features.
        
        We also apply external perturbations along $\mathbf{e}^B_x$ and $\mathbf{e}^B_y$. However, since the 
        motion controller is not robust to perturbations of high magnitude, we sample the external forces
        from a normal distribution $\mathcal{N}(0, 10)\,$\si{\newton} clipped between $[-30, 30]\,$\si{\newton} for a duration
        in the range $[1,4]\,$\si{\second}. We also perform gait randomization, where we select the trot gait with
        a probability of 0.7, crawl with a probability of 0.2, and amble \& pace with a probability of 0.05 each.

            \begin{center}
        \begin{table}[htbp]
            \caption{Domain randomization properties utilized in Session 1 for training the
            footstep planning policy.}
            \centering
            \begin{tabular}{ |c|c| }
                \hline
                 \textbf{Property} & \textbf{Sampling Distribution}  \\
                \hline
                Gravity, $g=\SI{9.81}{\metre\per\square\second}$ & $\mathcal{U}(0.96g, 1.04g)$ \\ 
                \hline
              Actuation Torque Scaling & $\mathcal{U}(0.9, 1.1)$ \\ 
                \hline
            Robot Link Mass Scaling & $\mathcal{U}(0.93, 1.07)$ \\ 
                \hline
          Robot Link Length Scaling & $\mathcal{U}(0.97, 1.05)$ \\ 
                \hline
            \end{tabular}
            \label{table:footstep_planner_dynamics_randomization}
        \end{table}
        \end{center}
        
        \subsubsection{Reinforcement Learning}
            For Session 1, we employ an interactive direct policy learning approach to perform guided updates of our footstep planning
            policy, $\pi_{\theta_f}$, in order to increase sample efficiency during RL policy optimization. 
            We initialize policy parameters, $\theta_f$,
            using Xavier initialization~\cite{glorot2010understanding}. We consider a fixed-length episode of 1024 desired feet position samples, implying 1024 gait strides are executed in each episode. During training, we also perform 
            locomotion over flat terrain. In this case,
            we use the motion controller to track the feet position generated by the policy
            and store it in a buffer $\mathcal{D}$. Additionally, we generate the desired feet position using the foothold optimizer of the 
            Dynamic Gaits motion controller for the same robot state, $\mathbf{s}_f$, and add it to $\mathcal{D}$. After desired number of episodes used for locomotion over flat ground, we perform back-propagation 
            through the footstep planning policy using the
            mean-squared error loss between the predicted and the expert samples. For uneven
            terrain, we utilize SAC to perform policy
            exploration so as to learn to generate feet position that avoid edges and slopes while ensuring the stability
            of the robot. We alternate between policy optimization through exploration and the \textit{interactive guided policy learning} during training.
            The ratio of episodes used to perform guided learning to the number of episodes used to perform policy exploration is reduced as
            training progresses. We refer to this approach as interactive guided policy optimization (IGPO). Algorithm~\ref{alg:footstep_planning} describes
            the IGPO approach used for training the footstep planning policy.
            It is important to
                note that since we use a curriculum based
                reward transition given by the factor $d_f$ as shown in eq.~\ref{eq:running_reward_fp} and
                eq.~\ref{eq:final_reward_fp}, we continuously push new state transition tuples to the
                SAC replay buffer while discarding the older samples to ensure that the action-value 
                function
                approximations utilized in SAC remain relevant to the latest objective weighted by $d_f$.
                Moreover, while performing guided learning, we set the target value of the
                state-dependent stochasticity based on $\sigma_{\theta_f}(\mathbf{s}_f)$ to zero.

            \begin{algorithm}[htbp!] 
            \begin{algorithmic}[1]
            \Statex \textbf{Input:} $\lambda_{d}$, $n_{eps}$, $l_{eps}$, $H$
            \Statex \textit{Initialize} $\mathcal{D}=\{\}$, $\alpha=1$

            \For{$k = 0,1,2\dots H$}
            \State $\theta_f, \mathcal{D}=\Call{GuidedLearning}{\theta_f,  \mathcal{D}, \alpha n_{eps}, l_{eps}}$
            \State $\theta_f=\Call{PolicyOptimization}{\theta_f,\left(1-\alpha\right) n_{eps}, l_{eps}}$
            \State $\alpha=\text{\textup{e}}^{\lambda_{d}t}$
            \EndFor

            \Function{GuidedLearning}{$\theta_f$, $\mathcal{D}$, $n_{eps}$, $l_{eps}$}
            \For{$i = 0$ \textbf{to} $n_{eps}-1$}
                \For{$t = 0$ \textbf{to} $l_{eps}-1$}
                \State Sample $a\sim\pi_{\theta_f}\left(a|s\right)$
                  \State Generate $a^\ast$ using \Call{DynamicGaits}{s}
                    \State $\mathcal{D}=\mathcal{D}\cup\left\{\left(s,a,a^\ast\right)\right\}$
                    \State $s, r = \Call{RLEnvironmentStep}{a}$
                \EndFor
                \State Sample batch of $\{(s,a_{j},a_{j}^{\ast})\}$ from $\mathcal{D}$
                \State Update $\theta_f$ by minimizing $\sum_{j}{||a_{j}-a_{j}^{\ast}||^{2}}$
            \EndFor
            \State \textbf{return} $\theta_f$, $\mathcal{D}$
            \EndFunction

            \Function{PolicyOptimization}{$\theta_f$, $n_{eps}$, $l_{eps}$}
            \For{$i = 0$ \textbf{to} $n_{eps}-1$}
            \State $\tau=\{\}$
            \For{$t = 0$ \textbf{to} $l_{eps}-1$}
              \State Sample $a\sim\pi_{\theta_f}\left(a|s\right)$
              \State $s,r= \Call{RLEnvironmentStep}{a}$
              \State $\tau=\tau\cup\left(s,a,r\right)$
            \EndFor
            \State Update $\theta_f$ using SAC
            \EndFor
            \State \textbf{return} $\theta_f$
            \EndFunction

            \end{algorithmic}
            \caption{\label{alg:footstep_planning} Interactive Guided Policy Optimization}
            \end{algorithm}

            As training progresses, we increase the complexity of the RL environment by scaling the maximum terrain elevation. As introduced in Section~\ref{section:terrain_generation},
            we represent the elements of $\mathbf{T}$ in the range of $[0, h_{max}]$ where $h_{max}=\SI{2}{\meter}$. During training, we scale $h_{max}$ using $d_{hf}$ which is updated by the rule 
            $d_{{hf}_{k+1}}\leftarrow{}d_{{hf}_k}^{0.9992}$
            where $k$ refers to the policy iterations and $d_{hf_{0}}=0.01$.
            
            For Session 2, we do not perform guided learning and only utilize SAC to make 
            footstep planning more robust to the changes in modeled system dynamics. We
            also set each of the curriculum factors to one. This additional training allows
            us to utilize the same footstep planner trained with ANYmal B for deployment on 
            ANYmal C. The training hyperparameters and policy convergence iterations are provided in the Appendix.
            
        \subsubsection{Testing}
            For the purpose of evaluation, we utilize the stochastic policy in a deterministic manner. Therefore,
            we consider the $\mu_{\theta_f}(\mathbf{s}_f)$ as the desired action, ${}_{H}r^\ast_{H{F_xy}}$, 
            and discard $\sigma_{\theta_f}(\mathbf{s}_f)$. We detail the test results in 
            Section~\ref{sec:results_footstep_planning}.
            
        \subsubsection{Deployment}
            The observations utilized for the footstep planner are accessible on the physical robots. 
            During deployment on the real robots, 
            we observe a computational delay of approximately \SI{35}{\milli\second} during elevation map 
            extraction and encoding. The elevation map processing is therefore executed in a parallel thread while the motion 
            tracking controller
            runs in the main thread at \SI{400}{\hertz}. The robot maintains the previous desired footholds while the
            forward pass through the elevation map encoder and the 
            footstep planning policy is performed, after which, the new motion plans for the updated 
            footholds are generated and tracked. This methodology remains consistent for both ANYmal B and C.
            A forward pass through the FP policy network
            requires approximately \SI{195}{\micro\second} on
            both the robots.
            
    \subsection{Domain Adaptive Tracker}
    \label{sec:training_dat}
        We utilize the terrain generator, actuator network and velocity command generator introduced in 
        Section~\ref{section:footstep_planner_training_eval} for training the DAT policy. In addition, 
        we use the footstep planning policy obtained after training Session 1 to train our domain
        adaptive tracking policy.
        
        \subsubsection{Domain Randomization}
            The randomization parameters and the corresponding sampling distributions representing
            the limits are provided in
            Table~\ref{table:dat_dynamics_randomization}. During training 
            of the DAT policy, the parameter randomization is approached using 
            a curriculum learning setup where initial limits of the distributions
            are set to 1. The lower and upper limits are then linearly expanded 
            after every policy iteration, $k$, based on the rule 
            $l_{l,k}=max(\frac{k}{2000}(l_{min} -1 ) + 1, l_{min})$ and 
            $l_{u,k}=min(\frac{k}{2000}(l_{max} -1 ) + 1, l_{max})$ respectively where the $l_{min}$
            and $l_{max}$ are the lower and upper limits presented in Table~\ref{table:dat_dynamics_randomization}. We
            also introduce external perturbations and gait randomization as done for the footstep planner.
            
            \begin{center}
            \begin{table}[htbp]
                \caption{Domain randomization properties utilized for training the
            domain adaptive tracking policy. Here $\tau_j^\ast$ is the expected torque
                output of the impedance controller.}
                \centering
                \begin{tabular}{ |c|c| }
                    \hline
                     \textbf{Property} & \textbf{Sampling Distribution}  \\
                    \hline
                    Gravity, $g=\SI{9.81}{\metre\per\square\second}$ & $\mathcal{U}(0.96g, 1.04g)$ \\ 
                    \hline
                  Actuation Torque Scaling & $\mathcal{U}(0.85, 1.15)$ \\ 
                    \hline
                Robot Link Mass Scaling & $\mathcal{U}(0.9, 1.10)$ \\ 
                    \hline
              Robot Link Length Scaling & $\mathcal{U}(0.85, 1.15)$ \\ 
                    \hline
                 \makecell{Actuation Damping Gain $G$ such that \\ $\tau_{j,t}=G\tau_{j,t}^\ast + (1-G)\tau_{j,t-1}$} & 
                 \makecell{$\mathcal{U}(0.8, 1)$} \\ 
                    \hline
                \end{tabular}
                \label{table:dat_dynamics_randomization}
            \end{table}
            \end{center}

    \subsubsection{Reinforcement Learning}
        We employ TD3 for training the DAT policy allowing us to perform constrained exploration. We observed that the 
        stochasticity 
        associated with policies trained using SAC and PPO required exhaustive hyperparameter
        tuning. Without significant tuning, the exploration resulted in 
        introduction
        of large corrective torques. This caused the whole-body controller to significantly drift from the generated motion plans
        causing aggressive
        recovery and eventual failure. This inhibited learning since the corrective torques mostly resulted in episode 
        termination. We therefore
        used TD3 to initially perform constrained exploration using uncorrelated 
        Gaussian sampling. During training, we scale the standard deviation of the standard Gaussian distribution
        according to the rule $d_{a,k+1} \leftarrow d_{a,k}^{0.9994}$
        where $k$ refers to the elapsed policy iterations and the initial scaling $d_{a,0}=0.01$. Additionally,
        we scale $\delta\tau_j$ by a factor of $40$ before adding it to the feed-forward torque generated
        by the whole-body controller.
        
        As in the case of training the footstep planner, we use the same curriculum approach of scaling the maximum 
        terrain elevation. We used the standard RLlib TD3 implementation for reinforcement learning. The hyperparameters are provided in 
        the Appendix.

    \subsubsection{Testing}
        Since we obtain a deterministic policy, we do not perform any modifications to the neural network
        and directly employ it as a function mapping state to action. The test results are 
        presented in Section~\ref{sec:results_dat}.

    \subsubsection{Deployment}
        The DAT is executed along with the whole-body controller in the main control loop
        at \SI{400}{\hertz}. The generated corrective torques are then added
        to the output of the whole-body controller, after which the desired joint states
        are forwarded to the actuator. The DAT remains 
        active when the recovery controller is deactivated.
        A forward pass through the DAT policy network
        requires approximately \SI{66}{\micro\second}.

    \subsection{Recovery Controller}
    \label{sec:training_rc}
        Our recovery controller training setup is directly built on top of our previous
        work~\cite{gangapurwala2020guided}. However, unlike in the case of our previous
        work, we do not employ a guided policy optimization strategy, since the
        guided learning focused on velocity command tracking as opposed to recovery control, and 
        instead utilize the training approach presented in~\cite{hwangbo2019learning} 
        performing reinforcement learning using PPO.
        
        For training the RC policy, we employ the terrain 
        generator introduced in 
        Section~\ref{section:terrain_generation}.
        Since we perform joint position control with the recovery 
        policy, 
        we also use the actuator network to model the actuation 
        dynamics for fixed $K_p=50$ and $K_d=0.1$ with
        zero velocity target. This is described in 
        eq.~\ref{eq:recovery_control_actuator_network}. We defer the
        reader to the original work~\cite{hwangbo2019learning} for 
        detailed description of the joint position target actuator
        network.

        \subsubsection{Domain Randomization}
            We use the same randomization strategy, including curriculum learning,
            as used for DAT for training the recovery controller.
            
        \subsubsection{Reinforcement Learning}
            Since our objective is to learn a recovery controller which can stabilize to perturbations in
            addition to performing state recovery, we introduce external perturbations in our training environment.
            However, in order to encompass a wide possibility of robot states when the recovery controller is activated,
            we initially train the RC policy to track reference velocity commands. We then transition from 
            this objective to that of recovery control. For this, we initially learn to locomote over flat ground after
            which we introduce larger external perturbations while also scaling up the terrain elevation. The approach
            is detailed in Algorithm~\ref{alg:recovery_control}.

            \begin{algorithm}[htbp!] 
            \begin{algorithmic}[1]
            \Statex \textbf{Input:} $n_{eps}$, $l_{eps}$, $H$
            \Statex \textit{Initialize} $e_{max}=1e-4$, 
            $f_{max}=0.01$, 
            $t_{f_{max}}=0.01$, $t_{f}=0$,
            $t_{a}=0$
            
            \For{$k = 0,1,2\dots H$}
                \For{$i = 0$ \textbf{to} $n_{eps}-1$}
                    \State $t_{vel}, s=
                    \Call{RLEnvironmentReset}{e_{max}}$
                    \For{$t = 0$ \textbf{to} $l_{eps}-1$}
                        \If{$t >= t_{vel}$}
                            \State Generate new velocity command
                            \State Sample 
                            $t_{vel}\sim\mathcal{U}(1,3)$
                        \EndIf
                        
                        \State Sample action
                        $a\sim\pi_{\theta_r}\left(a|s\right)$

                        \If{$t >= t_{a}$ and $t < t_{a} + t_{f}$}
                            \State Apply external force $f$
                        \ElsIf{$t > t_{a} + t_{f}$}
                            \State Sample external force, $f\sim\mathcal{N}(0, 45f_{max})$
                            \State Sample 
                            $t_{a}\sim\mathcal{U}(t,t+2)$ and 
                            $t_{f}\sim\mathcal{U}(0,4t_{f_{max}})$
                        \EndIf
                        
                        \State $s,r=\Call{RLEnvironmentStep}{a}$
                        \If{$s$ is a terminal state}
                        \State $t_{vel},s=\Call{RLEnvironmentReset}{e_{max}}$
                        \EndIf
                    \EndFor
                \EndFor
                \State Update $\pi_{\theta_r}$ using PPO
                \State $e_{max} \leftarrow e_{max}^{0.9996}$, $f_{max} \leftarrow f_{max}^{0.9984}$, $t_{f_{max}} \leftarrow t_{f_{max}}^{0.997}$
            \EndFor
            \Function{RLEnvironmentReset}{$e_{max}$}
                \State Sample terrain and scale elevation by $e_{max}$
                \State Randomize robot pose
                \State Randomize $\mathrm{q}_j$ where
                $max(|\mathrm{q}_j - \mathrm{q}_j^n|) < 0.3$
                \State Randomize $\dot{\mathrm{q}}_j$ where $max(|\dot{\mathrm{q}}_j|) < 1.0$
                \State Update state $s$ and set $t_{vel}=0$
            \State \textbf{return} $t_{vel}$, $s$
            \EndFunction

            \end{algorithmic}
            \caption{\label{alg:recovery_control} Recovery Control Training 
            Objective Transition}
            \end{algorithm}

    \subsubsection{Testing}
        As in the case of FP, we utilize the stochastic
        policy in a deterministic manner. In this regard, 
        the action $\mathbf{a}_r=\mu_{\theta_r}(\mathbf{s}_r)$. The
        desired joint positions, $\mathrm{q}_j^\ast=\mathbf{a}_r$
        are then forwarded to the actuator to generate the effective
        actuation torques. The results are presented in
        Section~\ref{sec:results_recovery_controller}.

    \subsubsection{Deployment}
        We query the robot state at 
        each control step executed at \SI{400}{\hertz} to activate RC according to the limits described
        in Section~\ref{sec:recovery_controller_description}. When active,
        RC is run in the main control loop to generate the desired
        joint positions. While the RC remains active, the footstep planner, motion controller 
        and domain adaptive tracker are deactivated. 
        A forward pass through the RC policy network
        requires approximately \SI{42}{\micro\second}.

    \section{Results and Discussion}
    \label{sec:results_and_discussion}
    This section
    details the results obtained for each of the 
    RL policies introduced in our framework. Note that,
    unless otherwise stated, the results presented here relate to the
    ANYmal B quadruped.

    \subsection{Footstep Planning}
    \label{sec:results_footstep_planning}
\begin{figure}
            \begin{subfigure}{.25\textwidth}
              \centering
              \includegraphics[width=0.98\linewidth]{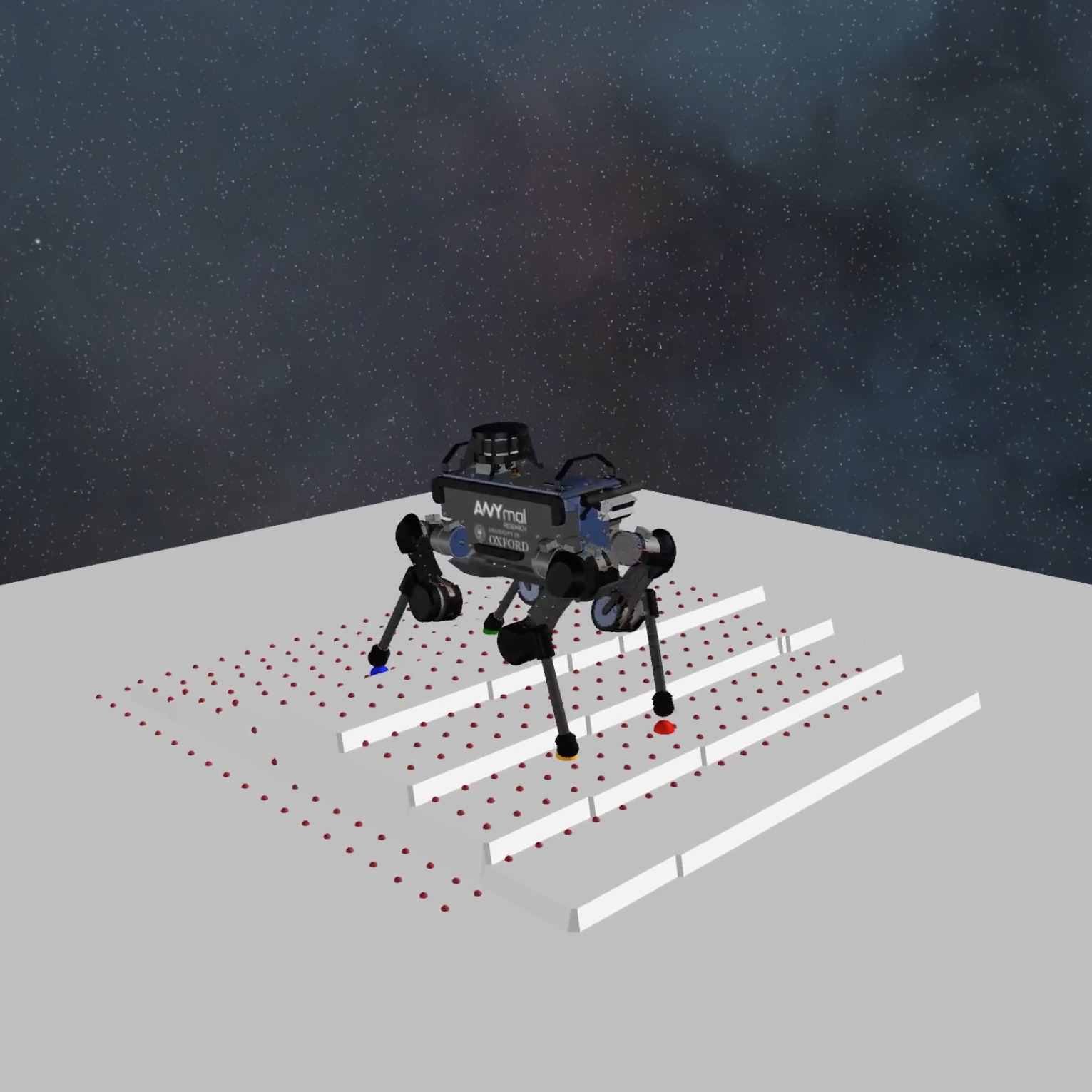}
            \end{subfigure}%
            \begin{subfigure}{.25\textwidth}
              \centering
              \includegraphics[width=0.98\linewidth]{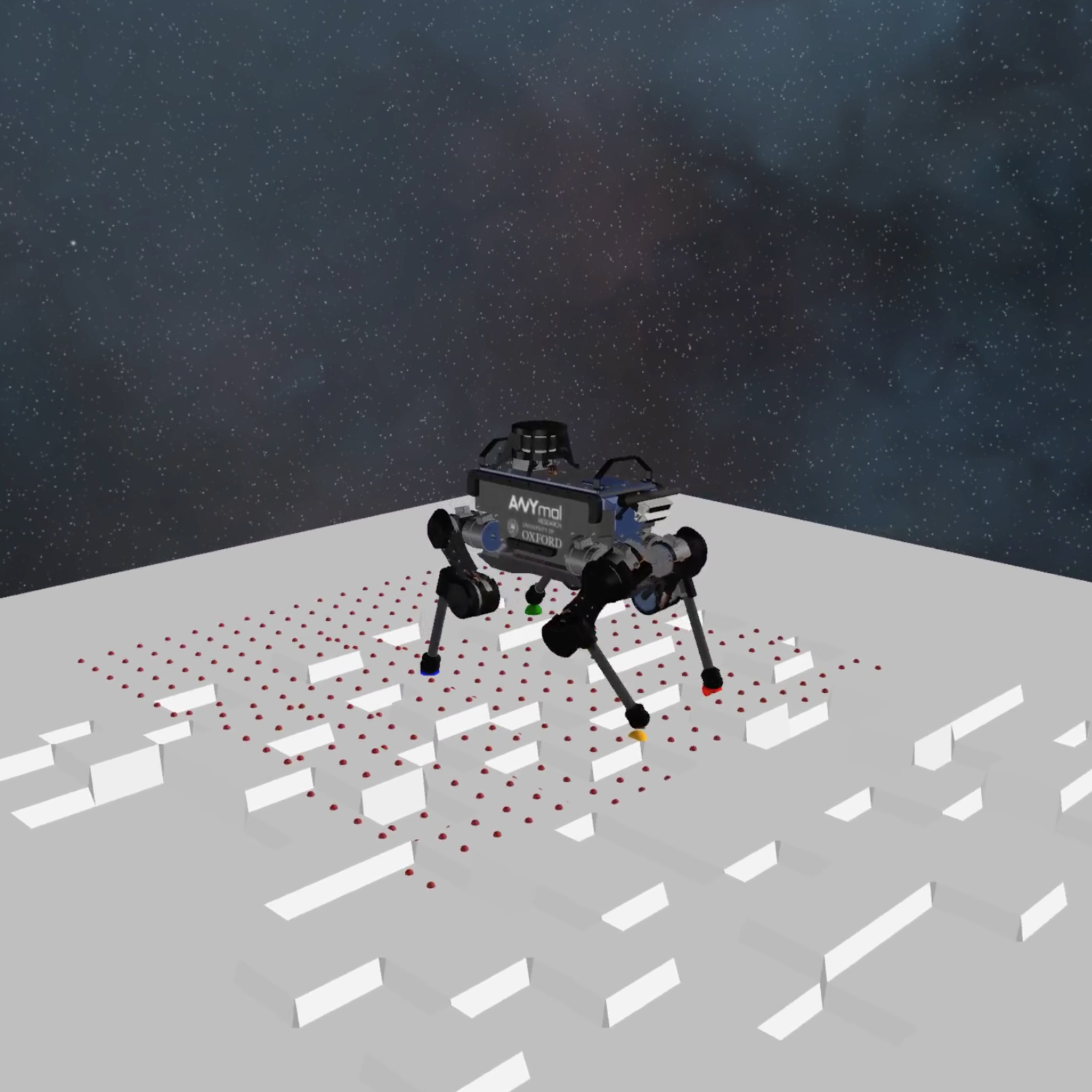}
            \end{subfigure} \vspace{0.1cm} \\
            \begin{subfigure}{.25\textwidth}
              \centering
              \includegraphics[width=0.98\linewidth]{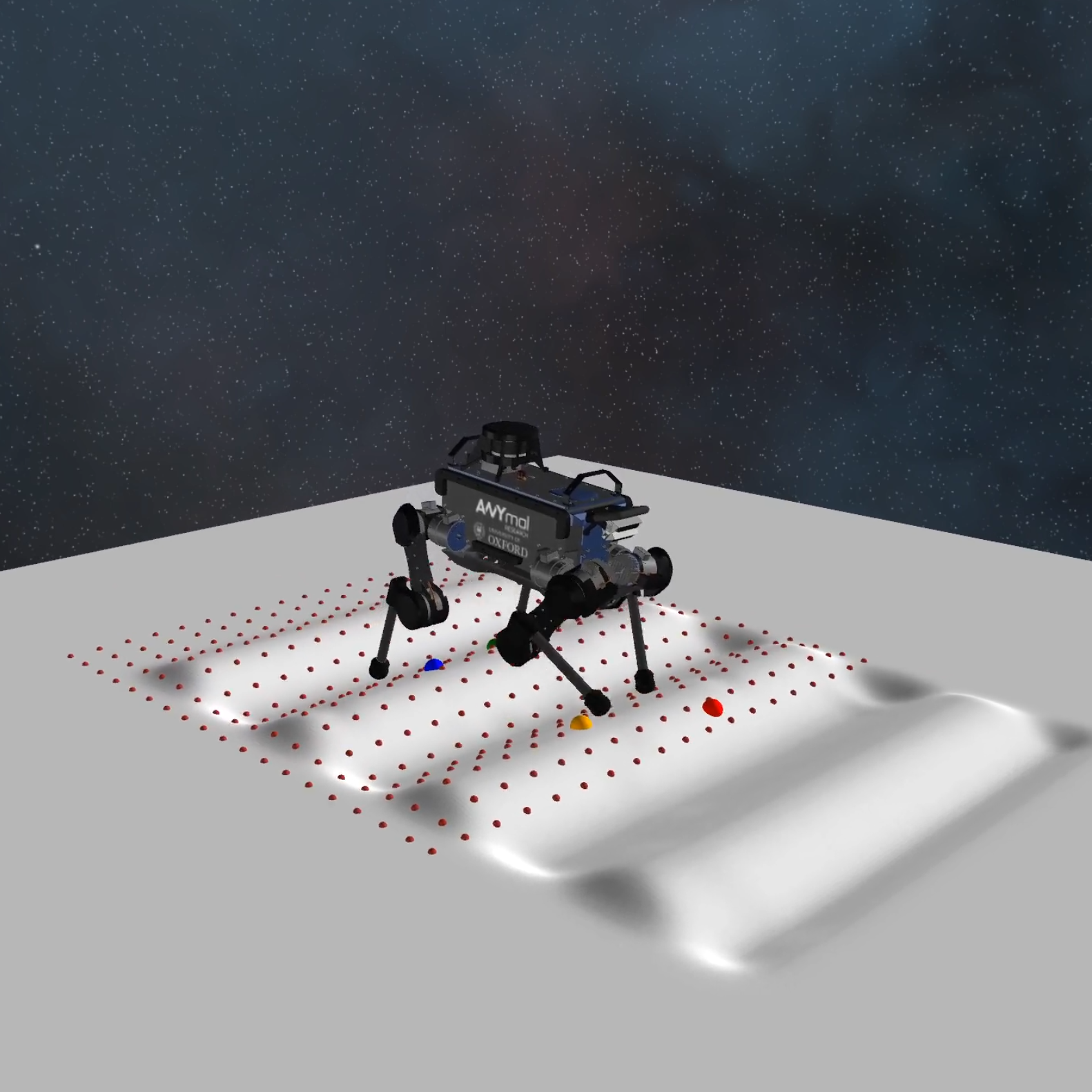}
            \end{subfigure}
            \begin{subfigure}{.25\textwidth}
              \centering
              \includegraphics[width=0.98\linewidth]{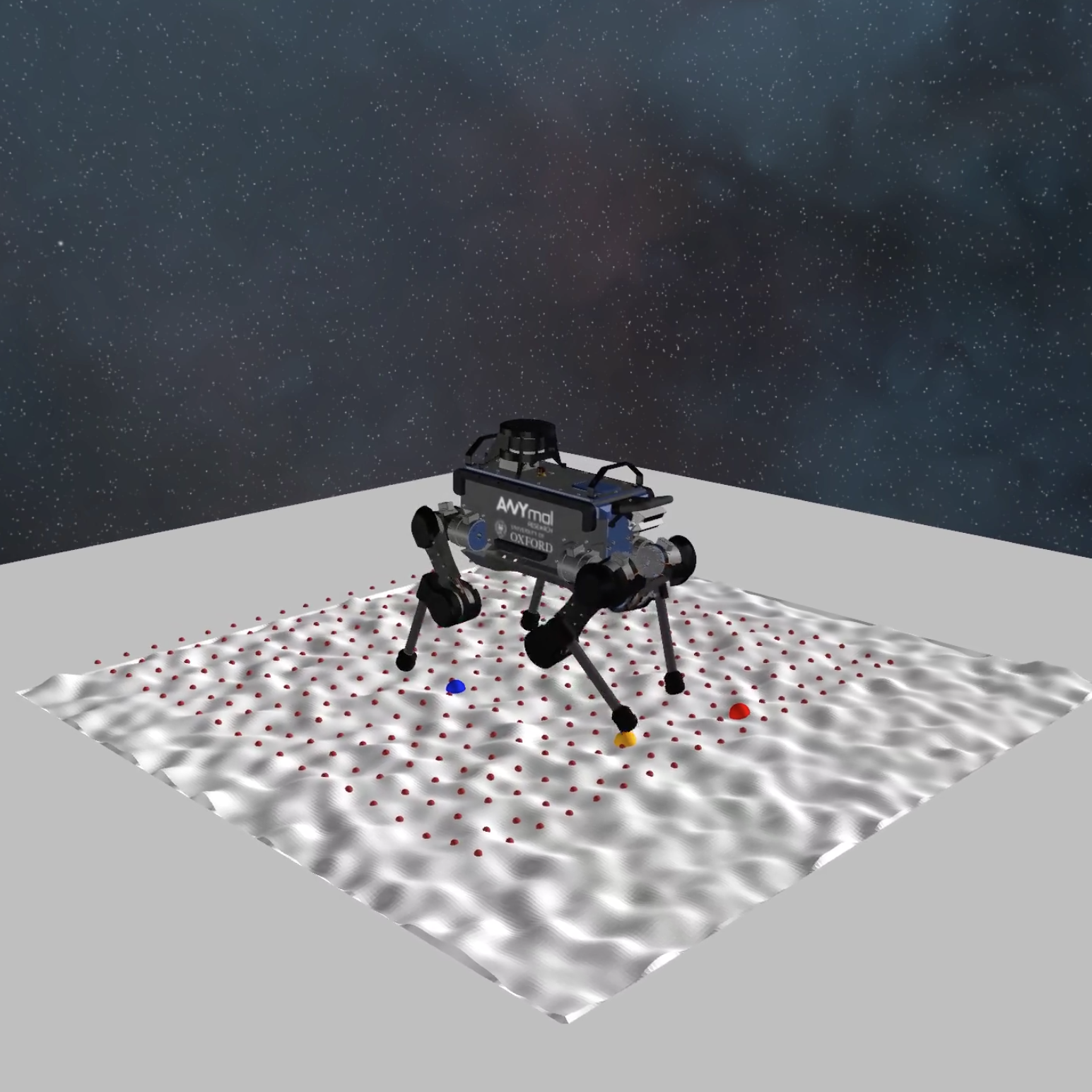}
            \end{subfigure}%
            \caption{\textbf{(Top)} the stairs (left) and brick (right) terrains. \textbf{(Bottom)} the wave (left) and unstructured (right) terrains.}
            \label{fig:terrains_examples}
            \vspace{-10pt}
        \end{figure}

        We evaluate the footstep planning policy, obtained after the 2 training sessions,
        by measuring its success rate when traversing
        different terrains. We set up our simulation platform to perform 
        1000 runs for each of the 3 footstep planners navigating using 2 different gaits - trot and crawl.
        The 3 planners are our RL footstep planner, a baseline perceptive planner and
        a baseline blind planner. In the case of the baseline planners, we use the original foothold
        optimizer of dynamic gaits to generate $r_F^d$ for tracking the desired base velocity commands. We
        further modify $r_{F_z}^d$ by extracting the terrain elevation at the touch-down location
        directly from $\mathbf{M}_H$ (including for the blind controller).
        For the case of perceptive controller, we additionally adjust $r_{F_{xy}}^d$ up to
        a radial deviation of \SI{0.05}{\meter} so as
        to minimize the cost $\mathbf{C}_\mathbf{M}(\hat{r}_{F_{xy},i}) + 2.5\Vert {\hat{r}_{F_{xy},i} - r^n_{F_{xy},i}} \Vert^2$
        for each of the feet.
        For this optimization, we compute the cost by sampling $\hat{r}_{F_{xy},i}$
        along the polar radial coordinate with a resolution
        of \SI{0.005}{\meter} and a polar angular resolution of $\pi/18\;\si{\radian}$.
        We test the performance over 1000 different terrains of varying dimensions for each of the 4
        object types - stairs, bricks, wave and unstructured ground as shown in Fig.~\ref{fig:terrains_examples}.
        Each of these objects are located at the center of a $5\times5\;$\si{\meter\squared} terrain, with the length of the 
        object, 
        along the
        direction of the robot's frontal axis, varying between \SI{2.0}{\meter}
        to \SI{3.6}{\meter}. We consider the following terrain properties.
        \begin{enumerate}
            \item  The amplitude for the unstructured terrain is varied between $[0.0125, 0.025]\,$\si{\meter}.
            \item  For stairs, the number of
                    steps range between $[3, 8]$. The maximum terrain elevation is varied
                    between $[0.25, 0.8]\,$\si{\meter} using clipped normal distribution, 
                    $\mathcal{N}(0.3, 0.1)\,$\si{\meter}. 
            \item   The wave terrain is 
                    generated using a sine function along its length with an
                    amplitude range of  $[0.05, 0.1]\,$\si{\meter}. The 
                    period of the sine function ranges between $\pi/2$ to $\pi$.
            \item For brick terrain, the bricks are made of deformed $10\times10$\si{\centi\meter\squared} patches with 
            extrusions $\{-h, 0, h\}$ where $h\in[0.02, 0.05]\,\si{\meter}$.
        \end{enumerate}
        
        The results obtained for traversal over these terrains are presented in \href{https://youtu.be/1Hm77CfaouE}{Movie S1}.

\begin{figure}
            \begin{subfigure}{.25\textwidth}
              \centering
              \includegraphics[width=.9\linewidth]{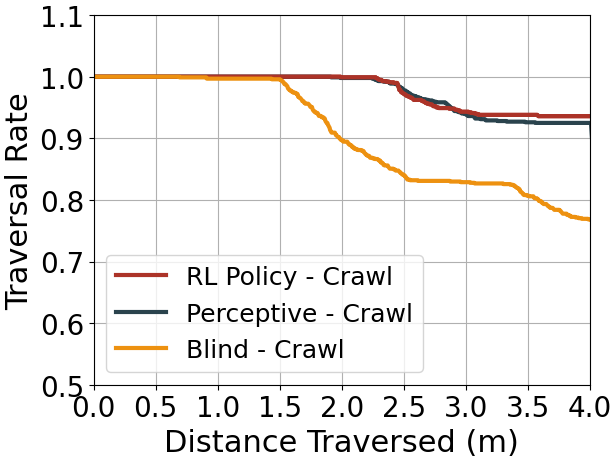}
              \caption{\footnotesize Stairs}
              \label{fig:stairs_traversal}
            \end{subfigure}%
            \begin{subfigure}{.25\textwidth}
              \centering
              \includegraphics[width=.9\linewidth]{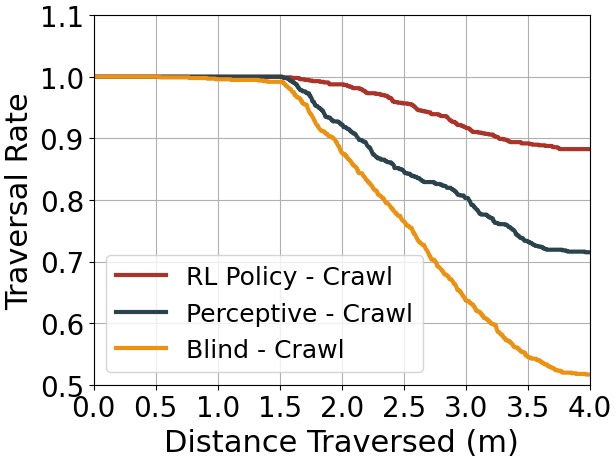}
              \caption{\footnotesize Bricks}
              \label{fig:bricks_traversal}
            \end{subfigure} \vspace{0.1cm} \\
            \begin{subfigure}{.25\textwidth}
              \centering
              \includegraphics[width=.9\linewidth]{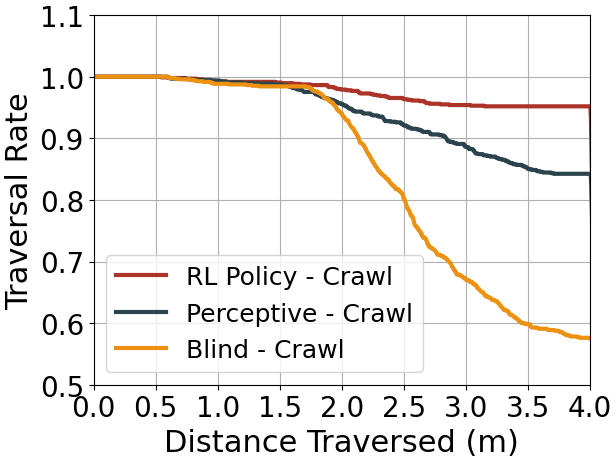}
              \caption{\footnotesize Wave}
              \label{fig:wave_traversal}
            \end{subfigure}
            \begin{subfigure}{.25\textwidth}
              \centering
              \includegraphics[width=.9\linewidth]{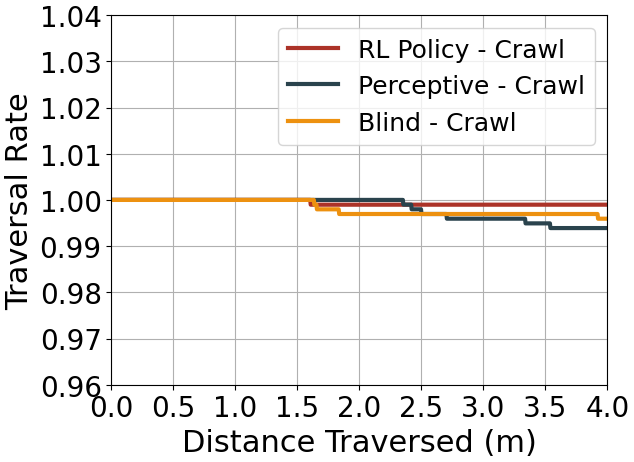}
              \caption{\footnotesize Unstructured}
              \label{fig:unstructured_traversal}
            \end{subfigure} \vspace{0.1cm} \\
            \begin{subfigure}{.25\textwidth}
              \centering
              \includegraphics[width=.9\linewidth]{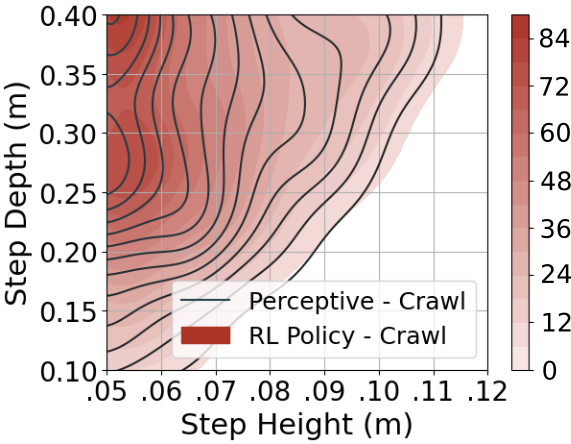}
              \caption{\footnotesize Estimate of success rate for\\ traversal over stairs with varying\\ depth and height}
              \label{fig:stairs_height_depth_sr}
            \end{subfigure}%
            \begin{subfigure}{.25\textwidth}
              \centering
              \includegraphics[width=.9\linewidth]{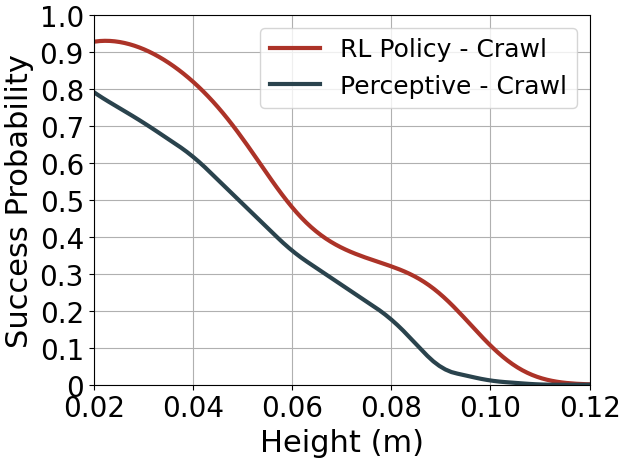}
              \caption{\footnotesize Estimate of success rate for\\terrains with varying brick height}
              \label{fig:bricks_height_sr}
            \end{subfigure}%
            \caption{\textbf{(a-d)} represent the traversal rate of each of the planning policies for the different types of terrains used for
            evaluation. This corresponds to the rate of success observed for the distance traversed.
            \textbf{(e)} is the plot representing the success rate for traversal over stairs with varying step height and depth.
            \textbf{(f)} represents the probability of success observed for brick terrain of varying brick heights.}
            \label{fig:traversal}
        \end{figure}

        For each run, we set the robot at the origin and generate a forward velocity command of
        \SI{0.3}{\meter/\second}. We denote the run as a success if the robot traverses
        a distance of \SI{4}{\meter}, and regard it as a failure if any of the FP
        termination criteria is observed.
        The success rates obtained for each of the
        experiments are represented in Table~\ref{table:footstep_planner}. 
        Figure~\ref{fig:traversal} presents the success rate for 
        traversing each terrain for each of the tested controllers with two gaits.
        \begin{center}
        \begin{table}[htbp]
            \caption{Success rates of policies measured for locomotion over different terrains generated for 1000 runs each.}
            \resizebox{0.48\textwidth}{!}{
            \centering
            \begin{tabular}{ |c|c|c|c||c|c|c| }
                \hline
                 & \multicolumn{3}{c||}{\textbf{Crawl}} & \multicolumn{3}{c|}{\textbf{Trot}} \\
                 & \textbf{Blind} & \textbf{Perceptive} & \textbf{RL Policy} & \textbf{Blind} & \textbf{Perceptive}  & \textbf{RL Policy} \\
                \hline
                \textbf{Unstructured} & 99.6 & 99.4 & \textbf{99.9} & \textbf{99.7}  & 99.6 & 99.6 \\ 
                \hline
                \textbf{Stairs} & 77.0 & 92.5 & \textbf{93.6} & 72.2 & 92.3 & \textbf{92.7} \\
                \hline
                \textbf{Wave} & 58.1 & 83.6 & \textbf{94.4} & 55.7 & 81.5 & \textbf{93.0}  \\
                \hline
                \textbf{Bricks} & 52.1 & 71.5 & \textbf{88.2} & 51.1 & 70.9 & \textbf{85.3} \\
                \hline
            \end{tabular}}
            \label{table:footstep_planner}
        \end{table}
        \end{center}

        The RL footstep planner was observed to exhibit
        preference for wider stance. This directly corresponds to a higher stability margin, a term
        introduced in the reward function during training.
        We measured ${}_{H}r_{HF}$ after every gait stride.
        For traversal over stairs, the absolute mean foot position in frame $H$
        was \SI{0.35}{\meter} and \SI{0.20}{\meter} along its $x$ and $y$ axes, similar to the expected 
        nominal stance value of \SI{0.35}{\meter} and 
        \SI{0.20}{\meter}. However, in the case of unstructured terrain, we computed the mean feet position to be
        $(0.34, 0.23)$\si{\meter}, $(0.35, 0.21)$\si{\meter} for the wave terrain, and $(0.34, 0.22)$\si{\meter} for 
        the bricks terrain.
        
        Figure.~\ref{fig:stairs_height_depth_sr} represents the success
        rate as a function of step height and depth evaluated using \textit{kernel density estimation} (KDE).
        
        The extrusions in the brick
        terrains often caused leg blockages. This was prominent for the blind planner. 
        The ability of the perceptive planner to minimize distance from the edges in the terrain significantly reduced the 
        problem with leg blockage. However, the difference in the feet height resulted in the robot toppling over in most
        cases. With the RL planner, we observed a wide stance and preference for foot placement along regions which minimized the
        deviation in feet height. For the case of the blind planner,
        we computed a mean deviation between the maximum and minimum foot height to be \SI{0.074}{\meter}. For the perceptive planner
        the mean height deviation was \SI{0.083}{\meter}, and \SI{0.057}{\meter} with the RL planner. 
        Figure~\ref{fig:bricks_height_sr}
        illustrates the success rate computed for different brick heights. In this
        case, the brick height only corresponds to the extrusion of the terrain implying that the maximum deviation
        in feet height is twice the height of the bricks. We observed that the RL planner fails to traverse brick terrains
        with maximum elevation of higher than \SI{0.12}{\meter} (corresponding to a maximum height deviation
        of \SI{0.24}{\meter}). This mainly occurs due to the limitations of the physical
        dimensions of the quadruped. 

        We also compared the generalization capability of the footstep planner obtained after training Session 1 and Session 2. We performed the
        experiments discussed above using the ANYmal C quadruped, adapting the motion controller for the parameter description
        of ANYmal C. The corresponding success rates are presented in Table~\ref{table:footstep_planner_compare_sessions}. As expected, we observe a better
        ability to generalize to a different platform with the policy trained using significant domain randomization.
        \begin{center}
        \begin{table}[htbp]
            \caption{Success rates obtained for locomotion over different terrains using the ANYmal C quadruped comparing the RL footstep planner
            performance after training Session 1 and Session 2.}
            \centering
            \begin{tabular}{ |c|c|c||c|c| }
                \hline
                 & \multicolumn{2}{c||}{\textbf{Crawl}} & \multicolumn{2}{c|}{\textbf{Trot}} \\
                 & \textbf{Session 1} & \textbf{Session 2} & \textbf{Session 1}  & \textbf{Session 2} \\
                \hline
                \textbf{Unstructured} & 99.9 & \textbf{99.9}  & 99.4 & \textbf{99.5} \\ 
                \hline
                \textbf{Stairs} & 93.0 & \textbf{93.4} & 90.5 & \textbf{91.9} \\
                \hline
                \textbf{Wave} & 85.7 & \textbf{91.2} & 82.4 & \textbf{88.4}  \\
                \hline
                \textbf{Bricks} & 76.3 & \textbf{85.8} & 71.6 & \textbf{82.2} \\
                \hline
            \end{tabular}
            \label{table:footstep_planner_compare_sessions}
        \end{table}
        \end{center}

    \subsection{Domain Adaptive Tracking}
    \label{sec:results_dat}
         In our experiments, the tracking error associated with the whole-body controller was negligible
         when the robot's base mass, of \SI{18.3546}{\kilogram}, was scaled in the range $\left[0.95, 1.05\right]$. 

\begin{figure}
            \begin{subfigure}{.25\textwidth}
              \centering
              \includegraphics[width=.9\linewidth]{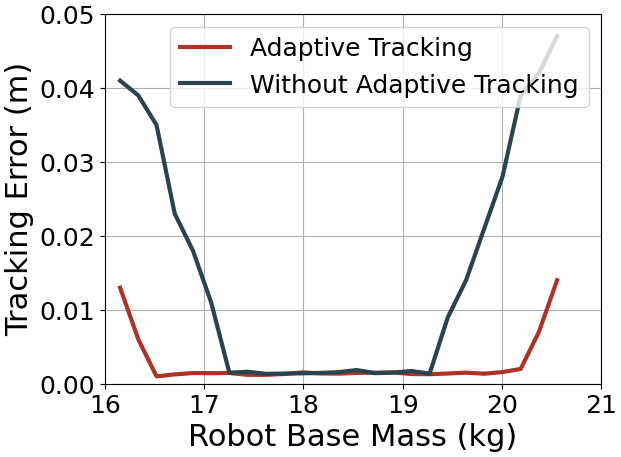}
              \caption{\footnotesize ANYmal B}
              \label{fig:adaptive_anymal_b}
            \end{subfigure}%
            \begin{subfigure}{.25\textwidth}
              \centering
              \includegraphics[width=.9\linewidth]{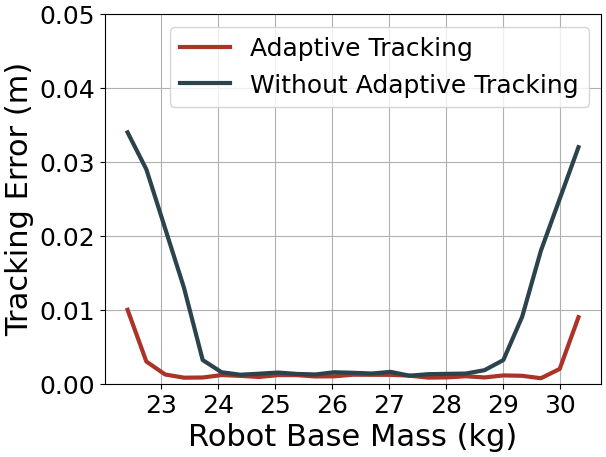}
              \caption{\footnotesize ANYmal C}
              \label{fig:adaptive_anymal_c}
            \end{subfigure}%
            \caption{The absolute mean tracking error observed before termination for varying physical parameters of the robot comparing the effects of the domain adaptive
            tracking policy.}
            \label{fig:adaptive}
        \vspace{-10pt}
        \end{figure}

         However, when we
         decreased the mass by a factor below 0.94, or increased it beyond 1.06,
         the motion controller became unstable
         resulting in increased oscillations. This was caused 
         due to the whole-body controller aggressively correcting for imprecise
         motion tracking. Introducing DAT helped stabilize whole-body motion tracking. We 
         were able to scale the base mass in the range of $[0.84, 1.17]$ without
         increasing the amplitude of the
         oscillations. Figure~\ref{fig:adaptive_anymal_b} illustrates the 
         $\mathrm{p}_{CoM_x}$ tracking error observed before episode termination for different robot base masses. 
         Here, the episode termination criteria include traversing a distance of \SI{5}{\meter} or
         failure of whole-body control.
        
        We also tested our domain adaptive tracking policy, without retraining, on the simulated model of the ANYmal C quadruped. Note that,
        while running experiments on the simulated ANYmal C, we increased the joint torque limit from \SI{\pm40}{\newton\meter} for the case of
        ANYmal B to a limit of \SI{\pm80}{\newton\meter}. 
        The tracking error measured for varying base mass of the ANYmal C quadruped
        is represented in Fig.~\ref{fig:adaptive_anymal_c}.
        With an original base mass
        of \SI{26.3732}{\kilogram}, the motion controller could track the forward velocity command of 
        \SI{0.3}{\meter/\second} even when we scaled this mass in the range $\left[0.91, 1.10\right]$.
        When we introduced the domain adaptive tracking policy, we were further able to extend this range
        to $\left[0.81, 1.19\right]$.

\begin{figure}
              \centering
              \includegraphics[width=.8\linewidth]{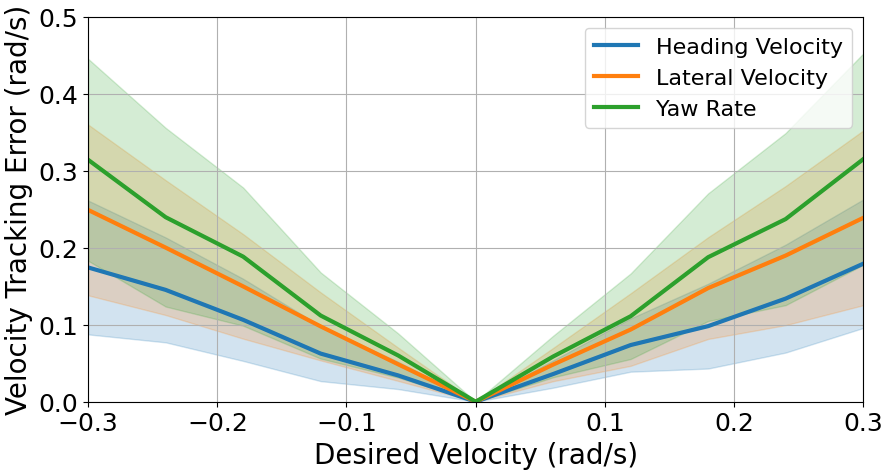}
              \caption{Velocity tracking error observed for the domain adaptive tracking policy 
              when running the motion controller on ANYmal C with the description parameters of ANYmal B.}
              \label{fig:domain_adaptive_tracking_vel}
        \end{figure}

        We performed another test on the simulated ANYmal C robot. However,
        in this case, we used the parameter description of the ANYmal B quadruped. 
        Therefore, the motion controller generated motion plans
        and control behavior for the ANYmal B quadruped even though the controller was used on ANYmal C. Note that, the domain adaptive policy was
        never trained using ANYmal C. Figure~\ref{fig:domain_adaptive_tracking_vel} represents
        the velocity tracking error observed in 1M simulation samples for randomly generated desired base velocity commands
        in the range of $[-0.3, 0.3]$\si{\meter\per\second} and $[-0.3, 0.3]$\si{\radian\per\second} for the linear and angular velocities respectively. Note that, in this
        case of domain adaptive tracking, we did observe an oscillatory behavior in the limbs while tracking the base velocity commands. This was largely due
        to the whole-body controller outputting desired joint states which were then aggressively corrected by the domain adaptive tracking policy. Moreover,
        the domain adaptive tracker failed to stably track absolute velocity commands beyond $\pm0.4$\si{\meter\per\second} for heading velocity, 
        $\pm0.3$\si{\meter\per\second} for lateral velocity, and $\pm0.5$\si{\radian\per\second} for yaw rate.
                
\begin{figure}
            \begin{subfigure}{.25\textwidth}
              \centering
              \includegraphics[width=.9\linewidth]{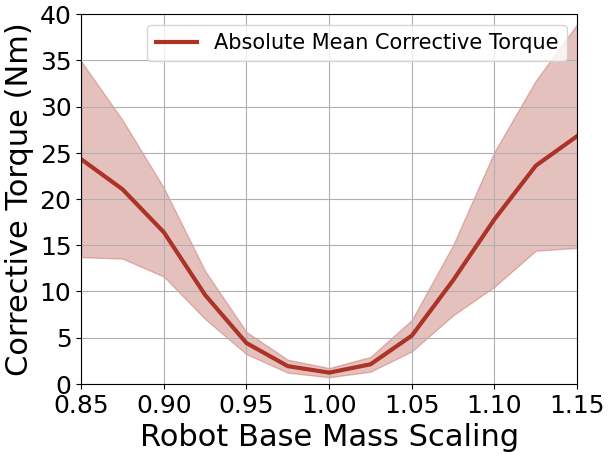}
            \end{subfigure}%
            \begin{subfigure}{.25\textwidth}
              \centering
              \includegraphics[width=.9\linewidth]{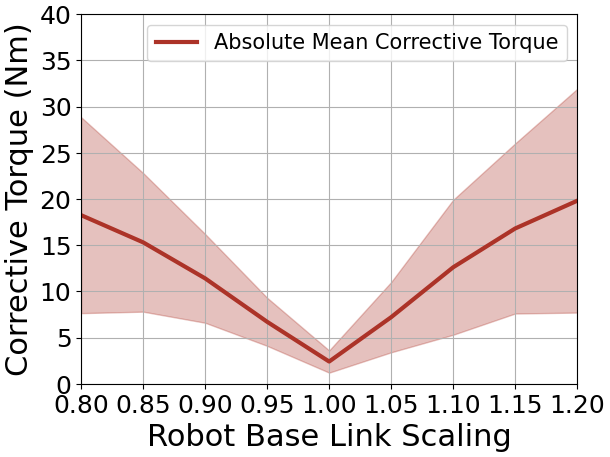}
            \end{subfigure}%
            \caption{The absolute mean corrective torques generated by the DAT to minimize motion plan tracking errors for ANYmal B base 
            mass (left) and link (right) scaling. The color bands represent one standard deviation.}
            \label{fig:adaptive_torque_scaling}
        \vspace{-12pt}
        \end{figure}

        Figure~\ref{fig:adaptive_torque_scaling} represents the mean and standard deviation of the absolute corrective joint torque generated by DAT
        to address ANYmal B base mass and link length scaling for locomotion on flat ground 
        for a heading velocity command of \SI{0.5}{\meter\per\second}.

        We have also provided a set of examples employing the domain adaptive tracker in \href{https://youtu.be/T-A8s-ajuww}{Movie S2}.
        
    \subsection{Recovery Controller}
    \label{sec:results_recovery_controller}
\begin{figure}
            \begin{subfigure}{.25\textwidth}
              \centering
              \includegraphics[width=.9\linewidth]{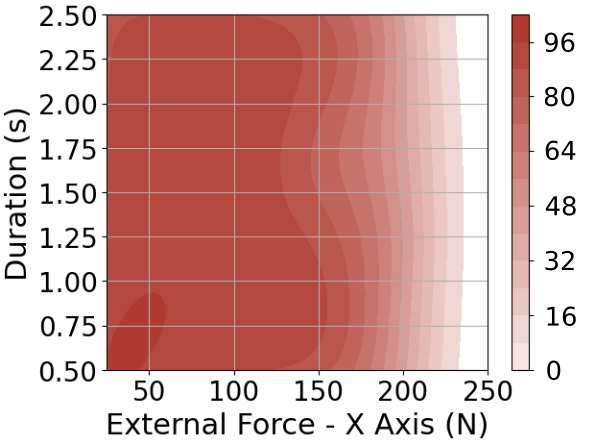}
              \caption{\footnotesize RL recovery controller -\\ANYmal B}
              \label{fig:rl_policy_heading}
            \end{subfigure}%
            \begin{subfigure}{.25\textwidth}
              \centering
              \includegraphics[width=.9\linewidth]{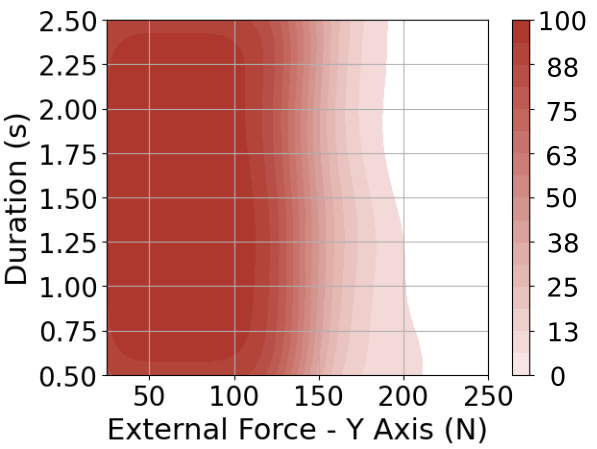}
              \caption{\footnotesize RL recovery controller -\\ANYmal B}
              \label{fig:rl_policy_lateral}
            \end{subfigure} \vspace{0.1cm} \\
            \begin{subfigure}{.25\textwidth}
              \centering
              \includegraphics[width=.9\linewidth]{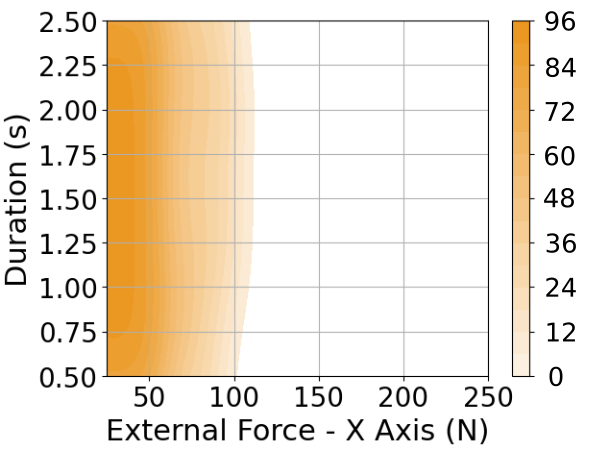}
              \caption{\footnotesize model-based controller only}
              \label{fig:whole_body_control_heading}
            \end{subfigure}%
            \begin{subfigure}{.25\textwidth}
              \centering
              \includegraphics[width=.9\linewidth]{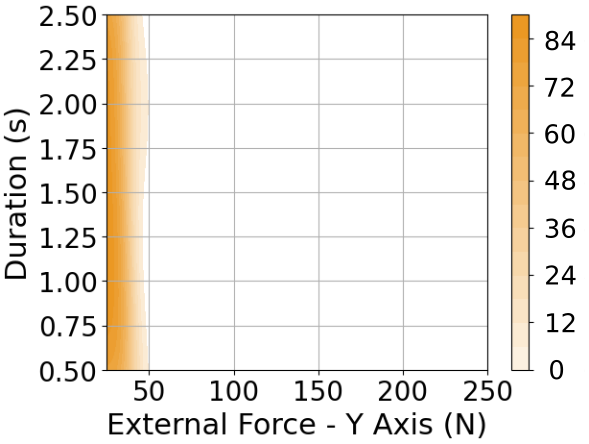}
              \caption{\footnotesize model-based controller only}
              \label{fig:whole_body_control_lateral}
            \end{subfigure} \vspace{0.1cm} \\
            \begin{subfigure}{.25\textwidth}
              \centering
              \includegraphics[width=.9\linewidth]{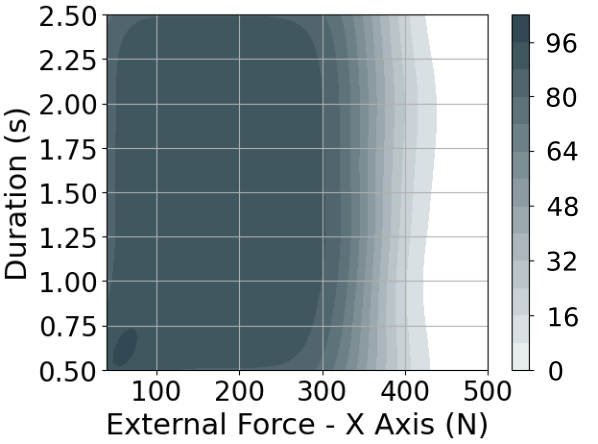}
              \caption{\footnotesize RL recovery controller -\\ANYmal C}
              \label{fig:rl_anymal_c_heading}
            \end{subfigure}%
            \begin{subfigure}{.25\textwidth}
              \centering
              \includegraphics[width=.9\linewidth]{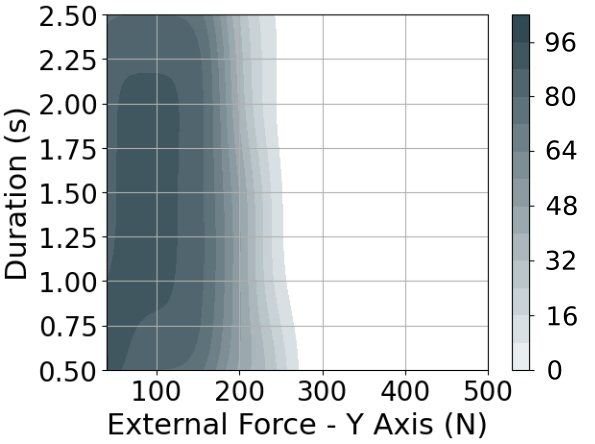}
              \caption{\footnotesize RL recovery controller -\\ANYmal C}
              \label{fig:rl_anymal_c_lateral}
            \end{subfigure} \\
            \caption{Bivariate kernel density estimates representing the success probability of the control policies to 
            stabilize upon external perturbations along the frontal (x) and the laeral (y) axes. Note that for \textbf{(e-f)}
            the x-axis scale of the plots is different.}
            \label{fig:perturbations}
        \end{figure}

        We tested the behavior of the RL recovery control policy and the motion controller for a range
        of external forces applied to the robot's base along the heading and lateral axes
        for duration between $[0.5, 2.5]\,$\si{\second}. The observations representing the KDE
        of the recovery rate as a function of the magnitude and duration of the external force 
        for each of the controllers are as shown in Fig.~\ref{fig:perturbations}. The color-bar labels represent
        the success probability of the robot to stabilize following external perturbations. We measure this 
        for both ANYmal B and ANYmal C and observe that compared to the motion controller which can handle
        perturbations of up to approximately \SI{100}{\newton} along the x-axis for the ANYmal B quadruped, the RL policy can handle
        external force of up to approximately \SI{225}{\newton} and \SI{400}{\newton} for ANYmal B and ANYmal C respectively.
        It is important to note that for the case of ANYmal C, we increased the position tracking gain to 60 compared to 
        45 for ANYmal B and also increased the joint torque limits to \SI{\pm80}{\newton\meter} compared to \SI{\pm40}{\newton\meter}
        for ANYmal B.
        The results obtained are also demonstrated in \href{https://youtu.be/-ToiFl-eBt8}{Movie S3}.
        
    \subsection{RLOC}
            \begin{figure}
            \begin{subfigure}{.25\textwidth}
              \centering
              \includegraphics[width=.9\linewidth]{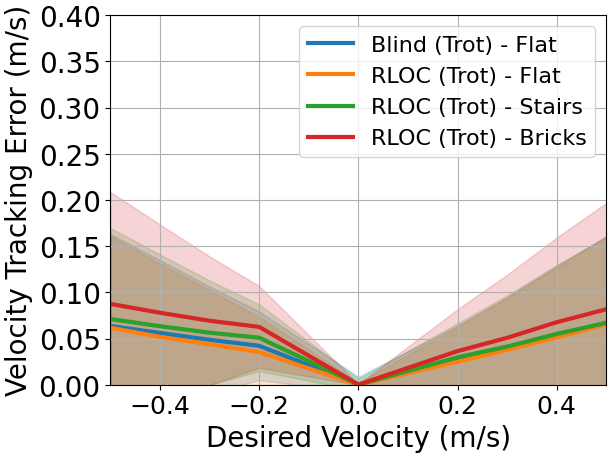}
              \caption{\footnotesize Forward Velocity Tracking}
              \label{fig:vel_tracking_heading}
            \end{subfigure}%
            \begin{subfigure}{.25\textwidth}
              \centering
              \includegraphics[width=.9\linewidth]{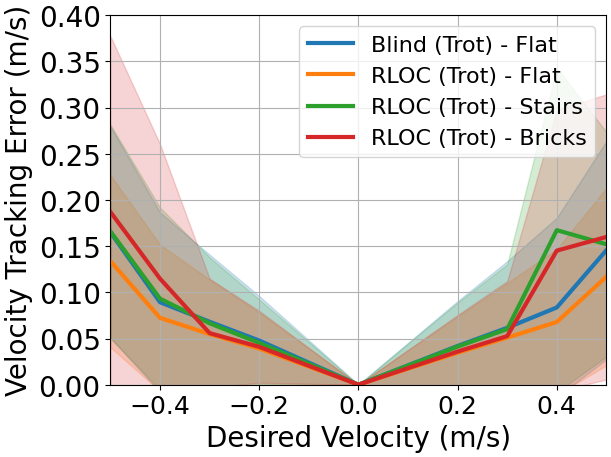}
              \caption{\footnotesize Lateral Velocity Tracking}
              \label{fig:vel_tracking_lateral}
            \end{subfigure}  \vspace{0.1cm} \\
            \begin{subfigure}{.25\textwidth}
              \centering
              \includegraphics[width=.9\linewidth]{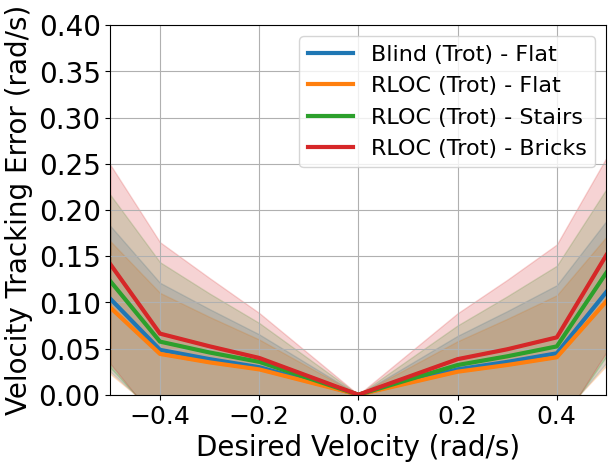}
              \caption{\footnotesize Yaw Rate Tracking}
              \label{fig:vel_tracking_yaw}
            \end{subfigure}%
            \begin{subfigure}{.25\textwidth}
              \centering
              \includegraphics[width=.9\linewidth]{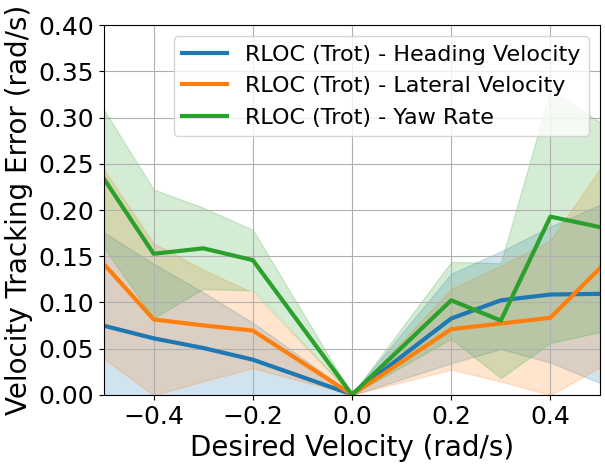}
              \caption{\footnotesize Velocity Tracking - Gazebo}
              \label{fig:vel_tracking_gazebo}
            \end{subfigure} \\
            \caption{\textbf{(a-c)} represent the velocity tracking error measured for 
            locomotion over even and uneven terrain for the ANYmal B quadruped tested in the RaiSim simulator. \textbf{(d)} represents
            the velocity tracking error on flat ground observed for the RLOC framework in the Gazebo simulator (with different actuation dynamics).}
            \label{fig:velocity_tracking}
        \end{figure}

        \begin{figure*}
              \centering
              \includegraphics[width=.98\linewidth]{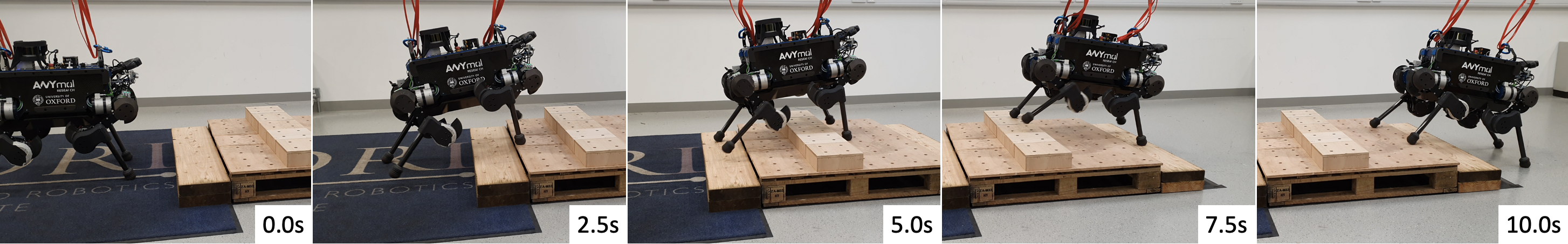}
              \caption{Snapshots of the ANYmal B quadruped walking over a staircase using our RLOC framework.}
              \label{fig:anymal_boxy_stairs_tests}
        \vspace{-12pt}
        \end{figure*}

        Figure~\ref{fig:velocity_tracking} represents
        the tracking error observed for the RLOC framework for traversal over different types of terrains. The improved
        tracking performance over the motion controller was a direct result of introduction of DAT.
        Removing DAT resulted in a very similar performance as that of the motion controller.
        
        The experimental setups for each of these terrains included the generation of a velocity
        command in the range of $[-0.5, 0.5]$\si{\meter/\second} for heading and lateral velocity tracking, and in the range of
        $[-0.5, 0.5]$\si{\radian/\second} for desired yaw rate tracking. Each episode comprised tracking of a 
        set velocity command for a duration of
        \SI{5}{\second}. We performed 100 runs for each command.

\begin{figure}
            \begin{subfigure}{.25\textwidth}
              \centering
              \includegraphics[width=.95\linewidth]{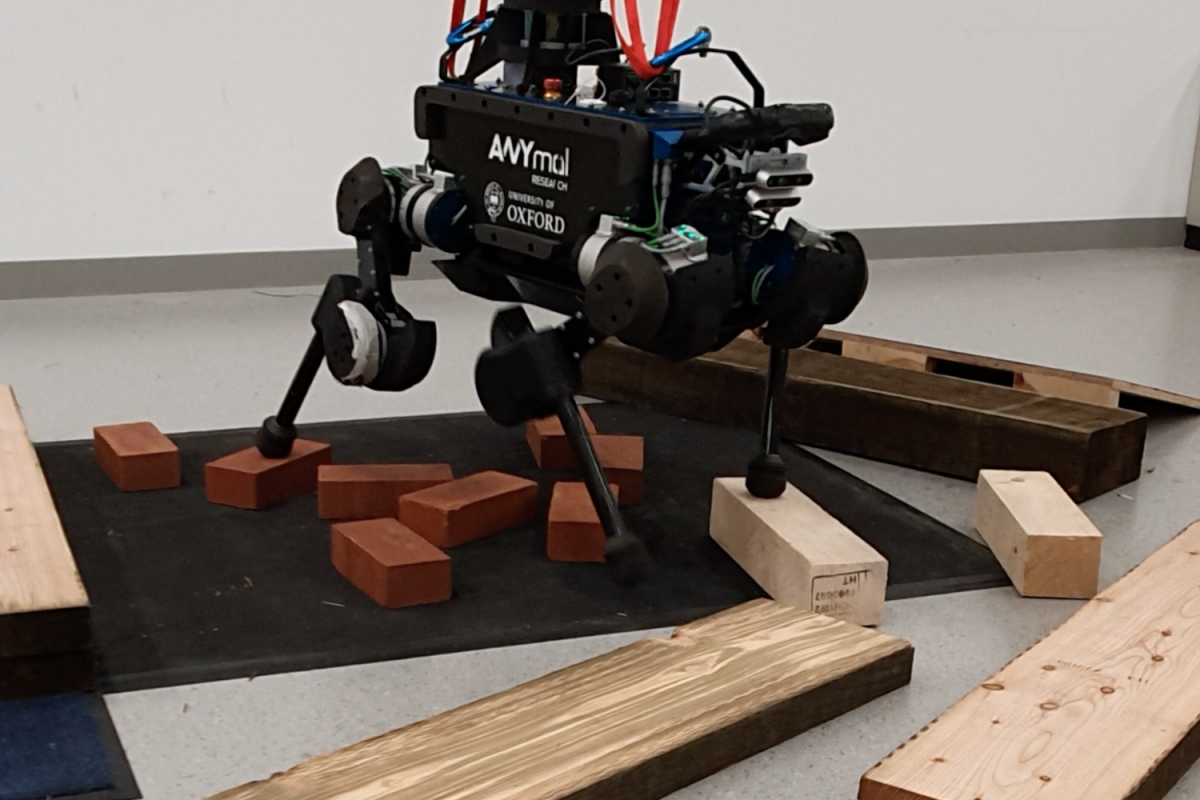}
              \label{fig:real_bricks_terrain}
            \end{subfigure}%
            \begin{subfigure}{.25\textwidth}
              \centering
              \includegraphics[width=.95\linewidth]{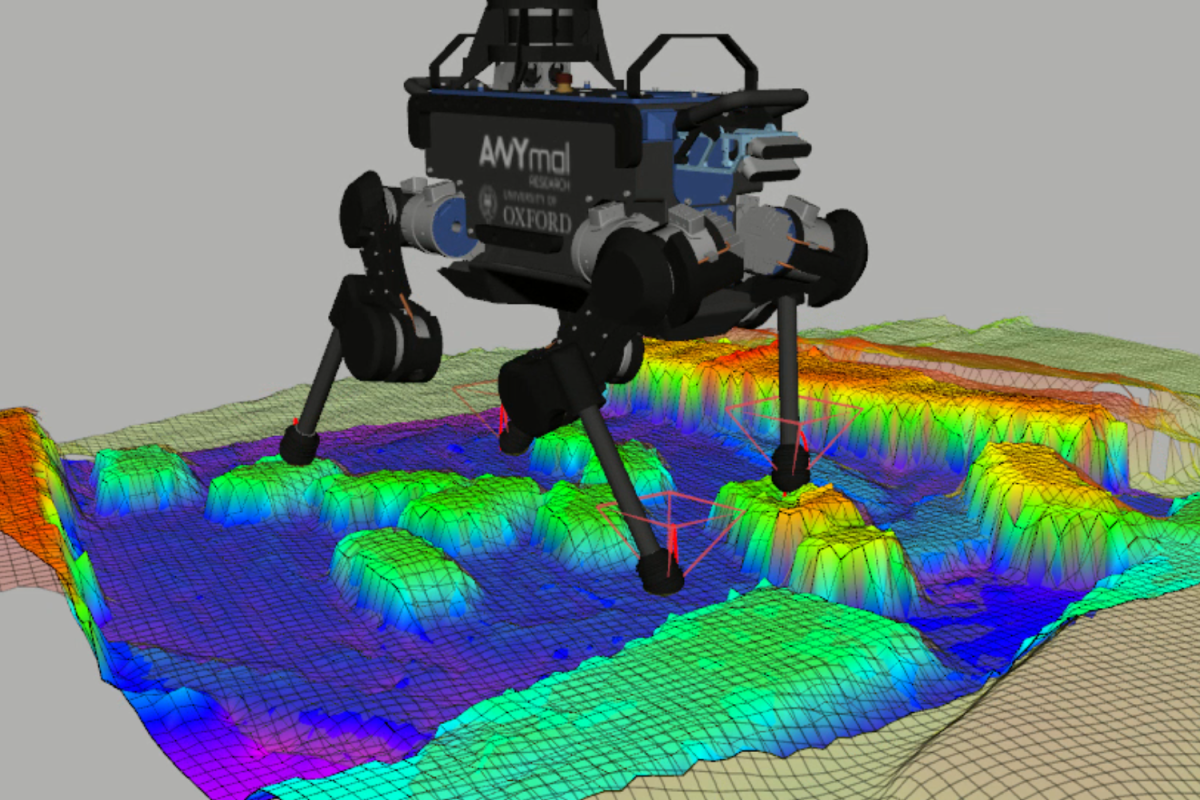}
              \label{fig:real_bricks_viz}
            \end{subfigure}  
            \caption{ANYmal B quadruped tested on a
             terrain comprising bricks (left); visualization of the corresponding robot state and elevation map (right).}
            \label{fig:real_bricks_elevation_smoothing}
        \vspace{-12pt}
        \end{figure}

        In order to quantify the effects of the recovery control policy, we performed experiments using the bricks and wave
        terrains for the trotting gait as introduced for evaluation of the FP. This is further presented in \href{https://youtu.be/-ToiFl-eBt8}{Movie S3}. Introduction of the recovery controller
        resulted in improved success rates. The results are summarized in
        Table~\ref{table:rloc_improvement}.
        \begin{center}
        \begin{table}[htbp]
            \caption{Improvement in success rate (SR) observed after introducing the recovery controller (RC) for traversal over bricks and wave terrains.
            The episodes here refer to the number of episodes (out of 1000) in which the RC was called. Moreover, the recovery rate (RR) represents the successful recovery
            performed using RC. The RR and SR are expressed as \%.}
            \resizebox{0.48\textwidth}{!}{
            \centering
            \begin{tabular}{ |c|c|c|c|c|c| }
                \hline
                 & \textbf{RC Calls} & \textbf{Episodes} & \textbf{RR} & \textbf{SR w/o RC} & \textbf{SR with RC} \\
                \hline
                \textbf{Bricks} & 184 & 131 & 62.5 & 85.3 & 90.2 \\
                \hline
                \textbf{Wave} & 39 & 22 & 71.8 & 93.0 & 94.8 \\
                \hline
            \end{tabular}}
            \label{table:rloc_improvement}
        \end{table}
        \end{center}
        We evaluated the velocity tracking performance of our framework in a slower but higher resolution \textit{Gazebo} simulator~\cite{koenig2004design}, 
        employing the ODE physics engine~\cite{smith2005open},
        as presented in Fig.~\ref{fig:vel_tracking_gazebo}. Figure~\ref{fig:anymal_boxy_real_tests} presents examples of the experiments performed on the real ANYmal B quadruped.
        The corresponding video clips are included in \href{https://youtu.be/GTI-0gl6Hg0}{Movie 1}. Figure~\ref{fig:anymal_boxy_stairs_tests} represents 
        the snapshots of the robot ascending and descending a staircase. In our experiments with the real robot, we observed motion behavior similar to
        the behavior in RaiSim especially for tests on simpler terrains such as a staircase. In the case of bricks, however, the
        state-estimation drift, sensor noise and the fast but less accurate nearest-neighbor interpolation technique used to obtain the elevation map along 
        occluded regions sometimes resulted in the elevation map being distorted especially along the edges, and also slightly shifted with respect to the robot. 
        The distortion of the elevation map is illustrated in Fig.~\ref{fig:elevation_map:occlusions}, Fig.~\ref{fig:real_bricks_elevation_smoothing} and \href{https://youtu.be/_SFOfxUgCzE}{Movie S4} which
        represent the real ANYmal B quadruped tested on a terrain comprising bricks, and the corresponding visualization of the 
        robot state and the elevation map. This blurring and slight shifting of the elevation map caused the robot to step close to
        the edges especially in environments composed of a large number of small objects.
        
        \begin{center}
        \begin{table}[htbp]
            \caption{Success rate obtained for traversal over different terrains using the ANYmal C quadruped comparing 
            the performance of RLOC with a 
            baseline perceptive controller~\cite{rudin2022learning}.}
            \resizebox{0.48\textwidth}{!}{
            \centering
            \begin{tabular}{ |c|c|c|c|c| }
                \hline
                 & \textbf{Unstructured} & \textbf{Stairs} & \textbf{Wave} & \textbf{Bricks} \\
                \hline
                \textbf{RLOC (Trot)} & \textbf{99.6} & 93.0 & \textbf{91.2} & \textbf{89.3} \\
                \hline
                \textbf{Baseline} & 99.4 & \textbf{93.2} & 83.5 & 70.9 \\
                \hline
            \end{tabular}}
            \label{table:rloc_vs_baseline}
        \end{table}
        \end{center}

        We compared the performance of RLOC on simulated ANYmal C with an open-sourced implementation of~\cite{rudin2022learning}
        as baseline. The
        success rates (for 1000 runs each) obtained for traversal over different terrains are provided in Table~\ref{table:rloc_vs_baseline}.
        While the two control frameworks performed similarly for locomotion over unstructured and stairs terrain, the baseline controller
        was unable to stably traverse the brick terrain. This is mostly due to the significantly larger terrain elevation sampling resolution 
        of \SI{0.1}{\meter} utilized in \cite{rudin2022learning} compared to \SI{0.02}{\meter} as used in RLOC. Reducing the sampling resolution
        for the baseline is expected to require significant retuning of the training environment and the policy architecture. 
        Although the baseline controller
        offered slightly better performance for traversing stairs, we observed that the controller was unable to reliably avoid edges
        while often staggering. In contrast, the motions exhibited by RLOC were a lot smoother. We suspect this behavioral comparison
        is also valid for \cite{miki2022learning}. This is attributed to the complex reward functions employed for training RLOC policies
        which encourage stable maneuvers (stability margin reward) while avoiding terrain edges (terrain cost map).
        
        We also transferred the RLOC framework to the real ANYmal C quadruped for both indoor and outdoor 
        locomotion as shown in Fig.~\ref{fig:anymal_coyote_real_tests}. We were 
        able to locomote outdoors over grassy and muddy terrain, and slopes. We observed a similar behavior
        as in the case of ANYmal B, however, the ANYmal C offered an advantage of higher joint torque limits enabling
        us to ascend steeper slopes. For this transfer, the only changes included the robot parameter
        description for the motion controller and the impedance controller gains. The experiments performed with the real ANYmal B and ANYmal C are
        presented in \href{https://youtu.be/GTI-0gl6Hg0}{Movie 1} and \href{https://youtu.be/rIr0tyqTjGw}{Movie S6}.

        We observed that ascending steep staircases where
        the ANYmal B base pitch exceeds $\pi/4\;\si{\radian}$ with respect to the horizontal frame results
        in failure to track heading velocity commands where the robot is unable to move
        forward. This is partially attributed to the joint torque
        limits while also resulting from the CoM motion planner that uses ZMP, implicitly assuming constant base height and attitude. For ANYmal C, we observe 
        velocity command tracking failure beyond a pitch angle of approximately $\pi/3\;\si{\radian}$. This 
        was the case for both simulated and real robot.

\section{Conclusion and Future Work}
 We presented an approach which bridges the gap between policies 
 trained using RL and controllers developed using model-based optimization methods.
 Our modular framework allowed us to greatly improve the success of traversal over complex terrains.
 For traversal over the bricks environment, 
 as presented in Section~\ref{sec:results_and_discussion}, we showed that our hybrid 
 approach, RLOC, was able to increase the success rate from 71.5\% (Table~\ref{table:footstep_planner}) to 90.2\% 
 (Table~\ref{table:rloc_improvement}) in comparison to our baseline perceptive controller. We showed that we were able to 
 traverse reliably (i.e. with a success rate of 90\% or above over 4m) and with a dynamic gait (trot at an effective velocity of
 approximately 0.25\si{\meter\per\second} on the real robot) over 
 uneven terrain even though the model-based locomotion controller was primarily developed and tuned for quasi-flat terrains.

To address ascending steep stairs and slopes, in our future work we aim to learn
a CoM and foot motion planner. Another direction for future work includes obtaining a hybrid long and short horizon planning 
policy to avoid obstacles and perform CoM and feet motion 
adaptations for traversal over extremely complex and non-stationary terrains.
    
\section*{Nomenclature}
\newcommand{\tab}[1]{\hspace{.05\textwidth}\rlap{#1}}

\begin{table}[h!]
    \centering
        \begin{tabular}{|c|c|}
            \hline
            Notation & Definition \rule{0pt}{2.6ex} \rule[-0.9ex]{0pt}{0pt} \\
            \hline
            ${}_{W}r_{WB}$ or $r_{B}$ & translation of frame $B$ represented frame $W$ \rule{0pt}{2.6ex} \\
            $\mathbf{R}_{WB}$ or $\mathbf{R}_B$ & rotation matrix of $B$ represented in $W$    \rule{0pt}{2.6ex} \\
            $\mathbf{e}_z^W$ & $z$-axis of $W$ \rule{0pt}{2.6ex} \\
            $\mathrm{q}_j$ & joint positions \rule{0pt}{2.6ex} \\
            ${}_{W}\mathrm{v}_{WB}$ or $\mathrm{v}_{B}$ & linear velocity of $B$ measured and expressed in $W$ \rule{0pt}{2.6ex} \\
            ${}_{W}\mathrm{\omega}_{WB}$ or $\mathrm{\omega}_{B}$ & angular velocity of $B$ measured and expressed in $W$ \rule{0pt}{2.6ex} \\
            $\dot{\mathrm{q}}_j$ & joint velocities \rule{0pt}{2.6ex} \\
            $\mathbf{R}_{B_z}$ & $z$ decomposition of $\mathbf{R}_B$ \rule{0pt}{2.6ex} \\
            $\tau_j$ & joint torques \rule{0pt}{2.6ex} \\
            $\cdot^\ast$ & desired quantity \rule{0pt}{2.6ex} \\
            $\Phi$ & feet motion parameterization tuple \rule{0pt}{2.6ex} \\
            $r_F^l$, $r_F^d$, $r_F^h$ & feet lift-off, touch-down and clearance positions \rule{0pt}{2.6ex} \\
            $\mathrm{c}_F$ & feet contact state \rule{0pt}{2.6ex} \\
            $t_F^l$, $t_F^d$ & feet lift-off and touch-down times \rule{0pt}{2.6ex} \\
            $\mathbf{M}_H$ & robocentric terrain elevation represented in frame $H$  \rule{0pt}{2.6ex} \\
            $\mathrm{Z}_{\mathbf{M}_{H}}$ & latent embedding of $\mathbf{M}_H$  \rule{0pt}{2.6ex} \\
            $\mathcal{C}$ & 2d convolution operator  \rule{0pt}{2.6ex} \\
            $\mathbf{s}_f$, $\mathbf{a}_f$, $\mathbf{r}_F$ & footstep planner state, action and reward  \rule{0pt}{2.6ex} \\
            $\mathbf{s}_d$, $\mathbf{a}_d$, $\mathbf{r}_D$ & domain adaptive tracker state, action and reward  \rule{0pt}{2.6ex} \\
            $\mathbf{s}_r$, $\mathbf{a}_r$, $\mathbf{r}_R$ & recovery controller state, action and reward  \rule{0pt}{2.6ex} \\
            $\mathbf{c}^\ast$ & desired velocity command  \rule{0pt}{3.6ex} \\
            \makecell{$\pi_{\theta_f}, \pi_{\theta_d}, \pi_{\theta_r}$} & \makecell{footstep planning, domain adaptive tracking \\ and recovery control policies}
            \rule{0pt}{3.6ex} \\
            $\mathcal{S}_m$ & stability margin function \rule{0pt}{2.6ex} \\
            $\mathbf{C}_\mathbf{M}$ & terrain cost map operator \rule{0pt}{2.6ex} \\
            $B_{\mathbf{M}_H}$ & elevation map update frame \rule{0pt}{2.6ex} \\
                     \hline 
        \end{tabular}
\label{tab:nomenclature}
\end{table}

\section*{Appendix} 
\label{sec:supplementary_materal}

\subsection{Motion Controller}
\label{section:appendix_motion_controller}
The dynamic gaits motion controller~\cite{bellicoso2018dynamic} generates the CoM motion plan by 
solving a nonlinear optimization problem using the sequential quadratic programming (SQP)
        approach~\cite{nocedal2006numerical} for a time horizon $t_H^c$ which corresponds to the periodicity of
        the active locomotion gait. 
        The SQP optimization parameters correspond to the spline coefficients 
        $\alpha=[{\alpha_0}^T \cdot \cdot \cdot {\alpha_i}^T \cdot \cdot \cdot {\alpha_{n_s}}^T]$ where $n_s$ is the
        number of different support polygons on the optimization horizon. This formulation, where the SQP also optimizes for ${\alpha_i^z}$ as opposed to
        assuming a fixed base height to ease on the computational complexity, 
        enables us to adapt the dynamic gaits motion controller for locomotion over 
        uneven terrain.
        For stability, dynamic gaits uses constraints on the ZMP.
        It also penalizes deviations from
        a \textit{path regularizer}, which represents
        an approximated CoM motion plan for a duration of $t_H^c$, to limit the drift of the robot base
        w.r.t. the reference footholds
        which may occur due to external perturbations and continuous updates of the motion plan. The path 
        regularizer, $\mathcal{T}_R^c$, is computed based on the estimated footprint center and the reference base
        velocity command, $\mathbf{c}^\ast$.
        We use a similar approach as presented in~\cite{bellicoso2018dynamic} to compute $\mathcal{T}_R^c$ and 
        introduce the following adaptations in the CoM motion plan optimization problem:
            (1) the terrain plane
            is estimated by fitting a planar surface through the current and planned footholds as opposed to current and previous footholds
            (2) unlike dynamic gaits which sets the height of the path regularizer $\mathcal{T}_{R_z}^c$ at a fixed reference value $h_{ref}$ with respect to the control frame, we adapt this value such that:
            \begin{equation*}
                \mathcal{T}_{R_z}^c = \mathbf{M}_{H}(\mathcal{T}_{R_x}^c, \mathcal{T}_{R_y}^c) + h_{ref} - 0.125 |\mathbf{R}_{C_y}|
            \end{equation*}
            where $\mathbf{R}_{C_y}$ is the angle of the terrain plane on the pitch axis of the robot. This modification allows us
            to increase the reachability of the feet on stairs and complex terrains.

        The support polygons (used in the dynamic constraints of the optimization) are defined by separate modules, the 
        foothold planner and contact scheduler.

        The foothold planning module in dynamic gaits is set up as a
        quadratic programming (QP) problem to obtain the $x$ and $y$ components of the
        desired footholds ${r^\ast_{F_{xy}}}$. The nominal footholds expressed in the
        horizontal reference frame ${{}_{H}r^n_{HF_{xy}}}$ corresponds to the projections of the 
        robot hip locations measured and expressed 
        in the horizontal frame along the vertical axis ${\mathbf{e}_z^H}$. The cost function of the
        QP problem then includes, among others, 
        a penalty term for deviations from the nominal foothold locations. The foothold
        optimizer gives us the desired feet touch down positions, ${r_F^d}^\ast$, where ${r^d_{F_{z}}}^\ast$
        is extracted from the terrain plane estimate, for each swing phase.
        
        The contact
        scheduler module, manages the feet contact-events for a full
        stride of
        an active gait while also supporting smooth gait transitions giving us the lift-off,
        $t_F^l$, and touch-down, $t_F^d$, timings for the feet swing and stance phases based on the
        assumption that $v_F^l$ and $v_F^d$ are zeros. This module also enables slowing down the execution of the gait in case a touch-down event is not detected at the expected time. The $\Phi_{o}$ and 
        $\Phi_{c}$ thus obtained is used for generating motion plans for ${r_F^d}^\ast$. Moreover, dynamic
        gaits considers a fixed foot clearance position ${r_F^h}$ for $\Phi_{o}$.
        
        In our RLOC framework, the foothold optimizer is replaced by the RL footstep planner which
        generates the desired feet touch-down positions, ${r_F^{FP}}$, for locomotion over
        uneven terrain. Additionally, ${r_{F,i}^h}$ for swing phase $\Phi_{o,i}$ of foot $i$ is now computed
        based on the maximum terrain elevation observed between the line, $l_{ld}^\Phi$, 
        connecting $r_{F_{xy}}^{l}$
        and ${r_{F_{xy}}^{FP}}$, given by
        \begin{equation}
            r_{{F_z},i}^h = \mathrm{max}(\mathbf{M}_{H}(X,Y)) + 0.05\,\si{\meter}
        \end{equation} where $\mathbf{M}_{H}$ is the elevation map expressed in the
        global reference frame, $X$ and $Y$ are the vectors containing the $x$ and $y$ coordinates
        of $l_{ld}^\Phi$ sampled at a resolution of \SI{0.02}{\meter}. The additional
        offset of \SI{0.05}{\meter} to the foot clearance position is introduced to avoid collisions during
        inaccurate motion tracking. To limit uncertain behavior that may occur due to 
        imprecise elevation map generation as a result of sensor
        noise and occlusions, $r^h_F$ is clipped to be less than $r^l_F + 0.25\,$\si{\meter}.
        
        In contrast to dynamic gaits, which uses the estimate of the terrain plane to determine the orientation of the friction cone, we compute the terrain normal at each foot using the information from the elevation map. The normals are estimated using central finite difference at the feet positions. Since the information from the elevation map is not always reliable, we also add the following checks:
         \begin{enumerate}
             \item When $\mathrm{c}_{F,i}=1$, the foot height is close to the terrain elevation such that $| r^d_{{F}_z,i} - \mathbf{M}_H(r^d_{{F_{xy},i}}) | < \epsilon_d$ where $\epsilon_d = \SI{0.04}{\meter}$.
             \item The terrain elevation within a deviation of $\pm\SI{4}{\centi\meter}$ centered at $r^d_{F_{xy},i}$ is planar.
             \item The estimated terrain normal at foot $i$, $\mathbf{e}_z^{T_i}$, stays in realistic bounds, i.e. $| \mathnormal{cos}^{-1}(\mathbf{e}_z^{T_i}\cdot\mathbf{e}_z^W) | < \pi/3\;\si{\radian}$.
         \end{enumerate}
         If any of these checks fail for a foot $i$, we fall back to a vertical estimate for the corresponding friction cone.
        
        In the original work, the CoM and foothold motion optimizers are run every time a new
        solution is obtained. In our training setup, however, we use a fixed optimization frequency of \SI{50}{\hertz} 
        which corresponds to a conservative lower bound of the planning frequency on the hardware. This
        allows us to compute plans at a frequency independent of the computation time enabling 
        execution of parallel simulations 
        during training of our RL policies.
        For deployment,
        we switch back to the original variable optimization frequency approach.
        Moreover,
        during evaluation, we remove the foothold optimization module from the motion controller and
        only utilize the RL footstep planner executed after every gait stride.
        
\subsection{Stability Margin}
\label{section:appendix_stability}
        In our work, we define the stability margin as the distance between the Instantaneous Capture Point 
        (ICP)~\cite{Pratt2007}
        and the edges of the feasible region as introduced in~\cite{Orsolino2020}. 
        The margin is positive when the ICP belongs to the feasible region and negative when it lies outside of it.
        A positive margin corresponds to a stable robot state whereas a negative margin implies that the robot is unstable and will 
        eventually fall unless 
        a recovery action is introduced. If the margin is bounded, however,
        it implies that the ICP does not diverge away from the ZMP and that the robot 
        can still manage to recover to a stable configuration before the robot eventually falls. 
        This is, for example, what happens for dynamic gait patterns such as trot and pace:
        in such cases the stability margin will be mostly negative 
        but the robot will not fall if it manages to maintain a suitable lower bound value of stability margin at all the time.

        The stability margin term, introduced in the RL reward functions, ensures the robot avoids
        unnecessary linear and angular accelerations and places its feet at the correct time instant and location. 
        A maximization of the stability margin over a crawl gait
        (only one swing foot at the time),
        for example, will result in the policy trying to 
        enlarge the support of the robot as much as possible corresponding to a larger feasible region.
        
        The computation of the stability margin requires the estimation of the feasible region, which can be done by means of an iterative projection algorithm up to an arbitrary tolerance~\cite{Bretl2008}. This procedure requires the solution of multiple linear programs, and even for low tolerance values, requires an estimated \SI{0.5}{\milli\second}
        for an optimized C++ implementation. For this reason we employ an MLP as an
        approximation of the analytical solution to perform extremely fast 
        stability estimation. In our C++ implementation, we can perform a forward pass through this MLP in less than \SI{4}{\micro\second} making it appealing for use in sample-inefficient RL training.

        We use an MLP with 2 hidden layers of size \{48, 48\} which maps a 47-dimensional input into a scalar stability margin. The network input, $\mathbf{s}_m$, is given by the tuple
        \begin{equation*}
         \mathbf{s}_m = \langle \mathbf{e}^B_z,
         {}_{B}{v}_{WB},
         {}_{B}{\omega}_{WB},\\
        {}_{B}\dot{\omega}_{WB},\\
        \overrightarrow{F}_B,
        \overrightarrow{\tau}_B,
        \mathbf{T}_\mu,
        {}_{B}r_{BF},\;
        \mathrm{c}_F,\;
        {N}_F\rangle
        \end{equation*} where
        $\overrightarrow{F}_B\in\mathcal{R}^3$ is the external force on base,
        $\overrightarrow{\tau}_B\in\mathcal{R}^3$ is the external torque applied on base, $\mathbf{T}_\mu\in\mathcal{R}$ is the terrain friction,
        and 
        $N_{F}$ represents the contact normals.
        
        Upon training the stability margin network, we observed an absolute prediction error mean of \SI{0.0051}{\meter} and a standard deviation of \SI{0.0018}{\meter} when
        tested on a dataset comprising of 50k test samples. 
        
In addition to efficiently obtaining an estimate of the stability margin, the use of a neural network, also enables computation of partial derivatives of the stability margin with respect to the robot's states thereby eliminating the need to compute the Jacobians by finite differences of the analytical solution.

\begin{figure}[!h]
  \includegraphics[width=\linewidth]{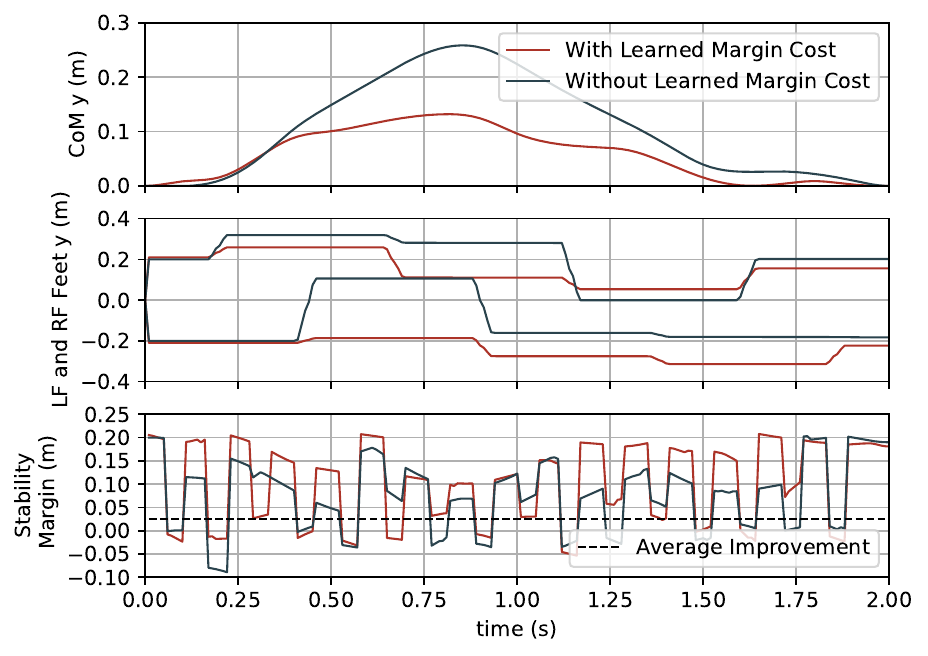}
  \caption{Trajectories referring to two motion plans of Anymal B on flat terrain optimized using TOWR with (blue) and without (red) the cost on the learned stability margin. We can see that maximizing the learned stability margin leads to a decrease in unnecessary base motions (upper plot) and in a larger stance (mid plot). These two elements together lead to an average improvement of the stability margin of about 2.5 cm (lower plot).}
  \label{fig:improvement}
\end{figure}

        This provides the possibility to formulate optimal control problems and gradient-descent based trajectory optimization problems that exploit the Jacobian with respect to the optimization variables of costs and constraints based on this MLP in order to maximize the overall stability margin or to make sure that this quantity lies within some specified boundaries. Fig.~\ref{fig:improvement} represents the values of the stability margin obtained for a straight walk of Anymal B on a flat terrain for \SI{2}{\second} covering a distance of \SI{0.5}{\meter}. Both the motions were obtained using the TOWR~\cite{Winkler2018} library, which employs a minimal parameterization of the state of the robot based on the single rigid body dynamics model. The red values refer to a motion plan obtained with the baseline library, which attempts to obtain a solution that satisfies all the feasibility constraints such as kinematic reachability and dynamic consistency without any minimization cost. The blue curve, instead, refers to the same trajectory optimization problem where a cost has been added that maximizes the learned stability margin. As an effect of the added cost, the obtained solution presents smaller acceleration peaks and a larger stance configuration of the feet. These two elements together result in an increased value of the stability margin both during the triple and the quadruple stance phases and in an average improvement of the stability margin of about \SI{0.025}{\meter}. Although the derivatives
        of the stability margin network represent an approximation of the Jacobians, we observed that this approximation
        still remained valid along the state regions sampled by TOWR, allowing us to obtain a feasible trajectory while 
        maximizing the stability margin.
        
        We observed a similar behavior while training our RL footstep planning policy. For a higher stability margin reward coefficient the footstep planning policy preferred a wider stance behavior. Upon increasing the reward coefficient for stepping close to the nominal feet position and lowering the stability margin coefficient,
        the footstep planning policy preferred a wider stance behavior only during unstable motion phases. This is demonstrated in \href{https://youtu.be/vSOU-USXK64}{Movie S5}.

    In our preliminary experiments, we observed that introducing the stability margin reward term in an environment setup with a simple reward
    function results in reduced base accelerations. As shown in \href{https://youtu.be/vSOU-USXK64}{Movie S5}, a locomotion policy obtained using the reward function
    $-(0.7 - {}_{B}v_{{WB}_x})^2$, is significantly more aggressive compared to the reward
    function augmented using the stability margin. This reward function can further be tuned to obtain the desired locomotion behavior. In our RL environment setups, we use the stability margin in our reward function to encourage the RL 
    agents to perform less aggressive maneuvers in order
    to maximize the stability metric.
    
    We extended this work on rapid dynamic stability estimation for use with OC and RL based approaches.
    The detailed network 
    training and deployment description
    can be found in~\cite{orsolinorapid}.

\subsection{Cost Map}
\label{section:appendix_costmap}
        In order to drive the RL footstep planner to generate desired feet positions which avoid terrain edges, we compute a penalty based on the cost map obtained using the
        1\textsuperscript{st} and 2\textsuperscript{nd} order derivatives of the elevation map local to the feet positions 
        within a deviation of $\pm\SI{0.065}{\meter}$ along $\mathbf{e}_x^{F}$ and $\mathbf{e}_y^{F}$ represented by $\mathbf{M}_{F_i}:\mathbb{R}^2\times\mathbb{R}\rightarrow\mathbb{R}^{27\times27}$ for foot $i$ sampled at
        a resolution of $\SI{0.005}{\meter}$.
        We compute the cost map by performing operations in the following order:
        \begin{enumerate}
            \item Smoothing: $\mathbf{C}_\mathbf{M}(r_{F_{xy},i})\leftarrow\mathcal{C}(\mathbf{M}_{F_i}, k_s)$ where
                \begin{equation*}
                        k_{s} = 
                        \begin{bmatrix}
                            0.1 & 0.1 & 0.1\\
                            0.1 & 0.2 & 0.1\\
                            0.1 & 0.1 & 0.1\\
                        \end{bmatrix}
                \end{equation*}
            \item Approximation of second order derivative using Laplacian kernel: $\mathbf{C}_\mathbf{M}(r_{F_{xy},i})\leftarrow\mathcal{C}(\mathbf{C}_\mathbf{M}(r_{F_{xy},i}), k_l)$ where
                \begin{equation*}
                        k_{l} = 
                        \begin{bmatrix}
                            -1 & -1 & -1\\
                            -1 & 8 & -1\\
                            -1 & -1 & -1\\
                        \end{bmatrix}
                \end{equation*}
            \item Modulus of the cost map to represent terrain edges without direction information: $\mathbf{C}_\mathbf{M}(r_{F_{xy},i})\leftarrow abs(\mathbf{C}_\mathbf{M}(r_{F_{xy},i}))$
            \item Estimation of deviation in heights along the edges using blurring kernel:
            $\mathbf{C}_\mathbf{M}(r_{F_{xy},i})\leftarrow\mathcal{C}(\mathbf{C}_\mathbf{M}(r_{F_{xy},i}), k_b)$
                    \begin{equation*}
                        k_{b} = \frac{1}{9}\times
                        \begin{bmatrix}
                            1 & 1 & 1\\
                            1 & 1 & 1\\
                            1 & 1 & 1\\
                        \end{bmatrix}
                \end{equation*}
            \item Circular distance filtering such that terrain edge at
            $\mathbf{M}_{F_i}(r_{F_{xy},i})$ has the maximum weighting for computing the cost
            penalty. This weight decreases with increase in the radial deviation from the center of the feet $r_{F_{xy},i}$.
            The circular distance filter can be represented as a matrix $\mathbf{M}_{c}\in\mathbb{R}^{21\times21}$
            where the center $\mathbf{M}_{c}(r_{F_{xy},i})=1$ and 
            $\mathbf{M}_{c}(\hat{r}_{F_{xy},i})=max(0.05 - ||\hat{r}_{F_{xy},i} - r_{F_{xy},i}||, 0) / 0.05$. The cost
            map is then modified by performing an element-wise product:
            $\mathbf{C}_\mathbf{M}(r_{F_{xy},i})\leftarrow{M}_{c}\odot\mathbf{C}_\mathbf{M}(r_{F_{xy},i})$
        \end{enumerate}
        
        Note that, we start with $\mathbf{M}_{F_i}:\mathbb{R}^2\times\mathbb{R}\rightarrow\mathbb{R}^{27\times27}$ which corresponds to 
        an area of $(2\times0.065)\times(2\times0.065)\,$\si{\meter\squared}, and following the above procedure, we obtain the desired $21\times21$-dimensional cost map
        which corresponds to an area of $(2\times0.05)\times(2\times0.05)\,$\si{\meter\squared}. This reduction
        is due to the convolutions performed without utilizing additional padding.

\begin{figure}
            \begin{subfigure}{.25\textwidth}
              \centering
              \includegraphics[width=.9\linewidth]{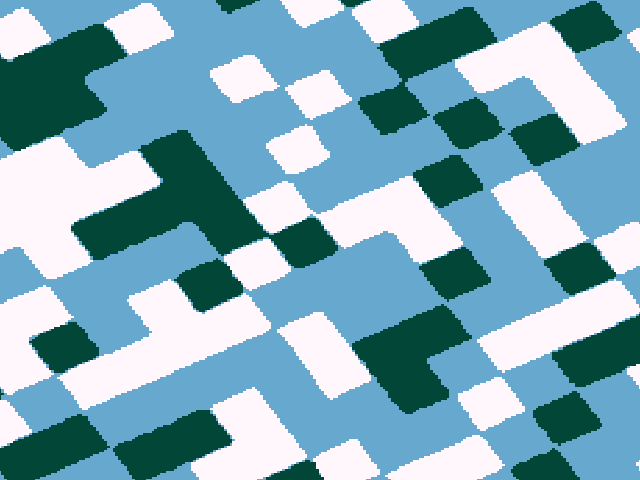}
            \caption{\footnotesize Terrain elevation}

              \label{fig:cost_map:input}
            \end{subfigure}%
            \begin{subfigure}{.25\textwidth}
              \centering
              \includegraphics[width=.9\linewidth]{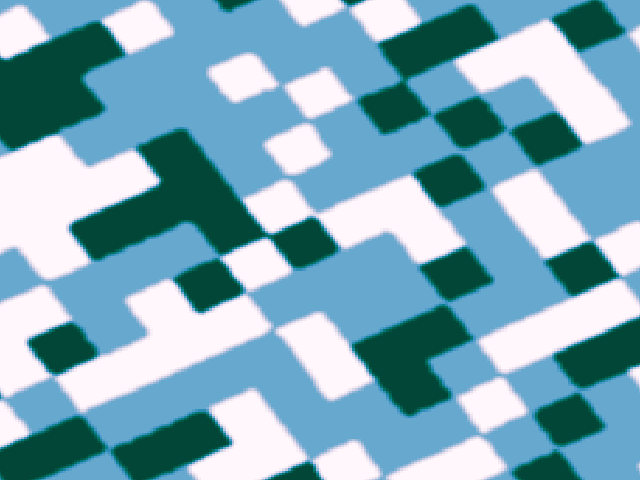}
               \caption{\footnotesize Elevation smoothing}
              \label{fig:cost_map:smooth}
            \end{subfigure} \vspace{0.125cm} \\
                        \begin{subfigure}{.25\textwidth}
              \centering
              \includegraphics[width=.9\linewidth]{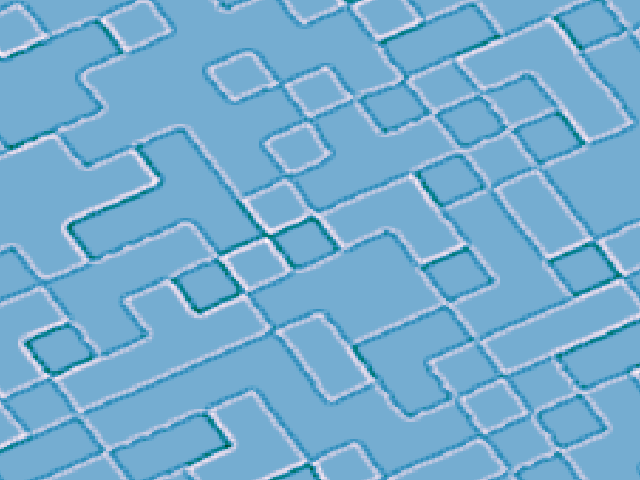}
               \caption{\footnotesize Elevation Laplacian}
              \label{fig:cost_map:laplacian}
            \end{subfigure}%
            \begin{subfigure}{.25\textwidth}
              \centering
              \includegraphics[width=.9\linewidth]{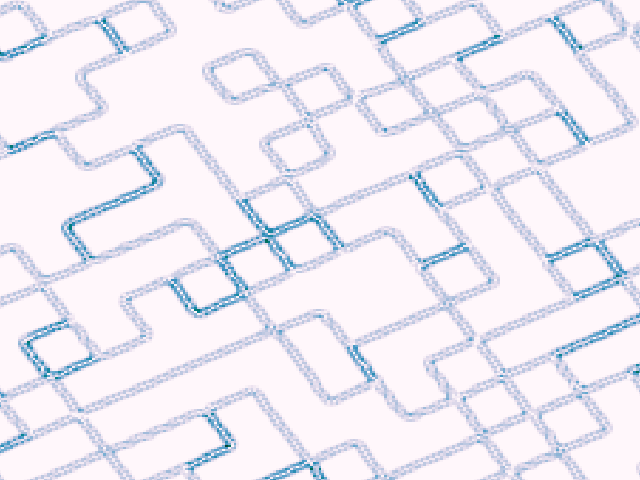}
               \caption{\footnotesize Elevation modulus}
              \label{fig:cost_map:slope}
            \end{subfigure} \vspace{0.125cm} \\
                        \begin{subfigure}{.25\textwidth}
              \centering
              \includegraphics[width=.9\linewidth]{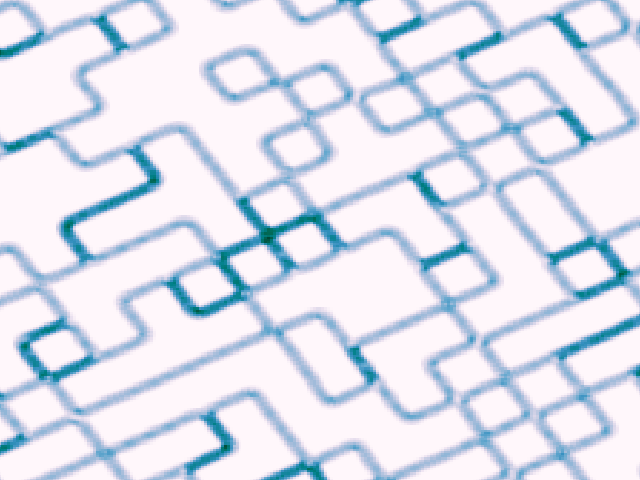}
               \caption{\footnotesize Elevation blurring}
              \label{fig:cost_map:blurred}
            \end{subfigure}%
            \begin{subfigure}{.25\textwidth}
              \centering
              \includegraphics[width=.9\linewidth]{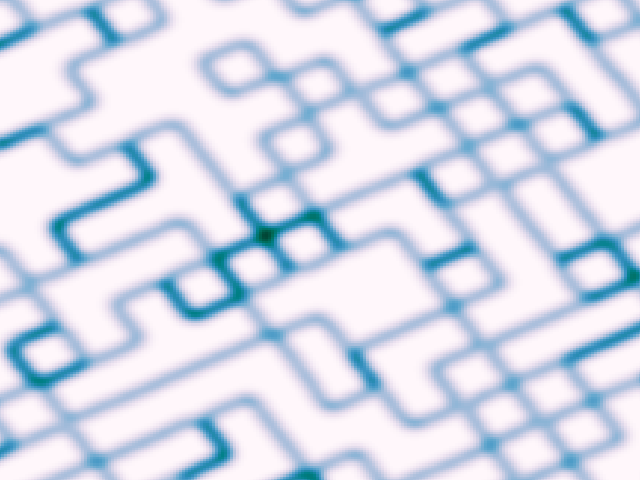}
             \caption{\footnotesize Terrain cost map}
              \label{fig:cost_map:output}
            \end{subfigure} \\
            \caption{Example of computation of the cost map for a bricks terrain. Green color represents upward extrusions corresponding to higher elevation while white color represents downward extrusions. For the cost map, dark blue represents positions with higher cost. Note that, the cost map shown here corresponds to the entire terrain which
            was obtained by performing convolution using the circular distance filter as opposed to element-wise multiplication which is done for extracting the cost at foot position in $\mathbf{M}_{F_i}$.}
            \label{fig:cost_map}
        \end{figure}

        Figure~\ref{fig:cost_map} shows an example of the cost map on a terrains with bricks and holes. In practice, the whole cost map does not need to be explicitly computed since only the cost at each foot position is needed so the transformation are only applied on small patches around the selected footholds.

\subsection{Elevation Map Encoding}
Our encoder network architecture design is based on the following reasons: (1) the encoder should capture as much information as possible, 
(2) the encoder forward pass time on the real robots should not exceed the control step time of \SI{2.5}{\milli\second} after the raw elevation map is computed in
a parallel thread to avoid change in robot state resulting in feet position tracking drifts. 
For this, we introduced the unstructured terrain in the dataset for
training the autoencoder. We observed that reducing the dimensionality
of the latent representation below 96 resulted in poor reconstruction of
the unstructured terrain, i.e. the encoder failed to detect
the unstructured terrain, and the autoencoder output an almost smooth
reconstruction.
A forward pass through the encoder network requires approximately \SI{1.7}{\milli\second}
on the real robots giving us sufficient time to perform state updates and forward passes through
the RL planning and control policy networks.

While training the encoder using the autoencoding strategy, we minimize the
loss
\begin{equation}
                \mathcal{L}_{\mathbf{M}_{H}} = ({\mathbf{M}}_H^n - {\mathbf{\hat{M}}}_H^n)^2 \,.
            \end{equation} It is a common practice to include a regularization term in this reconstruction
            loss to discourage overfitting, however, reconstruction from noisy encoder input implies that
            the auto-encoder is restricted from memorizing the training dataset. Instead, the encoder learns to
            map the input data to a lower-dimensional manifold which represents the most relevant features
            of the terrain while getting rid of the additional noise. Hence, it is not necessary for us
            to introduce additional regularization during training.

\subsection{Control Switching}
RLOC executes the footstep planner, motion controller and domain adaptive tracker
upon initialization. When any of the
robot state parameters exceeds a certain threshold described by the activation limit in Table~\ref{tab:recovery_activation_deactivation},
the recovery controller is activated while the footstep planner, motion controller and domain adaptive tracker
are deactivated. The recovery controller stabilizes the robot and the active control state
switches back to motion planning and
whole-body control. This switch is performed when each of the robot state parameters
lie within the deactivation limits presented in Table~\ref{tab:recovery_activation_deactivation}.

\begin{table}[h!]
                \centering
                \caption{Robot state criteria for activating and deactivating the recovery controller. Here $|\cdot|$
                refers to the $l_1$ norm of a vector, $\mathrm{abs}()$ is the element-wise modulus operator, 
                $\mathrm{max}()$ refers to the maximum element in a vector, $\mathrm{q}_j^n$ is the vector of nominal 
                joint positions and $(\mathrm{q}_{j} - \mathrm{q}^n_{j})_{HAA}$ refers to the vector of 
                elements relating to the four HAA (hip abduction-adduction) joints. HFE refers to
                hip flexion-extension and KFE refers to knee flexion-extension.}
                \begin{tabular}{|c|c|c|}
                \hline
                     State & Activation Limit & Deactivation Limit \\
                     \hline
                     $| \mathrm{v}_B |$ & 20.0 & 0.5 \rule{0pt}{2.6ex} \\
                     $| \mathrm{\omega}_B |$ & 5.0 & 0.5 \rule{0pt}{2.6ex} \\
                     $\mathrm{max}(\mathrm{abs}((\mathrm{q}_{j} - \mathrm{q}^n_{j})_{HAA}))$ & $\pi/4$ & $\pi/12$ 
                     \rule{0pt}{2.6ex} \\
                     $\mathrm{max}(\mathrm{abs}((\mathrm{q}_{j} - \mathrm{q}^n_{j})_{HFE}))$ & $\pi/4$ & $\pi/12$ 
                     \rule{0pt}{2.6ex} \\
                     $\mathrm{max}(\mathrm{abs}((\mathrm{q}_{j} - \mathrm{q}^n_{j})_{KFE}))$ & $\pi/3$ & $\pi/12$ 
                     \rule{0pt}{2.6ex} \\ 
                     $|\mathrm{cos}^{-1}(\mathbf{e}_z^{B}\cdot\mathbf{e}_z^{W})|$ & $\pi/3$ & $\pi/4$ \rule{0pt}{2.6ex} \\ 
                     $|\mathrm{cos}^{-1}(\mathbf{e}_z^{B}\cdot\mathbf{e}_z^{C})|$ & $\pi/4$ & $\pi/6$ \rule{0pt}{2.6ex} \\
                    \hline
                \end{tabular}
                \label{tab:recovery_activation_deactivation}
            \end{table}

\subsection{Terrain Generation}
\begin{figure}
            \begin{subfigure}{.23\textwidth}
              \centering
              \includegraphics[width=.75\linewidth]{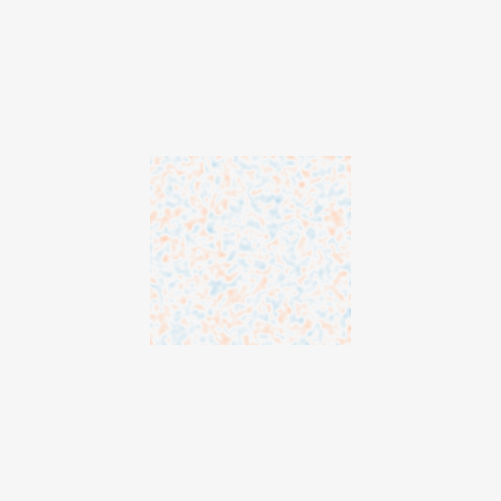}
              \label{fig:terrains_unstructured}
            \end{subfigure}%
            \begin{subfigure}{.23\textwidth}
              \centering
              \includegraphics[width=.75\linewidth]{figures/terrains_unstructured_sim.png}
              \label{fig:terrains_unstructured_sim}
            \end{subfigure} \vspace{0.1cm} \\
            \begin{subfigure}{.23\textwidth}
              \centering
              \includegraphics[width=.75\linewidth]{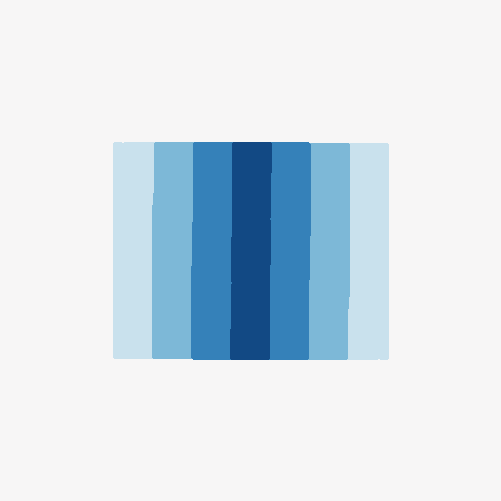}
              \label{fig:terrains_stairs}
            \end{subfigure}
            \begin{subfigure}{.23\textwidth}
              \centering
              \includegraphics[width=.75\linewidth]{figures/terrain_stairs_sim.png}
              \label{fig:terrains_stairs_sim}
            \end{subfigure} \vspace{0.1cm} \\
            \begin{subfigure}{.23\textwidth}
              \centering
              \includegraphics[width=.75\linewidth]{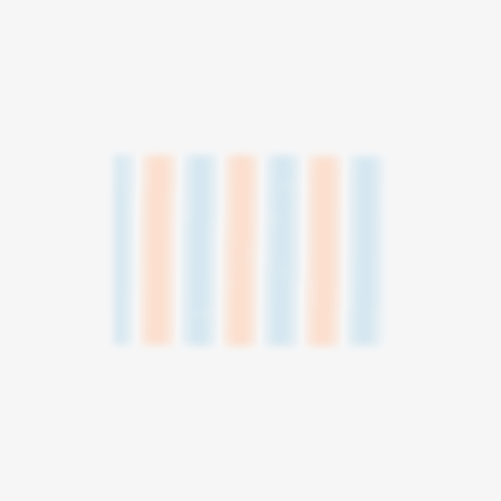}
              \label{fig:terrains_wave}
            \end{subfigure}%
            \begin{subfigure}{.23\textwidth}
              \centering
              \includegraphics[width=.75\linewidth]{figures/terrains_wave_sim.png}
              \label{fig:terrains_wave_sim}
            \end{subfigure} \vspace{0.1cm} \\
            \begin{subfigure}{.23\textwidth}
              \centering
              \includegraphics[width=.75\linewidth]{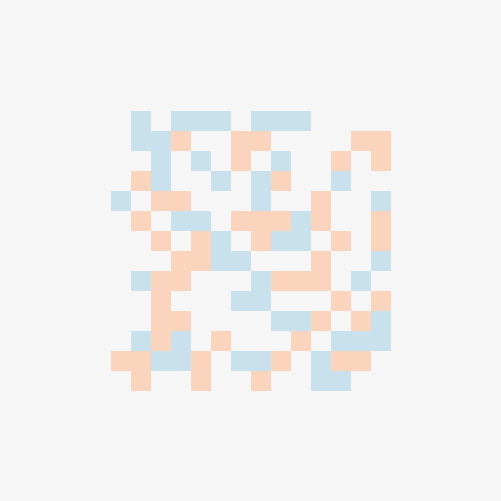}
              \label{fig:terrains_bricks}
            \end{subfigure}%
            \begin{subfigure}{.23\textwidth}
              \centering
              \includegraphics[width=.75\linewidth]{figures/terrains_bricks_sim.png}
              \label{fig:terrains_bricks_sim}
            \end{subfigure}%
            \caption{\textbf{(top-bottom)} represent the terrains comprising of unstructured ground, stairs, wave and brick objects respectively. 
            The left images represent the
            height maps imported in the RaiSim simulator as shown on the right.}
            \label{fig:terrains_png_sim}
        \end{figure}

        For terrain generation,
        we consider several objects such as a staircase,
        planks, bricks,
        unstructured sub-terrain and wave
        sub-terrain. Each terrain contains at least 1 and at most 5 objects. We position
        and rotate
        these objects within the
        terrain area to limit height
        deviations around close regions and increase the occupancy over the entire terrain,
        and represent their corresponding heights as elevation
        in the range of $[0, 65535]$ where 0 corresponds to no elevation
        and 65535 corresponds to an elevation of \SI{2}{\metre}. We automate this generation process using a Python tool 
        which stores a wide range of terrains, 10k in our case, in the form of single channel 16-bit PNG files. These PNGs are 
        then
        imported in RaiSim as heightmaps.
        Figure~\ref{fig:terrains_png_sim} represents
        the generated terrains for the unstructured ground, stairs, wave and brick objects along with their corresponding
        visualization in RaiSim.

\subsection{Actuation Dynamics}
The actuator network utilized for training footstep planner and domain adaptive tracker models an adaptive impedance controller which can be represented 
as \begin{equation}
            \tau_j = \mathbf{j}_R(\mathrm{q}_j^\ast, \mathrm{q}_j, \mathrm{\dot{q}}_j^\ast, 
            \mathrm{\dot{q}}_j, \tau_{j_{FF}}, K_p, K_d) \ .
\end{equation} The input to the actuator network is thus represented as $j_s\in\mathbb{R}^{21}$ such that
        \begin{equation}
            j_s = \langle s_{j_{pe}}, s_{j_{ve}}, s_{j_{te}}, s_{j_{vd}}, s_{j_{td}}, s_{j_{kp}}, s_{j_{kd}} \rangle \ .
        \end{equation}
        The definitions of the input terms are provided in Table~\ref{tab:actuator_network_input_defintions}.
        This actuator network contains 2 hidden layers 
        with 48 nodes in each of the layers and a 1-dimensional output layer representing the estimate of the joint
        torque. We perform a forward pass through the network at each time step for each of the 12
        joints of the quadruped.
        
        \begin{table}[h!]
                \centering
                \caption{Actuator input term definitions. Here, $i$ refers to the $i$-th joint. Subscript 
                $t$ refers to the measurement at the current
                control step and the 
                duration between $t-1$ and $t$ corresponds to \SI{2.5}{\milli\second}.}
                \begin{tabular}{|c|}
                    \hline
                     Actuator Network Input Terms \rule{0pt}{2.6ex} \rule[-0.9ex]{0pt}{0pt} \\
                     \hline
                     $ s_{j_{pe}}=[({\mathrm{q}^\ast_{j,i}}_{t} - {\mathrm{q}_{j,i}}_{t})^T \ \ ({\mathrm{q}^\ast_{j,i}}_{t-4} 
                     - {\mathrm{q}_{j,i}}_{t-4})^T \  \ ({\mathrm{q}^\ast_{j,i}}_{t-8} - {\mathrm{q}_{j,i}}_{t-8})^T]^T$ 
                     \rule{0pt}{2.6ex} 
                     \\
                     $ s_{j_{ve}}=[({\mathrm{\dot{q}}^\ast_{j,i}}{}_{t} - {\mathrm{\dot{q}}_{j,i}}{}_{t})^T \ \ 
                     (\mathrm{\dot{q}}^\ast_{{j,i}_{t-4}} -
                     \mathrm{\dot{q}}_{{j,i}_{t-4}})^T \  \ (\mathrm{\dot{q}}^\ast_{{j,i}_{t-8}} - 
                     \mathrm{\dot{q}}_{{j,i}_{t-8}})^T]^T$ 
                     \rule{0pt}{2.6ex} 
                     \\
                     $ s_{j_{te}}=[({\mathrm{\tau}_{j,i}}_{t} - {\mathrm{\tau}_{{j,i}_{FF}}}_{t})^T \ \ ({\mathrm{\tau}_{j,i}}_{t-4} - {\mathrm{\tau}_{{j,i}_{FF}}}_{t-4})^T \  \ ({\mathrm{\tau}_{j,i}}_{t-8} - {\mathrm{\tau}_{{j,i}_{FF}}}_{t-8})^T]^T$ 
                     \rule{0pt}{2.6ex} 
                     \\
                     $ s_{j_{vd}}=[(\mathrm{\dot{q}}^\ast_{{j,i}_{t}})^T \ \ 
                     (\mathrm{\dot{q}}^\ast_{{j,i}_{t-4}})^T \  \ (\mathrm{\dot{q}}^\ast_{{j,i}_{t-8}})^T]^T$ 
                     \rule{0pt}{2.6ex} 
                     \\
                     $ s_{j_{td}}=[{\mathrm{\tau}^T_{{j,i}_{FF}}}_{t} \ \ {\mathrm{\tau}^T_{{j,i}_{FF}}}_{t-4} \  \ {\mathrm{\tau}^T_{{j,i}_{FF}}}_{t-8}]^T$ 
                     \rule{0pt}{2.6ex} 
                     \\
                    $ s_{j_{kp}}=[{K_p}^T_{t} \ \ {K_p}^T_{t-4} \  \ {K_p}^T_{t-8}]^T$  \rule{0pt}{2.6ex} \\
                    $ s_{j_{kd}}=[{K_d}^T_{t} \ \ {K_d}^T_{t-4} \  \ {K_d}^T_{t-8}]^T$  \rule{0pt}{2.6ex} 
                    \rule[-0.9ex]{0pt}{0pt} \\
                     \hline 
                \end{tabular}
                \label{tab:actuator_network_input_defintions}
            \end{table}

        For training the adaptive impedance actuator network, we collected data from the real system using the dynamic gaits motion controller
        to track randomly sampled base reference velocities. We recorded the desired
        relevant input quantities at \SI{400}{\hertz} for 12 actuators thus generating
        a large dataset of 8.64M samples in approximately 30 minutes. However, since the impedance gains 
        depend on the contact state of the
        robot, during data collection, we obtained sufficient samples for the swing 
        and stance phases. However, during operation, we obtained significantly less samples for the
        contact-uncertainty state. As a result, using the actuator network trained with the naive supervised learning
        approach failed to properly track actuation commands for contact-uncertainty states in the simulator causing the motion controller
        to generate unstable motion tracking commands eventually resulting in failure. 
        To address this, we utilized SMOGN~\cite{branco2017smogn, smogn-python} to perform regression over imbalanced data by 
        resampling the dataset generated from the real robot. SMOGN utilizes a random undersampling approach for the majority
        samples and a synthetic sample generation method for the minority samples. Table~\ref{table:actuator_netwowrk_error} presents
        the mean absolute prediction error comparing the performance of networks trained using a naive supervised learning approach with resampled
        learning, tested on 50k samples for each of the different sets of PD gains.
        \begin{center}
        \begin{table}
            \caption{Effective joint torque prediction error comparing the actuator network performance for training with and 
            without re-sampling of the training dataset using SMOGN. Here $\mu_{e}$ is the mean prediction error and $\sigma_{e}$
            is the standard deviation of the prediction error observed over
            50k test samples.}
            \resizebox{0.48\textwidth}{!}{%
            \centering
            \begin{tabular}{ |c|c|c|c| }
                \hline
                 & \textbf{$K_{p}$=8, $K_{d}$=0.1} & \textbf{$K_{p}$=35, $K_{d}$=0.4} & \textbf{$K_{p}$=160, $K_{d}$=0.3} \\
                \hline
                \textbf{Baseline} & $\mu_{e}=0.82$, $\sigma_{e}=0.34$ & $\mu_{e}=0.77$, $\sigma_{e}=0.31$ & $\mu_{e}=1.73$, $\sigma_{e}=0.85$ \\ 
                \hline
                \textbf{SMOGN} & $\mu_{e}=0.83$, $\sigma_{e}=0.21$ & $\mu_{e}=0.71$, $\sigma_{e}=0.19$ & $\mu_{e}=0.96$, $\sigma_{e}=0.39$ \\
                \hline
            \end{tabular}}
            \label{table:actuator_netwowrk_error}
        \end{table}
        \end{center}

\subsection{Training}
Our motivation for selecting an RL training strategy is based
on three main factors: (1) convergence to a desired behavior,
(2) sample efficiency, and (3) training time.

We use SAC for training
our footstep planning policy mainly for its improved sample
efficiency over the widely used TRPO and PPO approaches. In our
preliminary experiments for training the footstep
planner, we observed that
SAC required approximately 
a fourth of policy iterations for convergence
compared to PPO for similar training parameters. For this
reason, we decided to use SAC for training our 
footstep planner.

In the case of domain adaptive tracker, we observed that
the stochasticity associated with SAC and PPO necessitated
precise hyperparamter tuning, without which the policy failed
to exhibit a stable locomotion behavior.
Even then, we observed little improvement
in domain adaptive tracking. This was largely due to
the domain adaptive tracker generating large torques which caused the
whole-body controller to aggressively correct for inaccurate tracking resulting in
episode termination.
Instead, we used TD3 which allowed us
to perform exploration using an uncorrelated Gaussian noise.
We thus employed a curriculum based exploration strategy
where we initialized the policy to output values close to 0s, while also
setting a small value for the Gaussian noise standard deviation. As training
progressed, we increased the standard deviation. This allowed for stable learning
and thus better convergence.

For the recovery controller, we use PPO for 2 main reasons: (1) Prior work
on end-to-end state-to-joint RL control has utilized PPO and provided 
reproducible steps~\cite{hwangbo2019learning}, and (2) we observed that, compared 
to off-policy approaches such as SAC and TD3,
PPO required significantly less time to perform policy updates
after each episode rollout. Since the footstep planner and domain adaptive
tracker employed the motion controller during training, much of the 
training overhead occurred during simulation rollouts as opposed to policy
updates. The recovery controller rollouts were significantly fast since no
additional computation, such as motion optimization, was required. PPO
thus enabled us to perform faster training even with reduced sample efficiency.
In our preliminary experiments,
we observed that for each policy iteration, PPO required 
approximately~\SI{3.2}{\second} whereas SAC required
approximately~\SI{13}{\second} for similar training parameters.
Although it is possible to reduce the policy update duration
for SAC with additional parameter tuning, for our test case, PPO
still offered policy convergence in less training time than SAC. We
therefore decided to use PPO for training the recovery controller.

We trained each of our RL policies on a desktop computer
with an Intel i7-8700 processor clocked at \SI{4.0}{\giga\hertz} and
an NVidia RTX 2080Ti graphics card. The footstep planner Session 1
training time and hyperparameters are shown in Table~\ref{tab:hyperparameters_footstep_planner}.
For training Session 2, we use the SAC hyperparameters shown in Table~\ref{tab:hyperparameters_footstep_planner}
for additional 100 policy update iterations requiring roughly 4 hours of training.
For training the domain adaptive tracking policy, we use the parameters provided in
Table~\ref{tab:hyperparameters_domain_adaptive_tracker}, and we use the parameters
provided in Table~\ref{tab:hyperparameters_recovery_controller} for training the recovery
controller.

\begin{table}[h!]
    \caption{Footstep planner Session 1 training time and hyperparameters.}
    \centering
        \begin{tabular}{|c|c|}
            \hline
            Parameter & Value \rule{0pt}{2.6ex} \rule[-0.9ex]{0pt}{0pt} \\
            \hline
            Discount factor, $\gamma$ & 0.994 \rule{0pt}{2.6ex} \\
            Learning rate & 1e-4 \rule{0pt}{2.6ex} \\
            Buffer size & 1e5 \rule{0pt}{2.6ex} \\
            Batch size & 8192 \rule{0pt}{2.6ex} \\
            Prioritized replay buffer & False \rule{0pt}{2.6ex} \\
            Learning starts & 24576 \rule{0pt}{2.6ex} \\
            Parallel environments, $n_{env}$ & 24 \rule{0pt}{2.6ex} \\
            Soft update coefficient, $\tau$ & 1e-3 \rule{0pt}{2.6ex} \\
            Training frequency & 1 \rule{0pt}{2.6ex} \\
            Twin Q functions & True \rule{0pt}{2.6ex} \\
            Target entropy & Auto \rule{0pt}{2.6ex} \\
            IGPO $\lambda_d$ & 0.996 \rule{0pt}{2.6ex} \\
            IGPO $n_{eps}$ & 100 \rule{0pt}{2.6ex} \\
            Steps per iteration, $l_{eps}$ & $1024\times n_{env}$ \rule{0pt}{2.6ex} \\
            SAC time per iteration & $\sim$\SI{120}{\second} \rule{0pt}{2.6ex} \\
            Guided learning time per iteration & $\sim$\SI{105}{\second} \rule{0pt}{2.6ex} \\
            Total training time & $\sim$\SI{28}{\hour} \rule{0pt}{2.6ex} \\
            \hline 
        \end{tabular}
\label{tab:hyperparameters_footstep_planner}
\end{table}

\begin{table}[h!]
    \caption{Domain adaptive tracker training time and hyperparameters.}
    \centering
        \begin{tabular}{|c|c|}
            \hline
            Parameter & Value \rule{0pt}{2.6ex} \rule[-0.9ex]{0pt}{0pt} \\
            \hline
            Discount factor, $\gamma$ & 0.998 \rule{0pt}{2.6ex} \\
            Learning rate & 5e-4 \rule{0pt}{2.6ex} \\
            Buffer size & 1e6 \rule{0pt}{2.6ex} \\
            Batch size & 19200 \rule{0pt}{2.6ex} \\
            Learning starts & 76800 \rule{0pt}{2.6ex} \\
            Parallel environments, $n_{env}$ & 24 \rule{0pt}{2.6ex} \\
            Soft update coefficient, $\tau$ & 5e-3 \rule{0pt}{2.6ex} \\
            Prioritized replay & True \rule{0pt}{2.6ex} \\
            Prioritized replay $\alpha$ & 0.6 \rule{0pt}{2.6ex} \\
            Prioritized replay $\beta$ & 0.6 \rule{0pt}{2.6ex} \\
            Twin Q functions & True \rule{0pt}{2.6ex} \\
            Steps per iteration, $l_{eps}$ & $3200\times n_{env}$ \rule{0pt}{2.6ex} \\
            TD3 time per iteration & $\sim$\SI{4}{\second} \rule{0pt}{2.6ex} \\
            Total training time & $\sim$\SI{22}{\hour} \rule{0pt}{2.6ex} \\
            \hline 
        \end{tabular}
\label{tab:hyperparameters_domain_adaptive_tracker}
\end{table}

\begin{table}[h!]
    \caption{Recovery controller training time and hyperparameters.}
    \centering
        \begin{tabular}{|c|c|}
            \hline
            Parameter & Value \rule{0pt}{2.6ex} \rule[-0.9ex]{0pt}{0pt} \\
            \hline
            Discount factor, $\gamma$ & 0.996 \rule{0pt}{2.6ex} \\
            Learning rate & 5e-4 \rule{0pt}{2.6ex} \\
            Batch size & 115200 \rule{0pt}{2.6ex} \\
            Mini-batch size & 14400 \rule{0pt}{2.6ex} \\
            Parallel environments, $n_{env}$ & 48 \rule{0pt}{2.6ex} \\
            Entropy coefficient & 5e-3 \rule{0pt}{2.6ex} \\
            Value coefficient & 0.25 \rule{0pt}{2.6ex} \\
            GAE & True \rule{0pt}{2.6ex} \\            
            GAE $\lambda$ & 0.98 \rule{0pt}{2.6ex} \\
            Steps per iteration, $l_{eps}$ & $2400\times n_{env}$ \rule{0pt}{2.6ex} \\
            PPO time per iteration & $\sim$\SI{3}{\second} \rule{0pt}{2.6ex} \\
            Total training time & $\sim$\SI{6}{\hour} \rule{0pt}{2.6ex} \\
            \hline 
        \end{tabular}
\label{tab:hyperparameters_recovery_controller}
\end{table}

\section*{Acknowledgment}
S.G. trained the actuator network, stability margin network and the denoising convolutional autoencoder. S.G. setup the 
RL environments and performed training of footstep planning, domain adaptive tracking and recovery control policies. M.G.
extended the model-based motion controller for locomotion over uneven terrain and helped set up
simulation environments. R.O. extended the stability margin 
analysis to dynamic motions with 2 feet contacts. S.G. and M.G. wrote the control software for the real robot. 
S.G., M.G and R.O. performed hardware experiments. I.H. and M.F.
provided funding and supervised the project. S.G. took the lead in writing the manuscript. M.G., R.O.,
M.F. and I.H. provided critical feedback and helped shape the manuscript.

\bibliographystyle{IEEEtran}
\bibliography{references}

\vspace{-0.7cm}
\begin{IEEEbiography}[{\includegraphics[width=1in,height=1.25in,clip,keepaspectratio]{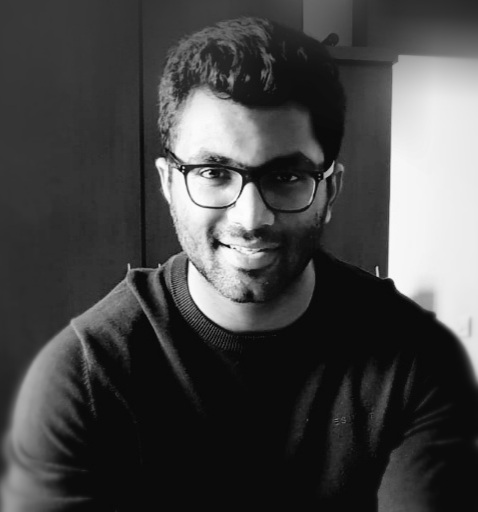}}]{Siddhant Gangapurwala}
completed his Bachelor of Engineering (B.E.) in Electronics from the University of Mumbai in 2016. In 2017, he joined the AIMS program as a doctoral candidate at the University of Oxford. The following year, Siddhant joined the Dynamic Robot Systems (DRS) group of the Oxford Robotics Insitute (ORI) to pursue his DPhil degree with focus on machine learning and optimal control based approaches for robotic locomotion over uneven terrain where he continues to work as a post-doctoral researcher on locomotion, manipulation and loco-manipulation.
\end{IEEEbiography}
\vspace{-0.78cm}
\begin{IEEEbiography}[{\includegraphics[width=1in,height=1in,clip,keepaspectratio]{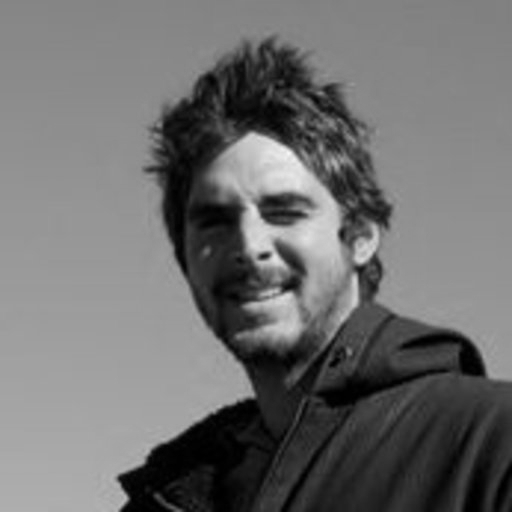}}]{Mathieu Geisert}
received his M. Eng. in Aerospace from Institut Supérieur de l’Aéronautique et de l’Espace SUPAERO (Toulouse, France) in 2013. 
He then joined the humanoid robotics team, GEPETTO, at LAAS-CNRS (Toulouse, France) where he worked for 5 years in different 
positions: as an intern, engineer, Ph.D. student and finally as a post doctoral researcher. In 2018, he joined the DRS group
as a post-doctoral researcher where his work focused on Planning, Optimal Control and Machine Learning for legged robots. 
In 2021, Mathieu joined Arrival, London, UK as a robotics research engineer.\end{IEEEbiography}
\vspace{-0.78cm}
\begin{IEEEbiography}
[{\includegraphics[width=1in,height=1.25in,clip,keepaspectratio]{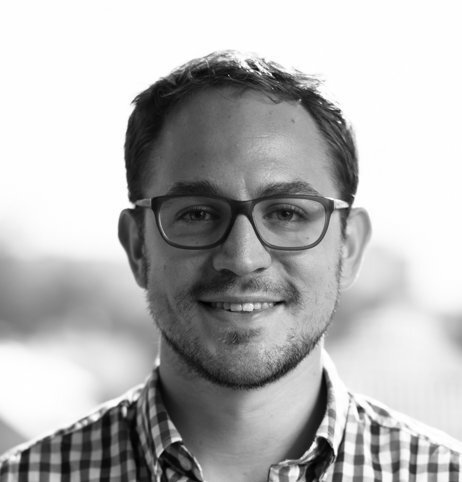}}]{Romeo Orsolino}
completed his B.Sc. in mechanical engineering from the University of Genova in 2013 and his M.Sc. in robotics engineering at the \'{E}cole Centrale de Nantes in 2015. In the same year, he  joined DLS at IIT for a Ph.D. focusing on motion planning for legged locomotion in rough terrains. In October 2019, he joined the DRS group as a post-doctoral researcher where he pursued his research on multi-contact motion planning, optimal control, dynamics and perception. In December 2020, Romeo joined Arrival Ltd. where continues to work on perceptive motion planning and control of dynamical systems.
\end{IEEEbiography}
\begin{IEEEbiography}
[{\includegraphics[width=1in,height=1.25in,clip,keepaspectratio]{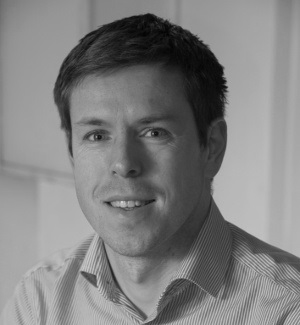}}]{Maurice Fallon}
studied Electronic Engineering at University College Dublin. His Ph.D. research in the field of acoustic source tracking was carried out in the Engineering Department of the University of Cambridge. After his Ph.D. he moved to MIT as a post doctoral researcher and later research scientist in the Marine Robotics Group (2008-2012). From 2012-2015 he was the perception lead of MIT’s team in the DARPA Robotics Challenge. From 2015, he was a Lecturer at University of Edinburgh where he led research in collaboration with NASA’s humanoid robotics program. He moved to Oxford in April 2017 and took up the Royal Society University Research Fellowship in October 2017.
\end{IEEEbiography}
\vspace{-1.1cm}
\begin{IEEEbiography}
[{\includegraphics[width=1in,height=1.25in,clip,keepaspectratio]{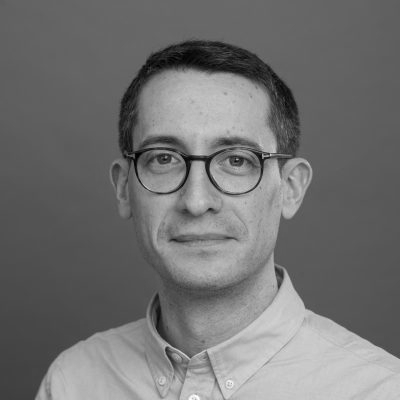}}]{Ioannis Havoutis}
is a Departmental Lecturer at the Oxford Robotics Institute. His research combines dynamic whole-body motion planning and control with machine learning, focusing on robots with arms and legs. He received his Ph.D. (2011) and M.Sc. with distinction (2007) from the University of Edinburgh, where he worked on machine learning for motion planning and control of articulated robots. He then joined the Dynamic Legged Systems Lab at the Advanced Robotics Department of IIT, where he led the Locomotion Group within the HyQ team.
He was a postdoctoral researcher at the Robot Learning \& Interaction Group of the Idiap Research Institute, where he worked on learning complex skills from demonstration.
\end{IEEEbiography}
\enlargethispage{-15cm}
\end{document}